%% file: spatial_smoothing_icml.tex
\documentclass{article}

\usepackage{microtype}
\usepackage{graphicx}
\usepackage{booktabs} %

\usepackage{hyperref}
\hypersetup{
    colorlinks=true,
	citecolor=MidnightBlue,
	linkcolor=OrangeRed,
	urlcolor=OrangeRed
}

\usepackage[dvipsnames]{xcolor}     %

\usepackage[accepted,nohyperref]{icml2022}

\usepackage{amsmath}
\usepackage{amssymb}
\usepackage{mathtools}
\usepackage{amsthm}

\usepackage{url}            %
\usepackage{booktabs}       %
\usepackage{multirow}       
\usepackage{amsfonts}       %
\usepackage{amssymb}        %
\usepackage{dsfont}         %
\usepackage{nicefrac}       %
\usepackage{microtype}      %

\usepackage{bm} 			%
\usepackage{kotex} 		%
\usepackage{xfrac} 		%
\usepackage[english]{babel}         %
\usepackage{blindtext} 	%
\usepackage{soul} 		%
\usepackage[normalem]{ulem}         %
\usepackage{csquotes} 	%
\usepackage{wrapfig}    %
\usepackage[inline]{enumitem}       %
\usepackage{aliascnt}   %
\usepackage[capitalise,nameinlink]{cleveref}   %
\usepackage{caption}    %
\usepackage{subcaption} %
\usepackage{makecell}   %
\usepackage{rotating}   %
\usepackage{chngcntr}

\makeatletter
\newtheoremstyle{indented}
  {4pt}  %
  {-1pt}  %
  {\itshape
   \addtolength{\@totalleftmargin}{1.0em}
   \addtolength{\linewidth}{-1.0em}
   \parshape 1 1.0em \linewidth}  %
  {}  %
  {\bfseries}  %
  {.}  %
  {.5em}  %
  {}  %
\makeatother

\theoremstyle{indented}

\crefname{thm}{theorem}{theorems}
\Crefname{thm}{Theorem}{Theorems}

\newaliascnt{lemma}{thm}

\aliascntresetthe{lemma}

\newaliascnt{prop}{thm}
\newtheorem{prop}[prop]{Proposition} 
\aliascntresetthe{prop}
\crefname{prop}{Proposition}{Propositions}
\Crefname{prop}{Proposition}{Propositions}

\usepackage{tikz}
\usetikzlibrary{shapes.misc,shadows}

\DeclareRobustCommand{\tcblack}[1]{
\begin{tikzpicture}[baseline=(char.base)]
\node(char)[
  draw,fill=black,
  shape=rounded rectangle,
  text height=5pt,
  drop shadow={opacity=.5,shadow xshift=0pt,shadow yshift=-0.6pt},
]
  {\color{white}{\normalfont #1}};
\end{tikzpicture}
}

\renewcommand{\algorithmicrequire}{\textbf{Input:}}
\renewcommand{\algorithmicensure}{\textbf{Output:}}

\icmltitlerunning{Blurs Behave Like Ensembles} 

\begin{document}

\twocolumn[
\icmltitle{Blurs Behave Like Ensembles: \\Spatial Smoothings to Improve Accuracy, Uncertainty, and Robustness}

\icmlsetsymbol{equal}{*}

\begin{icmlauthorlist}
\icmlauthor{Namuk Park}{comp}
\icmlauthor{Songkuk Kim}{yyy}
\end{icmlauthorlist}

\icmlaffiliation{yyy}{Yonsei University}
\icmlaffiliation{comp}{NAVER AI Lab}

\icmlcorrespondingauthor{Namuk Park}{namuk.park@navercorp.com}
\icmlcorrespondingauthor{Songkuk Kim}{songkuk@yonsei.ac.kr}

\icmlkeywords{Machine Learning, }

\vskip 0.3in
]

\printAffiliationsAndNotice{}  %

\input{meta/abstract}

\input{body/introduction}

\input{body/method}

\input{body/experiment}

\input{body/related-work}

\input{body/discussion}

\section*{Acknowledgement}

We thank the reviewers for valuable feedback. 
This research was supported by Basic Science Research Program through the National Research Foundation of Korea (NRF) funded by the Ministry of Education (2021R1A6A3A13045325).

\nocite{langley00}

\bibliography{meta/references.bib}
\bibliographystyle{icml2022}

\clearpage
\appendix
\counterwithin{figure}{section}
\counterwithin{table}{section}

\section*{Appendix Overview}

\cref{sec:conf} provide comprehensive resources, such as experimental details, to ensure reproducibility. In particular, \cref{sec:conf} provides the specifications of all models used in this work and detailed hyperparameter setups. 
Code is available at \url{https://github.com/xxxnell/spatial-smoothing}.

\cref{sec:ablation} provides the results of ablation studies. We report the predictive performances of \texttt{Prob} and \texttt{Blur} with various hyperparameters, and investigate several types of edge cases. 

\cref{sec:revisiting} further discusses prior works---namely, global average pooling, pre-activation, and \texttt{ReLU6}---as special cases of spatial smoothing. In particular, we provide numerical results to demonstrate that these methods improve accuracy, uncertainty estimation, and robustness simultaneously.

\cref{sec:extended-analysis} mainly provides rigorous discussions of three key properties of ensembles: \cref{thm:variance}, \cref{thm:frequency}, and \cref{thm:loss-landscape}. If a NN has these three properties at the same time, we can infer that the NN exploits the ensemble effect. Since these properties are necessary conditions for ensembles, they can be regarded as a checklist for ensembles.

\cref{sec:extended} provides detailed results of experiments, e.g., image classification and semantic segmentation. The results include predictive performances on various settings and robustness on corrupted datasets.

\cref{sec:prob-role} demonstrates that \texttt{Prob} plays an important role in spatial smoothing. \texttt{Blur} alone can improve  predictive performance of conventional CNNs only when activation ($\texttt{ReLU} \circ \texttt{BatchNorm}$) acts as \texttt{Prob}; however, otherwise, \texttt{Blur} harms the predictive performance. For example, \texttt{Blur} degrades the performance of pre-activation CNNs.

\input{appendix/experimental-setup}

\input{appendix/ablation-study}

\input{appendix/revisiting-gap}

\input{appendix/extended-analysis}

\input{appendix/extended-experiment}

\input{appendix/prob-analysis}

\end{document}

%% file: meta/abstract.tex
\begin{abstract}

Neural network ensembles, such as Bayesian neural networks (BNNs), have shown success in the areas of uncertainty estimation and robustness. However, a crucial challenge prohibits their use in practice. BNNs require a large number of predictions to produce reliable results, leading to a significant increase in computational cost. 
To alleviate this issue, we propose \emph{spatial smoothing}, a method that spatially ensembles neighboring feature map points of convolutional neural networks. By simply adding a few blur layers to the models, we empirically show that spatial smoothing improves accuracy, uncertainty estimation, and robustness of BNNs across a whole range of ensemble sizes. 
In particular, BNNs incorporating spatial smoothing achieve high predictive performance merely with a handful of ensembles. Moreover, this method also can be applied to canonical deterministic neural networks to improve the performances. A number of evidences suggest that the improvements can be attributed to the stabilized feature maps and the smoothing of the loss landscape. 
In addition, we provide a fundamental explanation for prior works---namely, global average pooling, pre-activation, and \texttt{ReLU6}---by addressing them as special cases of spatial smoothing. These not only enhance accuracy, but also improve uncertainty estimation and robustness by making the loss landscape smoother in the same manner as spatial smoothing.

\end{abstract}

%% file: body/introduction.tex
\section{Introduction}
\label{sec:introduction}

\input{resources/fig-concept}

In a real-world environment where many unexpected events occur, machine learning systems cannot be guaranteed to always produce accurate predictions. 
In order to handle this issue, we make system decisions more reliable by considering estimated uncertainties, in addition to predictions. Uncertainty quantification is particularly crucial in building a trustworthy system in the field of safety-critical applications, including medical analysis and autonomous vehicle control. 
However, canonical deep neural networks (NNs)---or deterministic NNs---cannot produce reliable estimations of uncertainties \citep{guo2017calibration}, and their accuracy is often severely compromised by natural data corruptions from noise, blur, and weather changes \citep{engstrom2019exploring,azulay2018deep}.

Bayesian neural networks (BNNs), such as Monte Carlo (MC) dropout \citep{gal2016dropout}, provide a probabilistic representation of NN weights. They aggregate a number of models selected based on weight probability to make predictions of desired results. 
Thanks to this feature, BNNs have been widely used in the areas of uncertainty estimation \citep{kendall2017uncertainties} and robustness \citep{ovadia2019can}. They are also promising in other fields like out-of-distribution detection \citep{malinin2018predictive} and meta-learning \citep{yoon2018bayesian}.

Nevertheless, there remains a significant challenge that prohibits their use in practice. BNNs require an ensemble size of up to fifty to achieve high predictive performance, which results in a fiftyfold increase in computational cost \citep{kendall2017uncertainties,loquercio2020general}. Therefore, if BNNs can achieve high predictive performance merely with a handful of ensembles, they could be applied to a much wider range of areas.

\subsection{Preliminary}\label{sec:introduction:prelim}

We would first like to discuss canonical BNN inference in detail, then move on to vector quantized BNN (VQ-BNN) inference \citep{park2019vqbnn}, an efficient approximated BNN inference.

\paragraph{Ensemble averaging for a single data point.}

Suppose we have access to model uncertainty, i.e., posterior probability of NN weight $p(\bm{w} \vert \mathcal{D})$ for training dataset $\mathcal{D}$. The predictive result of BNN is given by the following predictive distribution:
\begin{equation}
	p(\bm{y} \vert \bm{x}_{0}, \mathcal{D}) = \int p(\bm{y} \vert \bm{x}_{0}, \bm{w}) \, p(\bm{w} \vert \mathcal{D}) \, d\bm{w} \label{eq:bnn-inference}
\end{equation}
where $\bm{x}_{0}$ is observed input data vector, $\bm{y}$ is output vector, and $p(\bm{y} \vert \bm{x}, \bm{w})$ is the probabilistic prediction parameterized by the result of NN for an input $\bm{x}$ and weight $\bm{w}$. In most cases, the integral cannot be solved analytically. Thus, we use the MC estimator to approximate it as follows:
\begin{equation}
	p(\bm{y} \vert \bm{x}_{0}, \mathcal{D}) \simeq \sum_{i = 0}^{N-1} \frac{1}{N} \, p(\bm{y} \vert \bm{x}_{0}, \bm{w}_{i}) \label{eq:approx-bnn-inference}
\end{equation}
where $\bm{w}_{i} \sim p(\bm{w} \vert \mathcal{D})$ and $N$ is the number of the samples. This equation indicates that \emph{BNN inference is ensemble average of NN predictions for ``one observed data point''} $\bm{x}_0$ as shown on the left of \cref{fig:diagram}. Using $N$ neural networks in the ensemble would requires $N$ times more computational complexity than one NN execution.

\paragraph{Ensemble averaging for proximate data points.}

To reduce the computational cost of BNN inference, \emph{VQ-BNN \citep{park2019vqbnn} executes NN for ``an observed data point'' $\bm{x}_{0}$ only once, and complements the result with previously calculated predictions for ``other data points'' $\bm{x}_{i}$} as follows:
\begin{equation}
	p(\bm{y} \vert \bm{x}_{0}, \mathcal{D}) \simeq \sum_{i = 0}^{N - 1} \pi(\bm{x}_{i} \vert \bm{x}_{0}) \, p(\bm{y} \vert \bm{x}_{i}, \bm{w}_{i})
	\label{eq:approx-vqbnn-inference}
\end{equation}
where $\pi(\bm{x}_{i} \vert \bm{x}_{0})$ is the importance of data $\bm{x}_{i}$ with respect to the observed data $\bm{x}_{0}$, and it is defined as a similarity between $\bm{x}_{i}$ and $\bm{x}_{0}$. 
If we have access to previous predictions $\{ p(\bm{y} \vert \bm{x}_{1}, \bm{w}_{1}), \cdots \}$, the computational performance of VQ-BNN becomes comparable to that of one NN execution to obtain the newly calculated prediction $p(\bm{y} \vert \bm{x}_{0}, \bm{w}_{0})$.
To accurately infer the results, \emph{the previous predictions should consist of predictions for ``data similar to the observed data''}, i.e., $\bm{x}_i = \bm{x}_0 + \bm{\varepsilon}_i$ for small but non-zero $\bm{\varepsilon}_i$. The distribution of the proximate data points is called data uncertainty \citep{park2019vqbnn}.

Thanks to the temporal consistency of real-world data streams, aggregating predictions for similar data in data streams is straightforward. Since temporally proximate data sequences tend to be similar, we can memorize recent predictions and calculates their average using exponentially decreasing importance. In other words, \emph{VQ-BNN inference for data streams is simply temporal smoothing of recent predictions} as shown in the middle of \cref{fig:diagram}.

VQ-BNN has two limitations, although it may be a promising approach to obtain reliable results in an efficient way.
First, it was only applicable to data streams such as video sequences. Applying VQ-BNN to static images is challenging because it is impossible to memorize all similar images in advance. Second, \citet{park2019vqbnn} used VQ-BNN only in the testing phase, not in the training phase. We find that ensembling predictions for similar data helps in NN training by smoothing the loss landscape.

\begin{figure}

\centering

\vskip 0.2in
\raisebox{0pt}[\dimexpr\height-0.6\baselineskip\relax]{
\includegraphics[width=0.385\textwidth]{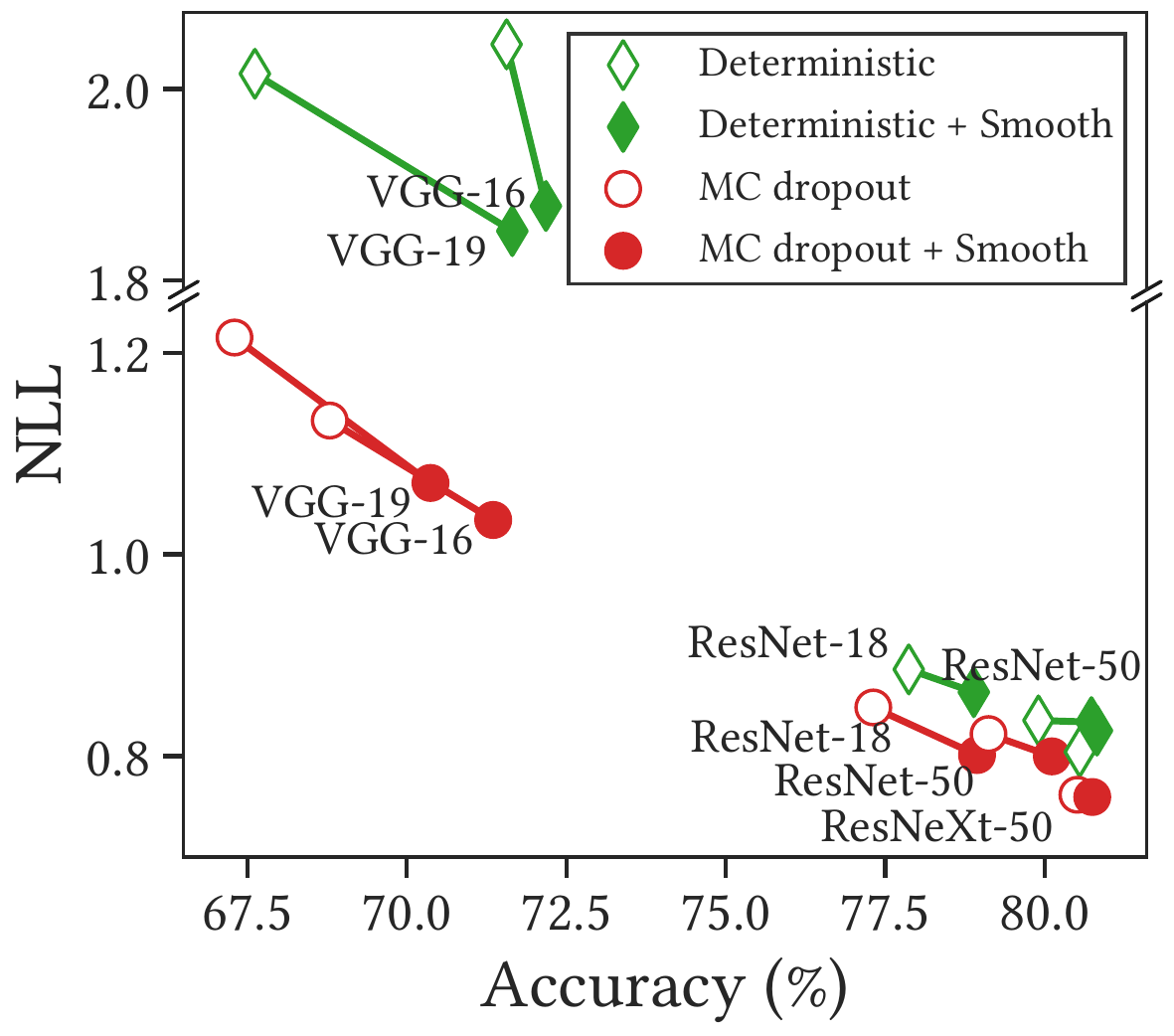}
}

\vspace{0pt}
\caption{
\textbf{Spatial smoothing improves both accuracy and uncertainty (NLL)}.
\texttt{Smooth} means spatial smoothing. 
Downward from left to the right ($\searrow$) means better accuracy and uncertainty.
}
\vskip -0.2in
\label{fig:accuracy-nll}
\end{figure}

\subsection{Main Contribution}
\label{sec:introduction:contribution}

\input{resources/fig-time-nll}

Our main contribution is threefold:

\tcblack{\small{1}}
Spatially neighboring points in visual imagery tend to be similar, as do feature maps of convolutional neural networks (CNNs). By exploiting this spatial consistency, \emph{we propose spatial smoothing as a method of aggregating nearby feature maps} to improve the efficiency of ensemble size in BNN inference. 
The right side of \cref{fig:diagram} visualizes spatial smoothing averaging neighboring feature maps.

\tcblack{\small{2}}
We empirically demonstrate that spatial smoothing improves the ensemble efficiency in vision tasks, such as image classification on CIFAR and ImageNet datasets, without any additional training parameters. \Cref{fig:performance:cifar} shows that negative log-likelihood (NLL) of ``MC dropout + spatial smoothing'' with an ensemble size of two is comparable to that of vanilla MC dropout with an ensemble size of fifty. We also demonstrate that spatial smoothing improves accuracy, uncertainty, and robustness all at the same time. \Cref{fig:accuracy-nll} shows that spatial smoothing improves both the accuracy and uncertainty of various deterministic and Bayesian NNs with an ensemble size of fifty on CIFAR-100.

\tcblack{\small{3}}
Global average pooling (GAP) \citep{lin2013network,zhou2016learning}, pre-activation \citep{he2016identity}, and \texttt{ReLU6} \citep{krizhevsky2010convolutional,sandler2018mobilenetv2} have been widely used in vision tasks. However, their motives are largely justified by the experiments. 
We provide an explanation for these methods by addressing them as special cases of spatial smoothing. Experiments support the claim by showing that the methods improve not only accuracy but also uncertainty and robustness.

%% file: resources/fig-concept.tex
\begin{figure}
\centering
\vskip -0.13in
\includegraphics[width=0.99\columnwidth,page=6]{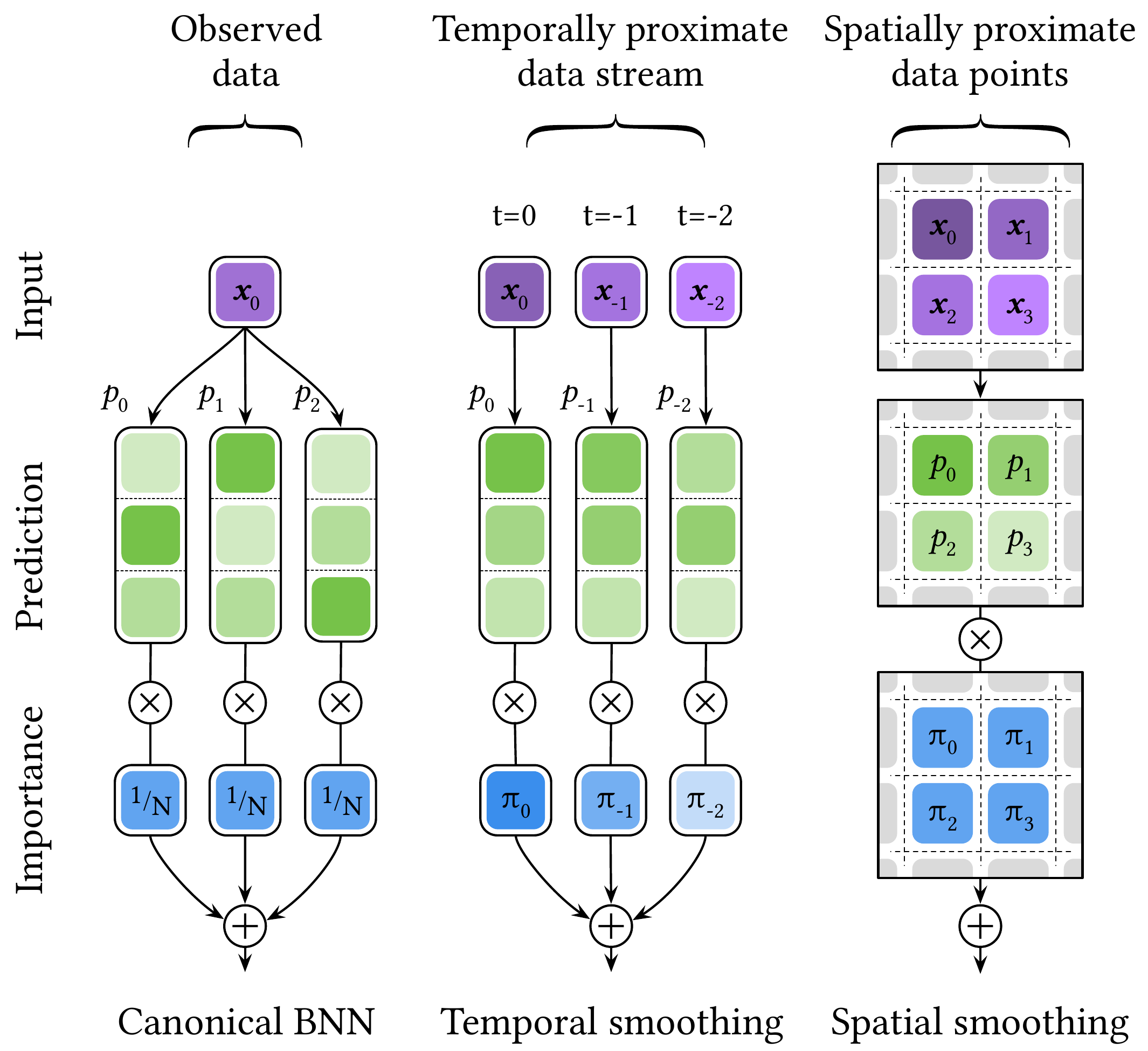}
\caption{
\textbf{Comparison of three different neural network ensembles: } canonical BNN inference, temporal smoothing \citep{park2019vqbnn}, and spatial smoothing (\emph{ours}). 
In this figure, $\bm{x}_0$ is observed data, $p_{i}$ is predictions $p(\bm{y} \vert \bm{x}_{0}, \bm{w}_{i})$ or $p(\bm{y} \vert \bm{x}_{i}, \bm{w}_{i})$, $\pi_i$ is importances $\pi(\bm{x}_{i} \vert \bm{x}_{0})$, and $N$ is ensemble size.
}
\vskip -0.33in

\label{fig:diagram}
\end{figure}

%% file: resources/fig-time-nll.tex
\begin{figure*}

\centering
\begin{subfigure}[b]{0.31\textwidth}
\centering
\includegraphics[width=\textwidth]{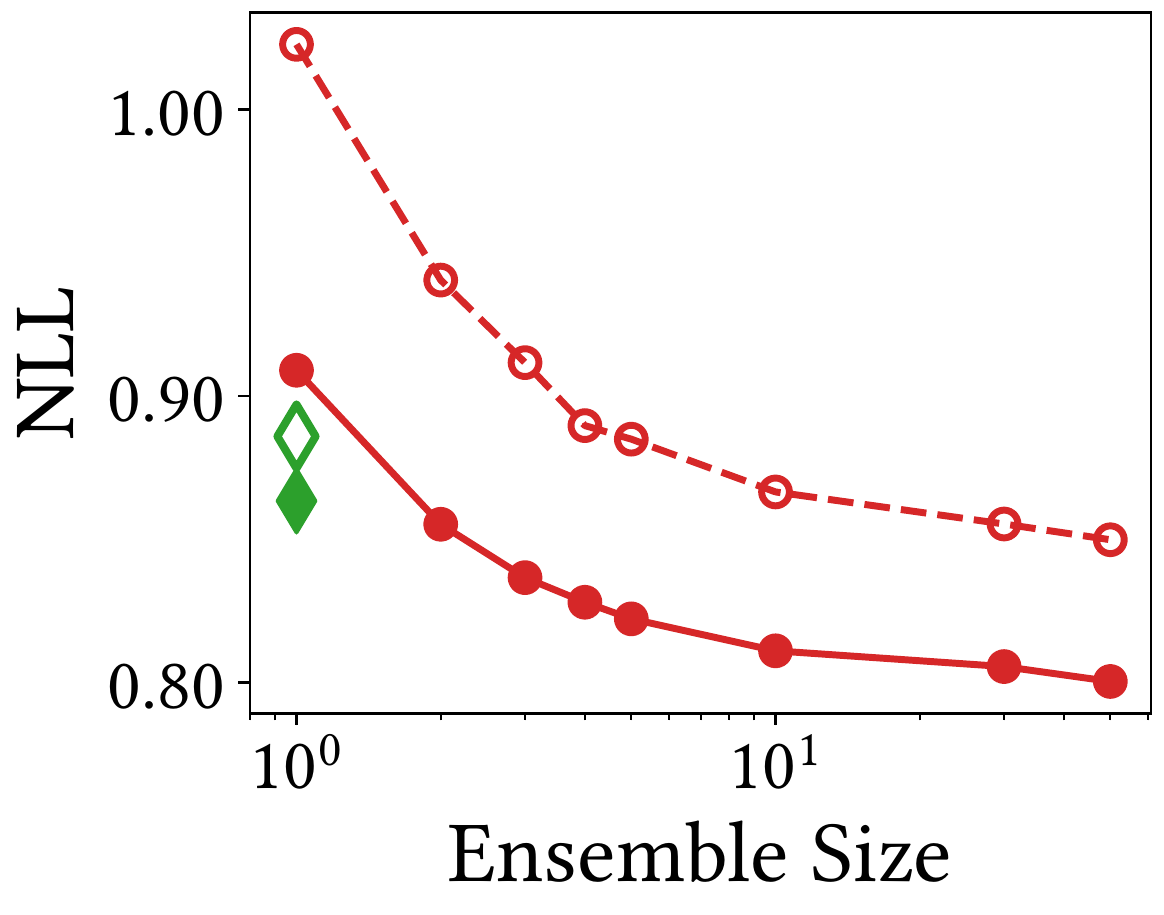}
\end{subfigure}
\hspace{0.5pt}
\begin{subfigure}[b]{0.31\textwidth}
\centering
\includegraphics[width=\textwidth]{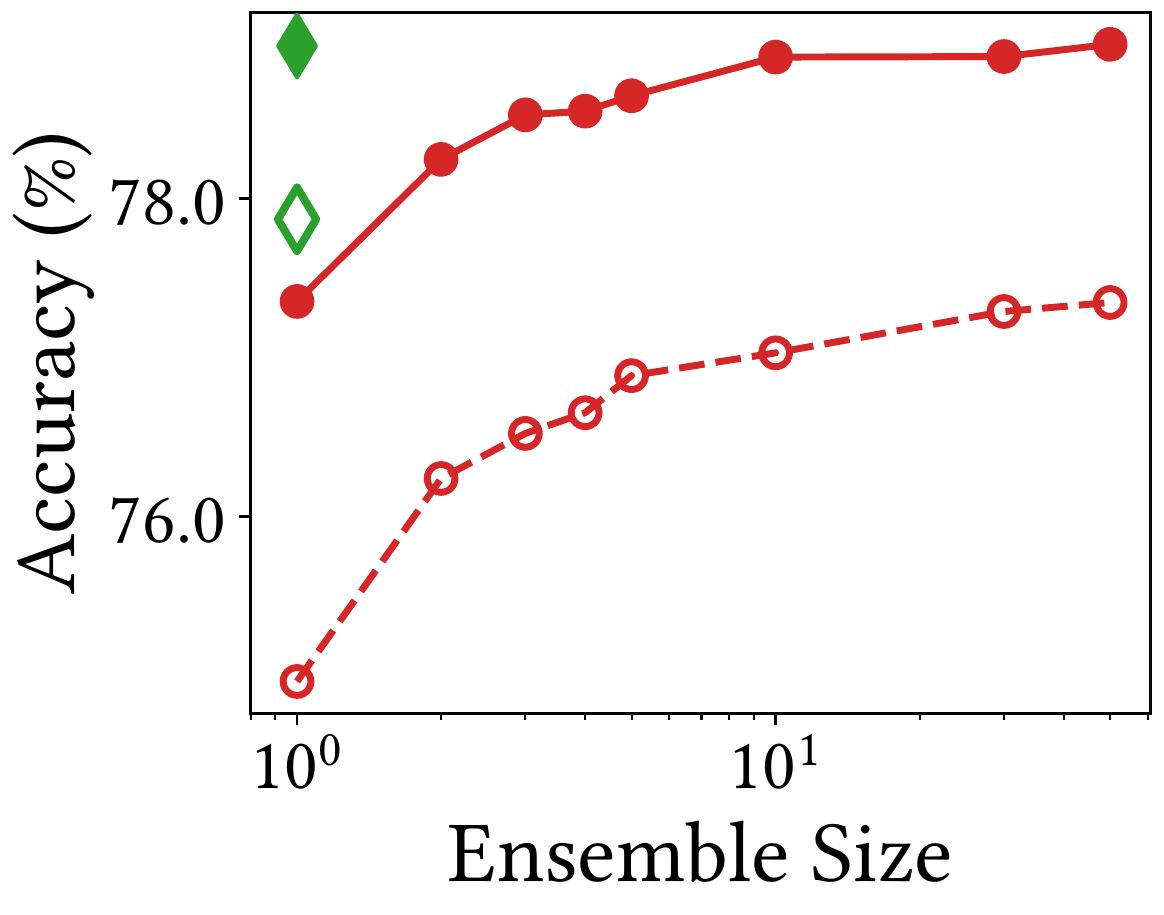}
\end{subfigure}
\hspace{0.5pt}
\centering
\begin{subfigure}[b]{0.31\textwidth}
\centering
\includegraphics[width=\textwidth]{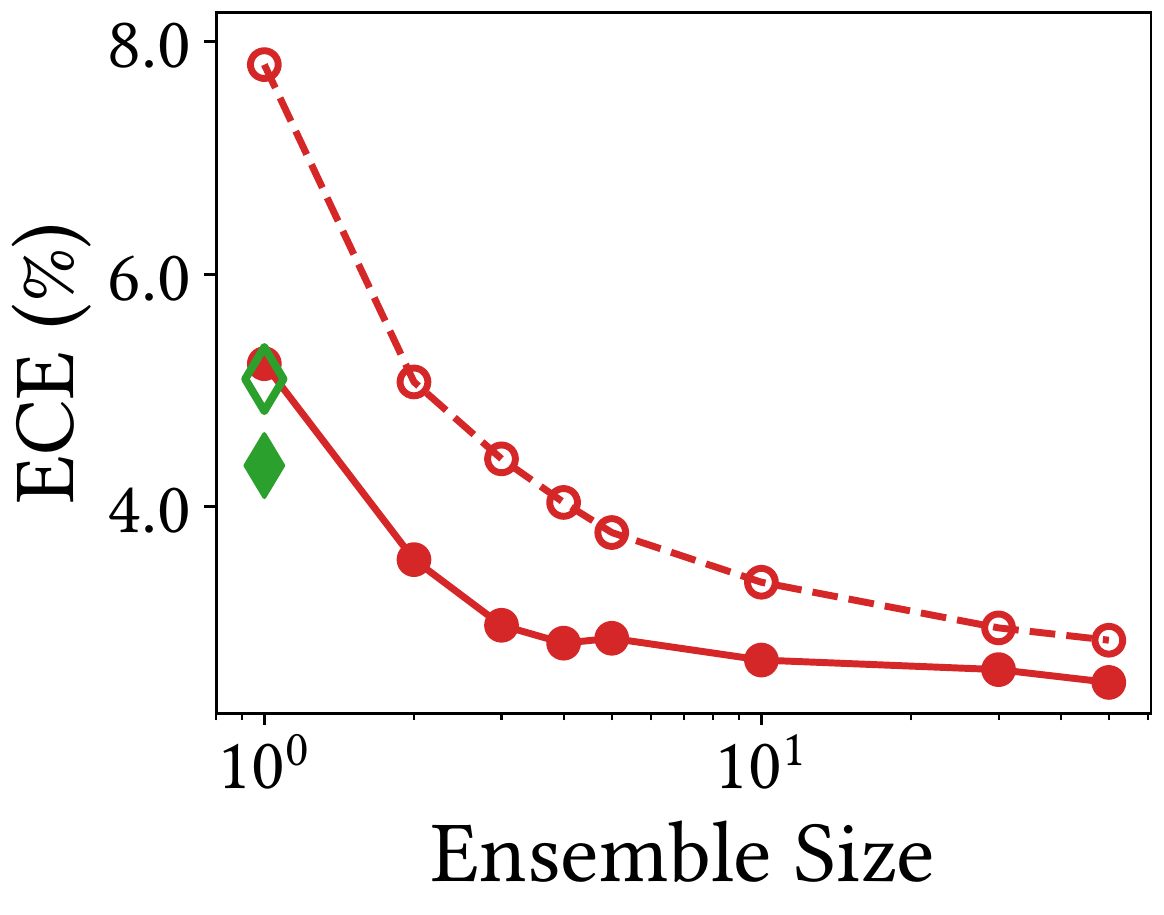}
\end{subfigure}
     
\vspace{3pt}
\centering
\includegraphics[height=0.028\textheight]{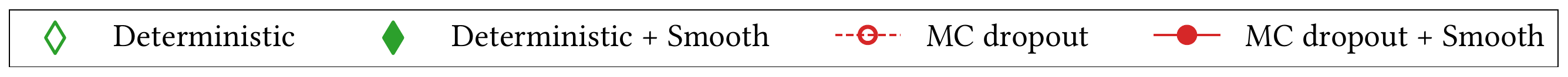}
 
\vskip -0.1in

\caption{
\textbf{Spatial smoothing improves both accuracy and uncertainty across a whole range of ensemble sizes.}
We report the predictive performance of ResNet-18 on CIFAR-100. See also \cref{fig:classification} for results on ImageNet.
}

\vskip -0.1in

\label{fig:performance:cifar}
\end{figure*}

%% file: body/method.tex
\section{Probabilistic Spatial Smoothing}\label{sec:spatial-smoothing}

To improve the computational performance of BNN inference, VQ-BNN \citep{park2019vqbnn} executes NN prediction only once and complements the result with previously calculated predictions as discussed in \cref{sec:introduction:prelim}. The key to the success of this approach largely depends on the collection of previous predictions for proximate data.
Gathering temporally proximate data and their predictions from data streams is easy because recent data and predictions can be aggregated using temporal consistency. On the other hand, gathering time-independent proximate data, e.g. images, is more difficult because they lack such consistency.

\subsection{Module Architecture for Ensembling Neighboring Feature Map Points}\label{sec:spatial-smoothing:architecture}

\input{resources/fig-diagram-block}

So instead of temporal consistency, we use spatial consistency---where neighboring pixels of images are similar---for real-world images. 
Under this assumption, we take the feature maps as predictions and aggregate neighboring feature maps. 

Most CNN architectures, including ResNet, consist of multiple stages that begin with increasing the number of channels while reducing the spatial dimension of the input volume. We decompose an entire BNN inference into several steps by rewriting each stage in a recurrence relation as follows:
\begin{align}
	p(\bm{z}_{i+1} \vert \bm{z}_{i}, \mathcal{D}) = \int p(\bm{z}_{i + 1} \vert \bm{z}_i, \bm{w}_i) \, p(\bm{w}_i \vert \mathcal{D}) \, d\bm{w}_i
	\label{eq:stage-inference}
\end{align}
where $\bm{z}_i$ is input volume of the $i$-th stage, and the first and the last volume are input data and output. $\bm{w}_i$ and $p(\bm{w}_i \vert \mathcal{D})$ are NN weight in the $i$-th stage and its probability. $p(\bm{z}_{i+1} \vert \bm{z}_{i}, \bm{w}_i)$ is output probability of $\bm{z}_{i+1}$ with respect to the input volume $\bm{z}_i$.
To derive the probability from the output feature map, we transform each point of the feature map into a Bernoulli distribution. 
To do so, a composition of \texttt{tanh} and \texttt{ReLU}, a function from value of range $[ -\infty, \infty ]$ into probability, is added after each stage.
Put shortly, we use neural networks for \emph{point-wise binary feature classification}.

\input{resources/fig-feature-variance}

Since \cref{eq:stage-inference} is a kind of BNN inference, it can be approximated using \cref{eq:approx-vqbnn-inference}. In other words, each stage predicts feature map points only once and complements predictions with similar but slightly different feature maps.
Under spatial consistency, it averages probabilities of spatially neighboring feature map points, which is well known as \emph{blur} operation in image processing.
For the sake of implementation simplicity, average pooling with a kernel size of 2 and a stride of 1 is used as a box blur. This operation aggregates four neighboring probabilities with the same importances.

In summary, as shown in \cref{fig:spatial-smoothing}, we propose the following \emph{probabilistic spatial smoothing} layer:
\begin{equation}
	\texttt{Smooth}(\bm{z}) = \texttt{Blur} \circ \texttt{Prob} \, (\bm{z})
\end{equation}
where $\texttt{Prob} (\cdot)$ is a point-wise function from a feature map to probability, and $\texttt{Blur} (\cdot)$ is importance-weighted average for aggregating spatially neighboring probabilities from feature maps. This \texttt{Smooth} layer is added before each downsampling layers, so we use four \texttt{Smooth} layers for ResNet. \texttt{Prob} and \texttt{Blur} are further elaborated below.

\input{resources/fig-fourier.tex}

\paragraph{\texttt{Prob}: Feature map to probability. }

\texttt{Prob} is a function that transforms a real-valued feature map into probability. We use  \texttt{tanh}--\texttt{ReLU} composition for this purpose. However, \texttt{tanh} is commonly known to suffer from the vanishing gradient problem. To alleviate this issue, we propose the following temperature-scaled \texttt{tanh}:
\begin{align}
	\texttt{tanh}_{\tau} (\bm{z}) = \tau \, \texttt{tanh} \, (\bm{z} / \tau)
\end{align}
where $\tau$ is a hyperparameter called temperature. $\tau$ is $1$ in conventional \texttt{tanh} and $\infty$ in identity function. 
$\texttt{tanh}_{\tau}$ imposes an upper bound on a value, but does not limit the upper bound to 1.

An unnormalized probability, ranging from $0$ to $\tau$, is allowed as the output of \texttt{Prob}. Then, thanks to the linearity of integration, we obtain an unnormalized predictive distribution accordingly. Taking this into account, we propose the following \texttt{Prob}:
\begin{align}
	\texttt{Prob}(\bm{z}) = \texttt{ReLU} \circ \, \texttt{tanh}_{\tau} (\bm{z})
	\label{eq:prob}
\end{align}
where $\tau > 1$. 
We empirically determine $\tau$ to minimize NLL, a metric that measures both accuracy and uncertainty. 

We expect other upper-bounded functions, such as $\texttt{ReLU6}(\bm{z}) = \texttt{ReLU} \circ \texttt{min}(\bm{z}, 6)$ and feature map scaling $\bm{z} / \tau$ with $\tau > 1$ which is \texttt{BatchNorm}, to be able to replace $\texttt{tanh}_{\tau}$ in \texttt{Prob}; and as expected, these alternatives improve uncertainty estimation in addition to accuracy. See \cref{sec:revisiting:preact} and \cref{sec:revisiting:relu6} for detailed discussions on activation ($\texttt{ReLU} \circ \texttt{BatchNorm}$) and \texttt{ReLU6} as \texttt{Prob}.

\paragraph{\texttt{Blur}: Averaging neighboring probabilities. }

\texttt{Blur} averages the probabilities from feature maps.
We primarily use the average pool with a kernel size of 2 and a stride of 1 as the implementation of \texttt{Blur} for the sake of simplicity.
Nevertheless, we could generalize \texttt{Blur} by using the following depth-wise convolution, which acts on each input channel separately, with non-trainable kernel
\begin{equation}
	\bm{K} = \frac{1}{\vert\vert \bm{k} \vert\vert_{1}^{2}} \, \bm{k} \otimes \bm{k}^{\top}
	\label{eq:blur}
\end{equation}
where $\bm{k}$ is a one-dimentional matrix, such as $\bm{k} \in \{ \left( 1 \right),  \left( 1, 1 \right), \left( 1, 2, 1 \right), \left( 1, 4, 6, 4, 1 \right), \cdots \}$. Different $\bm{k}$s derive different importances for neighboring feature maps. 
We experimentally show that most \texttt{Blur}s improve the predictive performance. The optimal $\bm{K}$ varies by model, suggesting that the experimental results in this paper have a potential for improvement.

\subsection{How Does Spatial Smoothing Help Optimization?}\label{sec:spatial-smoothing:how}

We demonstrate that spatial smoothing has the key properties of ensembles: it reduces feature map variances, filters high-frequency signals, and smoothens loss landscapes. In addition to the improved robustness against MC dropout, which randomly deletes spatial information (cf. \citet{veit2016residual}), these empirical perspectives suggest that \emph{spatial smoothing behaves like ensembles}. 
Since these properties are the positive attributes that can be expected from ensembles, spatial smoothing can be regarded as ensembles.

\paragraph{Feature map variance. }

BNNs have two types of uncertainties: One is model uncertainty and the other is data uncertainty \citep{park2019vqbnn}, the distribution of feature map points. Such randomness increases the variance of feature maps. To show that spatial smoothing aggregates the feature maps, we use the following proposition:
\begin{prop}\label{thm:variance}
Ensembles reduce the variance of predictions.
\end{prop}
Proof is omitted since it is straightforward. In our context, predictions are output feature map points of a stage.
We investigate model and data uncertainties of the predictions along NN layers to show that spatial smoothing reduces randomnesses and ensembles feature maps. \Cref{fig:fm-variance} shows the model uncertainty and data uncertainty of Bayesian ResNets including MC dropout layers. 
In this figure, the uncertainty of MC dropout's feature map only accumulates, and almost monotonically increases in every NN layer. In contrast, the uncertainty of the feature map of ``MC dropout + spatial smoothing'' significantly decreases in the spatial smoothing layers, suggesting that the smoothing layers ensemble the feature map. In other words, they make the feature map more accurate and stabilized input volumes for the next stages.
Deterministic NNs do not have model uncertainty but data uncertainty. Therefore, spatial smoothing improves the performance of deterministic NNs as well as Bayesian NNs.

\input{resources/fig-loss-landscape}

\paragraph{Fourier analysis. }

We also analyze spatial smoothing through the lens of Fourier transform: 
\begin{prop}\label{thm:frequency}
Ensembles filter high-frequency signals.
\end{prop}
Proof is provided in \cref{sec:extended-analysis:fourier}.
\Cref{fig:fourier:amplitude} shows the two-dimensional Fourier transformed output feature map at the end of the stage 1. It reveals that MC dropout has almost no effect on the low-frequency ($< 0.3 \pi$) ranges, but adds high-frequency ($\geq 0.3 \pi$) noises. Since spatial smoothing is a low-pass filter, it effectively filters high-frequency signals, including the noises caused by MC dropout. 

We also find that CNNs are particularly vulnerable to high-frequency noises. To demonstrate this claim, following \citet{shao2021adversarial}, we measure accuracy with respect to data with frequency-based random noise $\bm{x}_{\text{noise}} = \bm{x}_0 + \mathcal{F}^{-1}\left( \mathcal{F} (\delta) \odot \textbf{M}_f \right)$, where $\bm{x}_0$ is clean data, $\mathcal{F}(\cdot)$ and $\mathcal{F}^{-1}(\cdot)$ are Fourier transform and inverse Fourier transform, $\delta$ is Gaussian random noise, and $\textbf{M}_f$ is frequency mask as shown in \cref{fig:fourier:mask}. \Cref{fig:fourier:robustness} exhibits the results. In sum, the results show that high-frequency noises significantly impair accuracy. Spatial smoothing improves the robustness by effectively removing high-frequency noises, including those caused by MC dropout.

\paragraph{Loss landscape. }

Lastly, we show that randomness hinders NN training, and ensembles help optimization:
\begin{prop}\label{thm:loss-landscape}
The randomness of predictions sharpens the loss landscapes, and ensembles flatten them.
\end{prop}
Proof is provided in \cref{sec:extended-analysis:dropout}.
Since a sharp loss function disturbs NN optimization \citep{keskar2016large,santurkar2018does,foret2020sharpness}, reducing the randomness helps NNs learn strong representations. Ensembles with multiple NN predictions in training phases flatten the loss function by averaging out the randomness.
Consequently, \emph{an ensemble of BNN outputs in training phases significantly improves the predictive performance}. See \cref{fig:training-phase-ensemble} for numerical results. However, we do not use this training phase ensemble because it significantly increases training time. We use spatial smoothing instead since it ensembles feature map points without adding training time.

We visualizes the loss landscapes \citep{li2017visualizing}, i.e., the contours of NLL on training datasets. \Cref{fig:loss-landscape:do} 
shows that the loss landscapes of MC dropout fluctuate and have irregular surfaces due to randomness. As \citet{li2017visualizing,foret2020sharpness} pointed out, this may lead to poor generalization and predictive performance. Spatial smoothing reduces randomness, as discussed above, and \emph{spatial smoothing aids optimization by stabilizing and flattening the loss landscapes of BNNs}, as shown in \cref{fig:loss-landscape:dosm}.

\begin{figure}[ht]
\centering

\vskip 0.12in

\raisebox{0pt}[\dimexpr\height-0.6\baselineskip\relax]{
\includegraphics[width=0.31\textwidth]{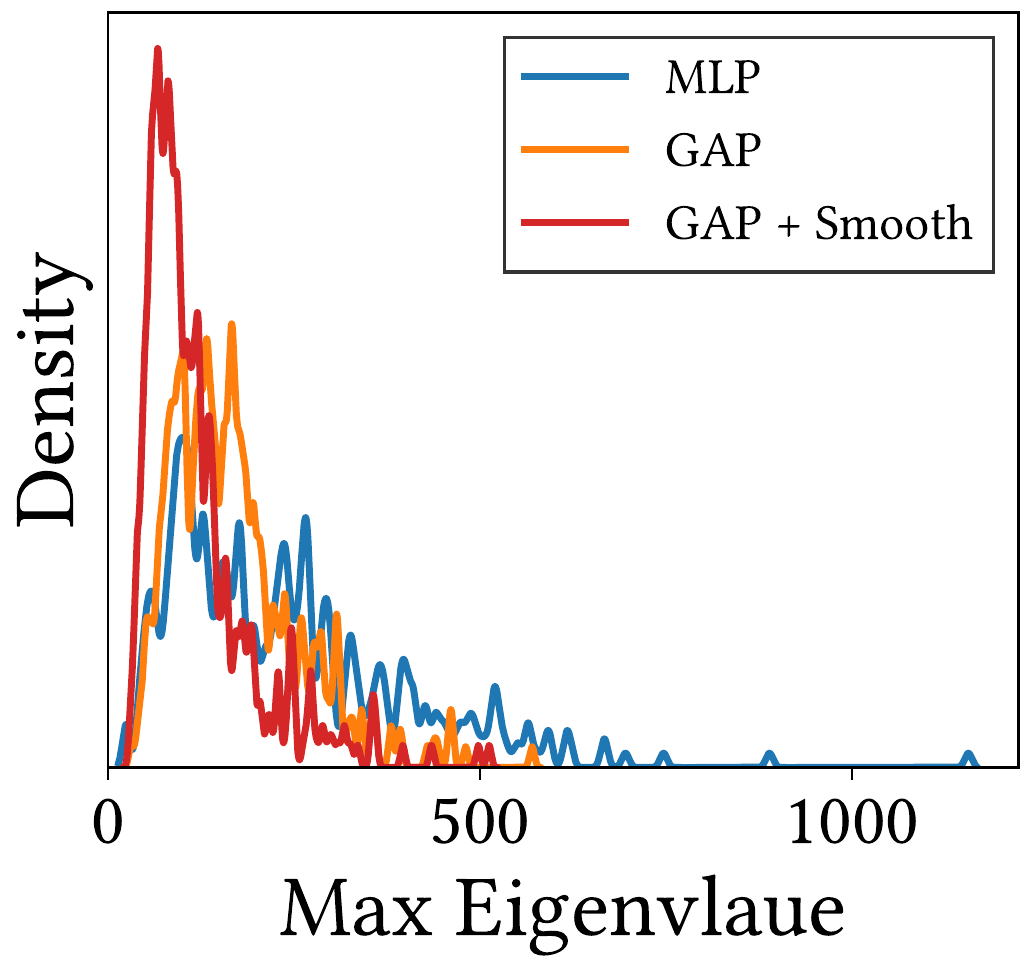}
}

\caption{
\textbf{Both GAP and spatial smoothing suppress large Hessian eigenvalue outliers}, i.e., they flatten the loss landscapes. 
Compare with \cref{fig:loss-landscape}.
}
\label{fig:loss-landscape:hmes}

\vskip -0.15in

\end{figure}

Furthermore, we use Hessian to quantitatively represent the sharpness of loss landscapes. 
The larger the Hessian eigenvalue, the sharper the loss landscape. 
To efficiently investigate the Hessian eigenvalues of the models in \cref{fig:loss-landscape}, we propose \emph{``Hessian max eigenvalue spectrum''}. A detailed description of the Hessian max eigenvalue spectra method is provided in \cref{sec:hmes}.
The results of the experiments with a batch size of 128 are provided in \cref{fig:loss-landscape:hmes}, which consistently show that spatial smoothing reduces the magnitude of Hessian eigenvalues and suppresses outliers. Since large Hessian eigenvalues disturb NN training \citep{ghorbani2019investigation}, we arrive at the same conclusion that spatial smoothing helps NN optimization. 
In addition, we propose the conjecture that the flatter the loss landscape, the better the uncertainty estimation, and vice versa.

\subsection{Revisiting Global Average Pooling}\label{sec:spatial-smoothing:gap}

\begin{table}[ht]
\centering
\vskip 0.0in
\caption{%
\textbf{MLP does not overfit the training dataset.} 
We report training NLL ($\text{NLL}_\text{train}$) and testing NLL ($\text{NLL}_\text{test}$) of ResNet-50 on CIFAR-100. 
}\label{tab:classifier-represent}
\vskip -0.2in
\begin{tabular}{ccc}\\\toprule  
\textsc{Classifier} & $\text{NLL}_\text{train}$ & $\text{NLL}_\text{test}$\\\midrule
GAP & \textbf{0.0061} & \textbf{0.822} \\  \midrule
MLP & 0.0071 & 1.029 \\  \bottomrule
\end{tabular} 
\vskip -0.02in
\end{table}

The success of GAP classifier in image classification is indisputable. 
The initial motivation and the most widely accepted explanation for this success is that GAP prevents overfitting by using far fewer parameters than multi-layer perceptron (MLP) \citep{lin2013network}. 
However, we discover that the explanation is poorly supported. We compares GAP with other classifiers including MLP. Contrary to popular belief, \cref{tab:classifier-represent} suggests that \emph{MLP does not overfit  the training dataset}. 
MLP underfits or gives comparable performance to GAP on the training dataset.
On the test dataset, GAP provides better results compared with MLP. 
See \cref{tab:classifier} for more detailed results.

Our argument is that GAP is an extreme case of spatial smoothing. In other words, GAP is successful because it aggregates feature map points and smoothens the loss landscape to help optimization. 
To support this claim, we visualizes the loss landscape of MLP as shown in \cref{fig:loss-landscape:mlp}. It is chaotic compared to that of GAP as shown in \cref{fig:loss-landscape:do}.
In conclusion, \emph{averaging feature maps tends to help neural networks learn strong representations}. 
Hessian shows the consistent results as demonstrated by \cref{fig:loss-landscape:hmes}.

%% file: resources/fig-diagram-block.tex
\begin{figure}[ht]
\centering

\raisebox{0pt}[\dimexpr\height-0.6\baselineskip\relax]{
\includegraphics[height=0.17\textheight,page=2]{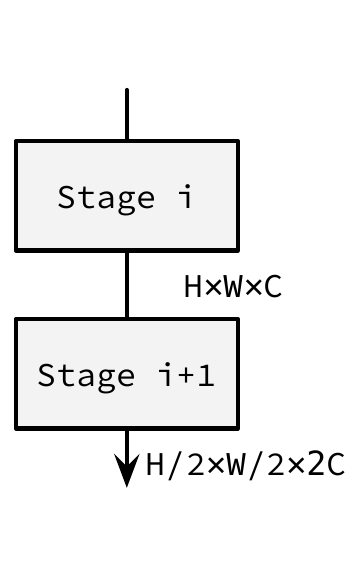}
\hspace{1pt}
\includegraphics[height=0.17\textheight,page=3]{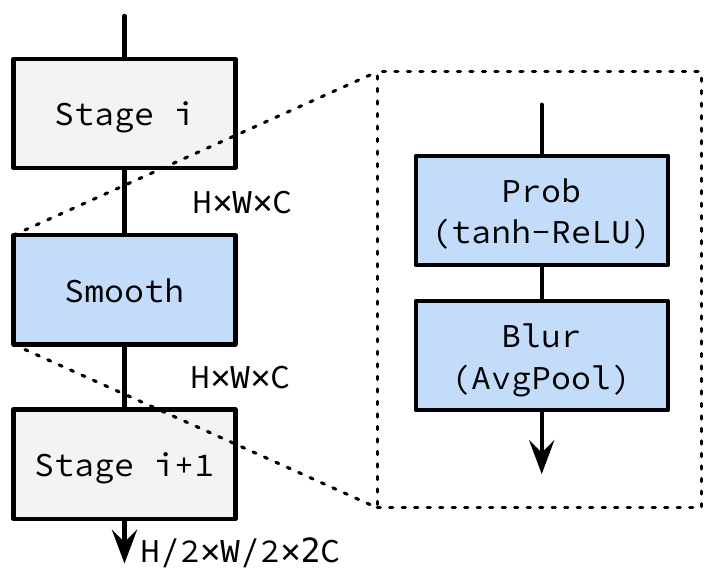}
}

\caption{
\textbf{Stages of CNNs such as ResNet (\emph{left}) and the stages incorporating spatial smoothing layer (\emph{right})}.
}
\label{fig:spatial-smoothing}
\end{figure}

%% file: resources/fig-feature-variance.tex
\begin{figure*}
\vskip 0.2in
\begin{center}

\begin{subfigure}[b]{0.44\textwidth}
\centering
\includegraphics[width=0.9\textwidth]{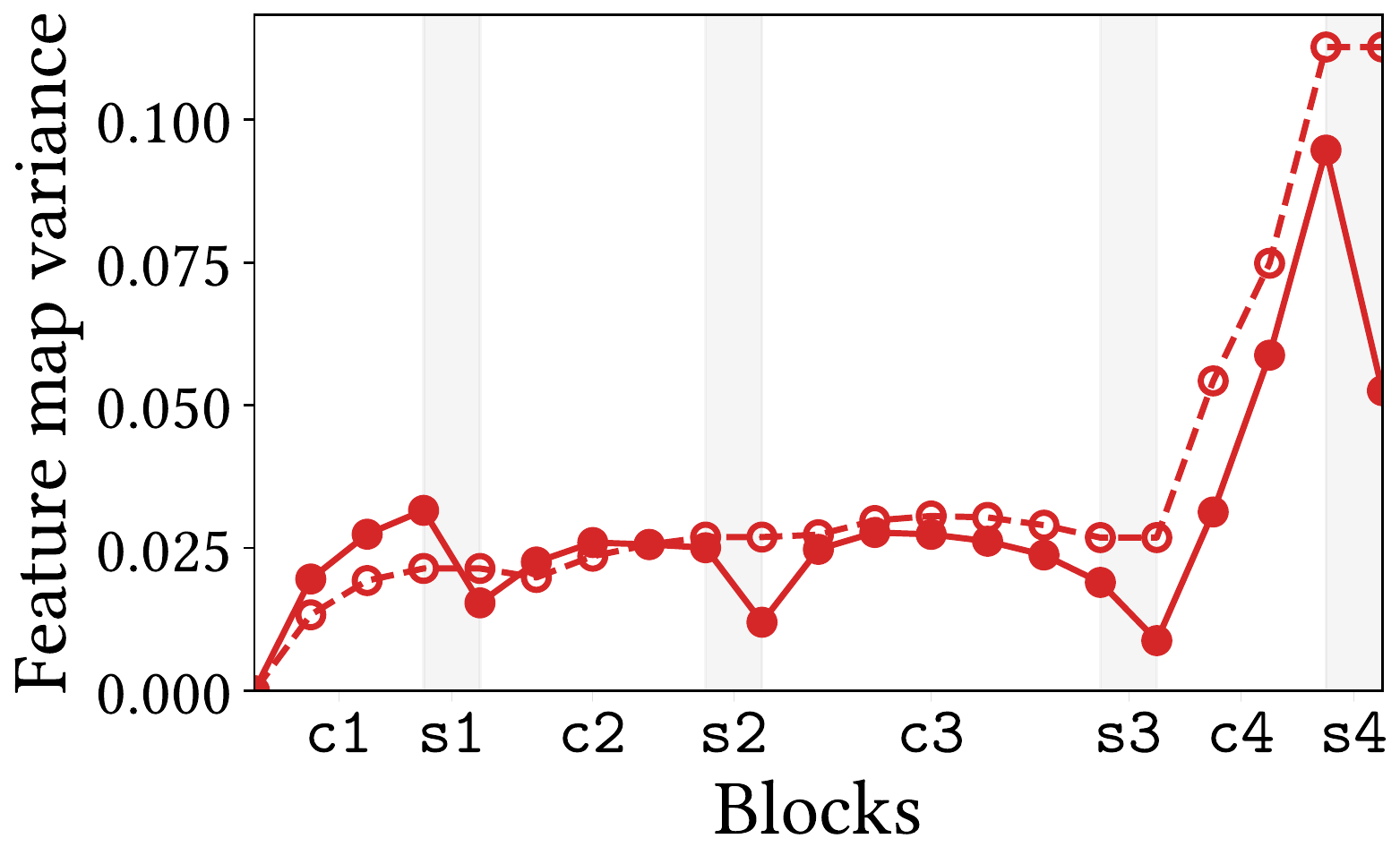}
\caption{Model uncertainty}
\end{subfigure}
\hspace{4pt}
\begin{subfigure}[b]{0.44\textwidth}
\centering
\includegraphics[width=0.9\columnwidth]{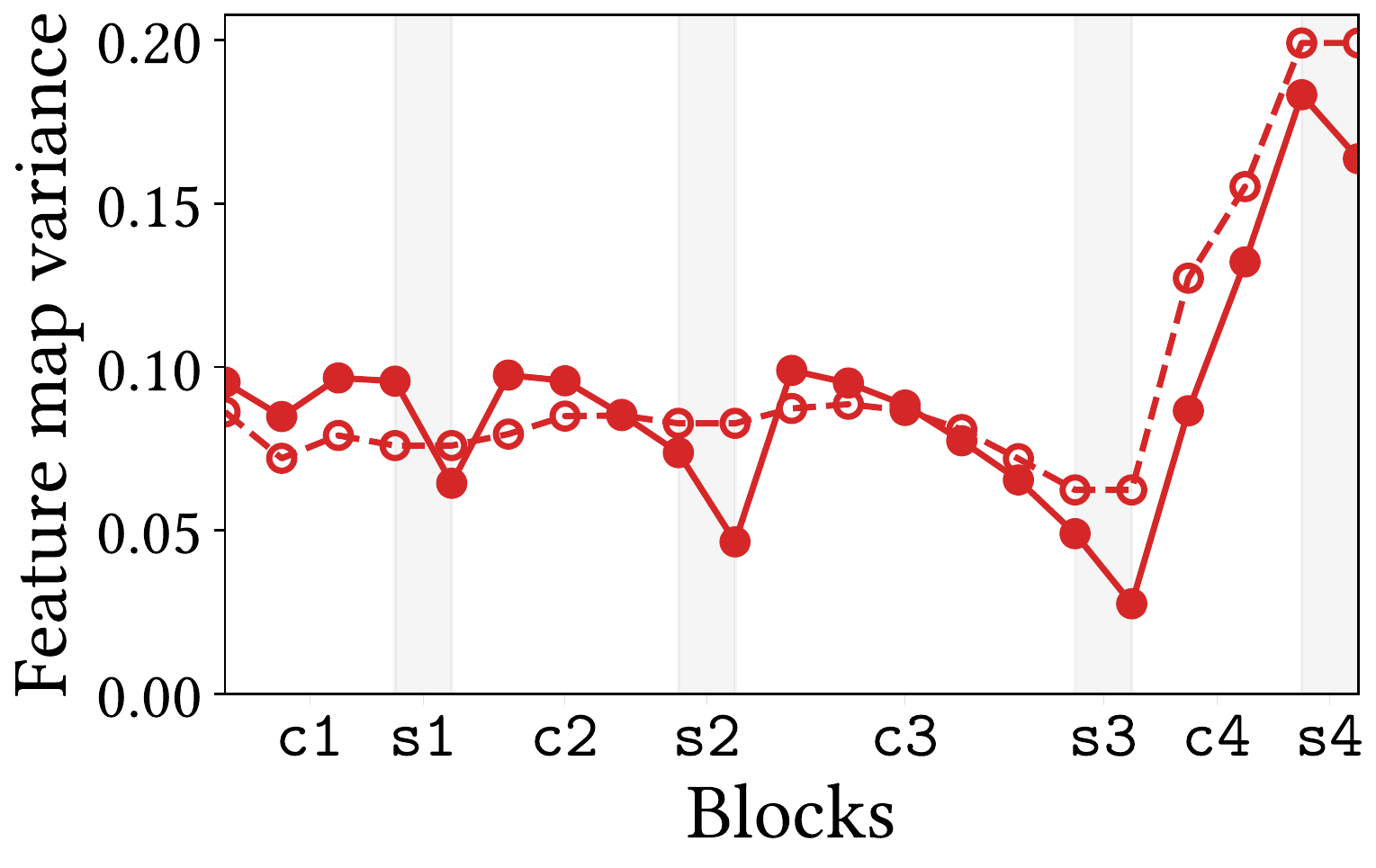}
\caption{
Data uncertainty
}
\end{subfigure}

\vspace{3pt}
\centering
\includegraphics[height=0.028\textheight]{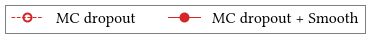}

\vspace{-3pt}
\caption{
\textbf{Spatial smoothings (gray area) reduce feature map variances}, suggesting that they ensemble feature map points.
We provide standard deviations of feature maps by block depth with ResNet-50 on CIFAR-100. $\texttt{c1}$ to $\texttt{c4}$ and $\texttt{s1}$ to $\texttt{s4}$ each stand for stages and spatial smoothing layers. The standard deviations are averaged over the channels. 
\emph{Left:} Model uncertainty is represented by the average standard deviation of several feature maps obtained from multiple NN executions.
\emph{Right:} Data uncertainty is represented by the standard deviation of feature map points obtained from one NN execution.
}
\label{fig:fm-variance}

\end{center}

\vskip -0.15in 
\end{figure*}

%% file: resources/fig-fourier.tex
\begin{figure*}
\centering
\begin{subfigure}[b]{0.252\textwidth}
\centering

\vskip 0.1in

\includegraphics[width=0.95\textwidth,page=4]{resources/diagrams.pdf}
\vspace{14pt}
\caption{Frequency mask}
\label{fig:fourier:mask}
\end{subfigure}
\hspace{10pt}
\centering
\begin{subfigure}[b]{0.325\textwidth}
\centering
\includegraphics[width=0.95\textwidth]{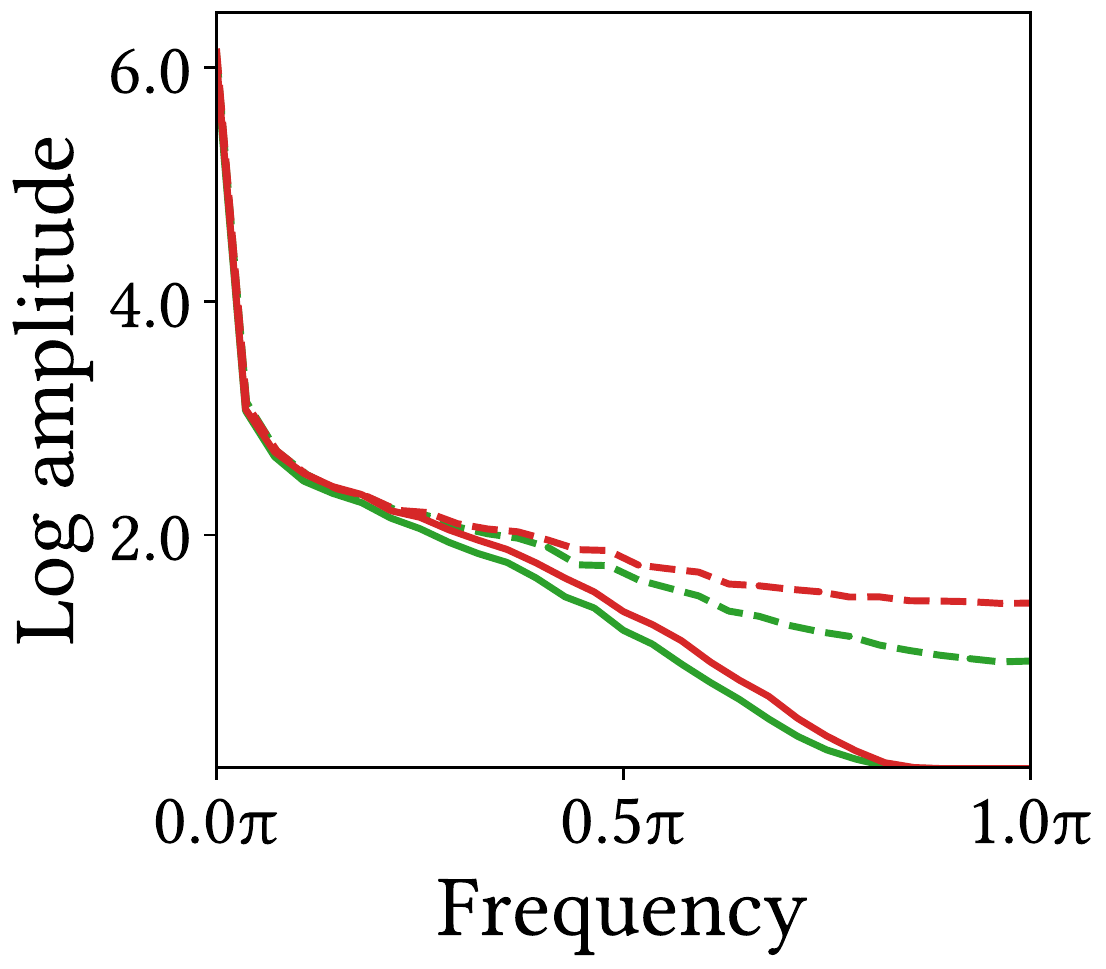}
\caption{Log amplitude at \texttt{c1}}
\label{fig:fourier:amplitude}
\end{subfigure}
\centering
\begin{subfigure}[b]{0.303\textwidth}
\centering
\includegraphics[width=0.95\textwidth]{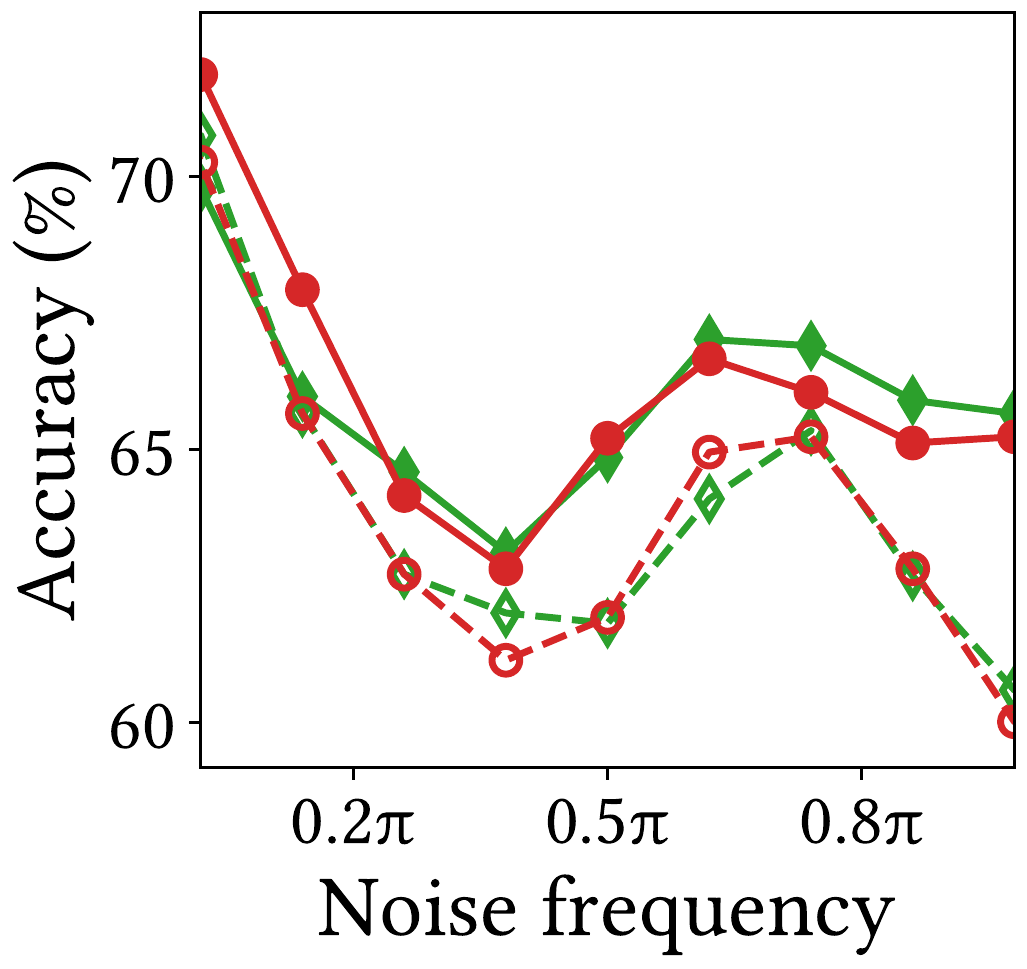}
\caption{Robustness for noise frequency}
\label{fig:fourier:robustness}
\end{subfigure}
     
\vspace{3pt}
\centering
\includegraphics[height=0.026\textheight]{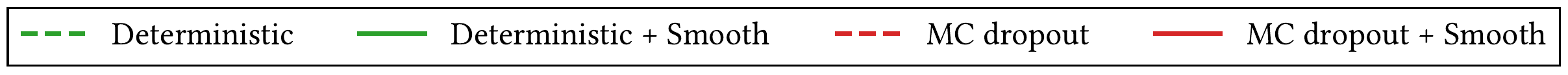}

\vspace{-2pt}
\caption{
\textbf{MC dropout adds high-frequency noises, and spatial smoothing filters high-frequency signals.}
In these experiments, we use ResNet-50 for ImageNet.
\emph{Left: } 
Frequency mask $\textbf{M}_f$ with $w = 0.1 \pi$. 
\emph{Middle: }
Diagonal components of Fourier transformed feature maps at the end of the stage 1. 
\emph{Right: }
The accuracy against frequency-based random noise. 
ResNets are vulnerable to high-frequency noises. Spatial smoothing improves the robustness against high-frequency noises. 
}
\label{fig:fourier}

\vskip -0.05in

\end{figure*}

%% file: resources/fig-loss-landscape.tex
\begin{figure*}

\centering

\vskip 0.1in

\begin{subfigure}[b]{0.28\textwidth}
\centering
\includegraphics[width=\textwidth]{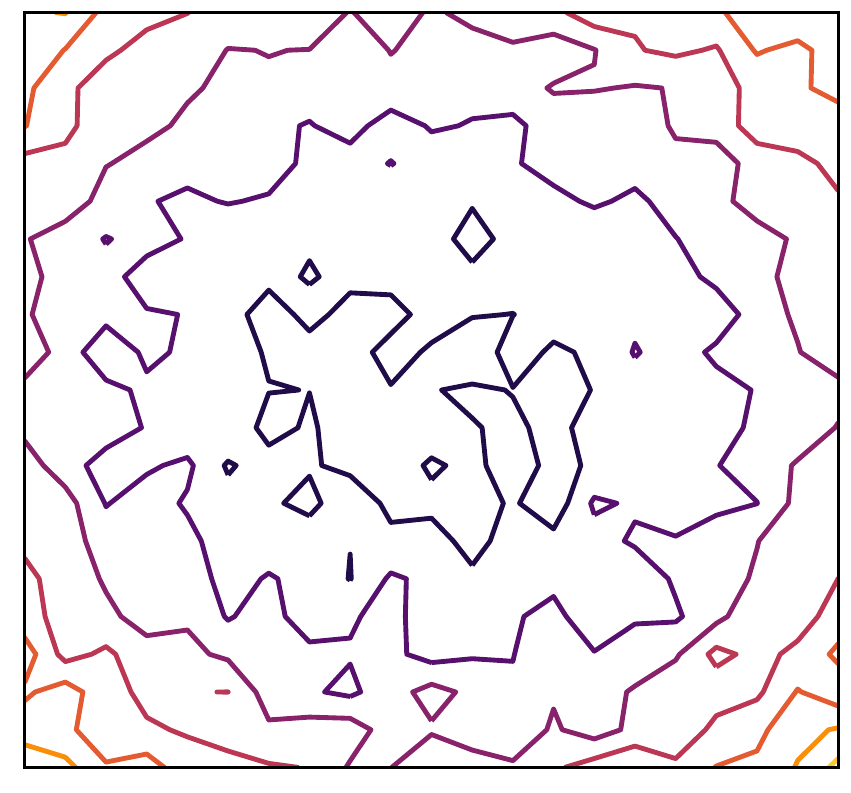}
\caption{MLP classifier}
\label{fig:loss-landscape:mlp}
\end{subfigure}
\hspace{1pt}
\centering
\begin{subfigure}[b]{0.28\textwidth}
\centering
\includegraphics[width=\textwidth]{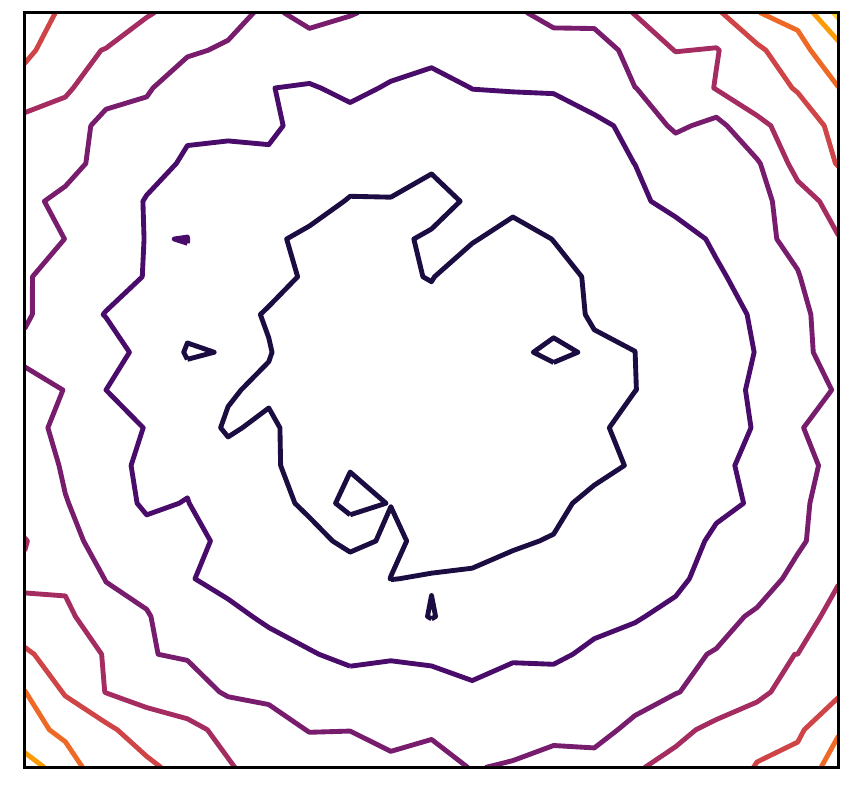}
\caption{GAP classifier (\emph{vanilla})}
\label{fig:loss-landscape:do}
\end{subfigure}
\hspace{1pt}
\centering
\begin{subfigure}[b]{0.28\textwidth}
\centering
\includegraphics[width=\textwidth]{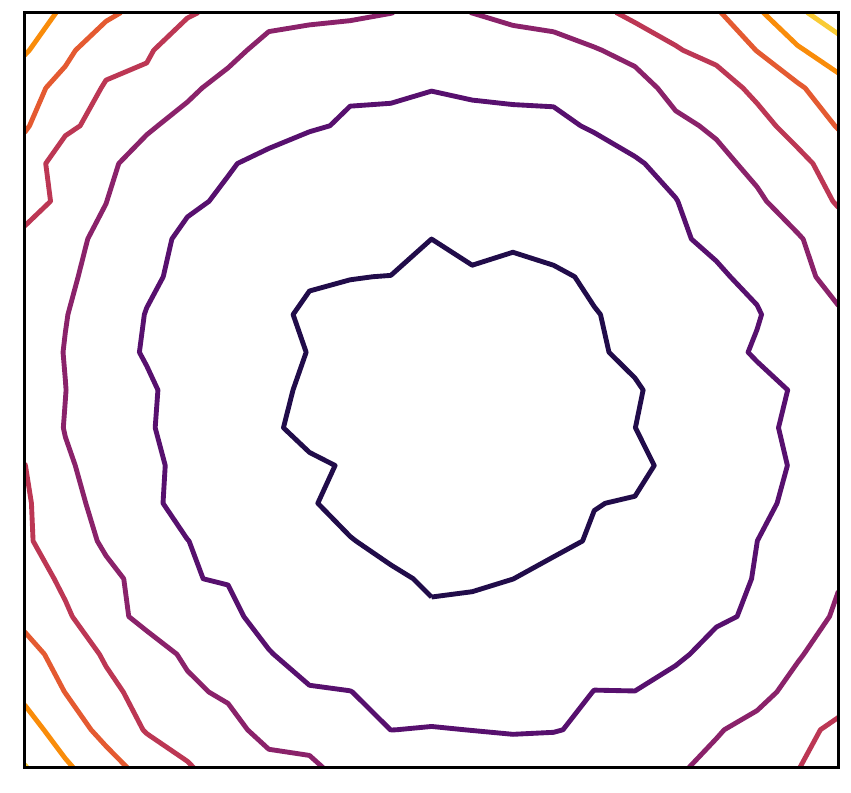}
\caption{GAP classifier + \texttt{Smooth}}
\label{fig:loss-landscape:dosm}
\end{subfigure}

\vskip -0.05in

\caption{
\textbf{Both GAP and spatial smoothing smoothen the loss landscapes.} To demonstrate this, we present the loss landscape visualizations of ResNet-18 models with MC dropout on CIFAR-100. 
}
\label{fig:loss-landscape}

\vskip -0.10in

\end{figure*}

%% file: body/experiment.tex
\section{Experiments}
\label{sec:experiments}

This section presents two experiments. 
The first experiment is image classification through which we show that spatial smoothing not only improves the ensemble efficiency, but also the accuracy, uncertainty, and robustness of both deterministic NN and MC dropout.
The second experiment is semantic segmentation on data streams through which we show that spatial smoothing and temporal smoothing \citep{park2019vqbnn} are complementary. In all experiments, we report the average of three evaluations, and the standard deviations are significantly smaller than the improvements. See \cref{sec:conf} for more detailed configurations.

Three metrics are measured in these experiments: NLL ($\downarrow$\footnote{The arrows indicate which direction is better.}), accuracy ($\uparrow$), and expected calibration error (ECE, $\downarrow$) \citep{guo2017calibration}. NLL represents both accuracy and uncertainty, and is the most widely used as a proper scoring rule. ECE measures discrepancy between accuracy and confidence.

\subsection{Image Classification}
\label{sec:experiments:classification}

We mainly discuss ResNet \citep{he2016deep} in image classification, but various models---e.g., VGG \citep{simonyan2014very}, ResNeXt \citep{xie2017aggregated}, and pre-activation models \citep{he2016deep}---on various datasets---e.g., CIFAR-\{10, 100\} and ImageNet---show the same trend as shown in \cref{tab:classification}.
Spatial smoothing also improves deep ensemble \citep{lakshminarayanan2017simple}, another non-Bayesian probabilistic NN method, as shown in \cref{fig:classification}.

\paragraph{Performance. }

\Cref{fig:performance:cifar} shows the predictive performances of ResNet-18 on CIFAR-100. The results indicate that \emph{spatial smoothing improves both accuracy and uncertainty} in many respects. Let us be more specific. First, spatial smoothing improves the efficiency of ensemble size. In these examples, the NLL of ``MC dropout + spatial smoothing'' with an ensemble size of 2 is comparable to or even better than that of MC dropout with an ensemble size of 50. In other words, ``MC dropout + spatial smoothing'' is 25$\times$ faster than MC dropout with a similar predictive performance. 
Second, the predictive performance of ``MC dropout + spatial smoothing'' is better than that of MC dropout, at an ensemble size of 50. As discussed in \cref{thm:loss-landscape}, flat loss landscapes in training phase lead to better performance.
Third, spatial smoothing improves the predictive performance of deterministic NN, as well as MC dropout.

\paragraph{Robustness. }

\input{resources/tab-semantic-segmentation-sim}

To evaluate robustness against data corruption, we measure predictive performance of ResNet-18 on CIFAR-100-C \citep{hendrycks2019benchmarking}. This dataset consists of data corrupted by 15 different types, each with 5 levels of intensity each.
We use mean corruption NLL (mCNLL, $\downarrow$), the averages of NLL over intensities and corruption types, to summarize the performance of corrupted data in a single value.
See \cref{eq:cnll} for a rigorous definition.

\begin{figure}[ht]
\centering

\vskip 0.05in

\raisebox{0pt}[\dimexpr\height-0.6\baselineskip\relax]{
\includegraphics[width=0.30\textwidth]{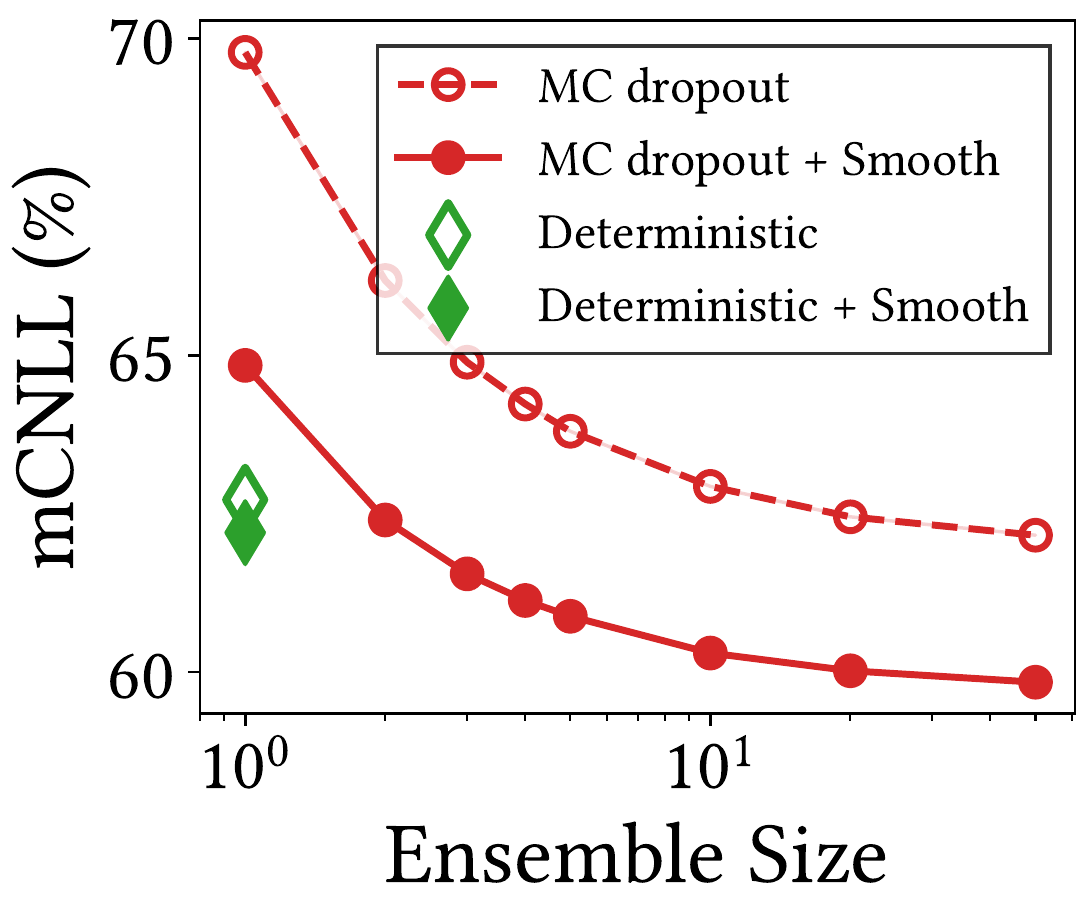}
}

\vspace{-2pt}

\caption{
\textbf{Spatial smoothing improves corruption robustness.}
We report mCNLL of ResNet-18 on CIFAR-100-C.
}
\label{fig:robustness:cnll}

\vskip -0.1in

\end{figure}

\Cref{fig:robustness:cnll} shows that spatial smoothing not only improves the efficiency but also corruption robustness across a whole range of ensemble size. See \cref{fig:robustness} for more detailed results.
Likewise, spatial smoothing also improves adversarial robustness and perturbation consistency ($\uparrow$) \citep{hendrycks2019benchmarking,zhang2019making}, shift-transformation invariance. See \cref{tab:adversarial}, \cref{tab:consistency}, and \cref{fig:perturbation} for more details.

\subsection{Semantic Segmentation}

\cref{tab:semseg-performance} summarizes the result of semantic segmentation on CamVid dataset \citep{brostow2008segmentation} that consists of real-world 360$\times$480 pixels videos. The table shows that spatial smoothing improves predictive performance, which is consistent with the image classification experiment. Moreover, the result reveals that \emph{spatial smoothing and temporal smoothing \citep{park2019vqbnn} are complementary}. 
See \cref{tab:extended:semseg-performance} for more detailed results.

%% file: resources/tab-semantic-segmentation-sim.tex
\begin{table*}[t]
\vskip 0.0in
\setlength\extrarowheight{3pt}

\caption{
\textbf{Spatial smoothing and temporal smoothing are complementary.} 
We provide predictive performance of MC dropout in semantic segmentation. \textsc{Spat} and \textsc{Temp} each stand for spatial smoothing and temporal smoothing. 
\textsc{Acc} and \textsc{Cons} stand for accuracy and consistency. 
The numbers in brackets denote the performance improvements over the baseline.
}\label{tab:semseg-performance}
\vskip -0.2in
\begin{center}
\begin{small}
\begin{sc}

  \begin{tabular}{cccccccccccc}
    \toprule

    Spat & Temp & NLL & \thead{Acc\\(\%)} & \thead{ECE\\(\%)} & \thead{Cons\\(\%)} \\
    \midrule
    $\cdot$ & $\cdot$ & 0.298 \color{Gray}(-0.000) & 92.5 \color{Gray}(+0.0) & 4.20 \color{Gray}(-0.00) & 95.4 \color{Gray}(+0.0) \\  
    \checkmark & $\cdot$ & 0.284  \color{Green}(-0.014) & 92.6 \color{Green}(+0.1) & 3.96 \color{Green}(-0.24) & 95.6 \color{Green}(+0.2) \\
    $\cdot$ & \checkmark & 0.273 \color{Green}(-0.025) & 92.6 \color{Green}(+0.1) & 3.23 \color{Green}(-0.97) & 96.4 \color{Green}(+1.0) \\  
    \checkmark & \checkmark & \textbf{0.260 \color{Green}(-0.038)} & \textbf{92.6 \color{Green}(+0.1)} & \textbf{2.71 \color{Green}(-1.49)} & \textbf{96.5 \color{Green}(+1.1)} \\

    \bottomrule
  \end{tabular}

\end{sc}
\end{small}
\end{center}

\vskip -0.10in
\end{table*}

%% file: body/related-work.tex
\section{Related Work}\label{sec:related-work}

Spatial smoothing can be compared with prior works in the following areas.

\paragraph{Anti-aliased CNNs. }

Local means \citep{zhang2019making,zou2020delving,vasconcelos2020effective,sinha2020curriculum} were introduced for the shift-invariance of deterministic CNNs in image classification. They were motivated to prevent the aliasing effect of subsampling, and used variants of \texttt{Blur} alone. Although these local filtering can result in a loss of information, \citet{zhang2019making} experimentally observed an increase in accuracy that was beyond expectation. 

However, we show that \emph{the predictive performance improvement of anti-aliased CNNs is not due to anti-aliasing effect of local mean.} 
In particular, \texttt{Prob} plays a key role in the improvement of the predictions as discussed in \cref{sec:prob-role}, and the performance improvement of anti-aliased CNNs is due to the cooperation of \texttt{Blur} and activation as \texttt{Prob}, which was not intended in prior works.
Several experimental results, e.g., \cref{fig:extended:preact}, support this claim by showing that \texttt{Blur} that does not cooperate with activation harms their predictive performance, and adding \texttt{Prob} before \texttt{Blur} surprisingly improves NNs.
Furthermore, our spatial smoothing, which exploits \texttt{Prob}, significantly outperforms \citet{sinha2020curriculum}'s anti-aliased CNN by up to +6.1 percent point on CIFAR-100 in accuracy. 

We provide a fundamental explanation for this phenomenon: \emph{spatial smoothing (\texttt{Prob}--\texttt{Blur}) behaves like an ensemble}. 
An ensemble not only improves accuracy, but also uncertainty and robustness of deterministic and Bayesian NNs \citep{lakshminarayanan2017simple,wilson2020bayesian}. 
For a discussion on non-local means \citep{wang2018non} and self-attention \citep{dosovitskiy2020image}, see \cref{sec:conclusion}.

\paragraph{Sampling-free BNNs. }

Sampling-free BNNs \citep{hernandez2015probabilistic,wang2016natural,wu2018deterministic} predict results based on a single or couple of NN executions. To this end, it is assumed that posterior and feature maps follow Gaussian distributions. However, the discrepancy between reality and assumption accumulates in every NN layer. Consequently, to the best of our knowledge, most of the sampling-free BNNs could only be applied to shallow models, such as LeNet, and were tested on small datasets.
\citet{postels2019sampling} applied sampling-free BNNs to SegNet; nonetheless, \citet{park2019vqbnn} argued that they do not predict well-calibrated results.

\paragraph{Efficient deep ensembles. }

Deep ensemble \citep{lakshminarayanan2017simple,fort2019deep} is another probabilistic NN approach for predicting reliable results.
BatchEnsemble \citep{wen2020batchensemble,dusenberry2020efficient} ensembles over a low-rank subspace to make deep ensemble more efficient.
Depth uncertainty network \citep{antoran2020depth} aggregates feature maps from different depths of a single NN to predict results efficiently. Despite being robust against data corruption, it provides weaker predictive performance compared to deterministic NN and MC dropout.

%% file: body/discussion.tex
\section{Discussion}
\label{sec:conclusion}

We propose spatial smoothing, a non-trainable module motivated by \emph{a spatial ensemble}, for improving NNs. Three different aspects---namely, feature map variance, Fourier analysis, and loss landscapes---show that spatial smoothing behaves like an ensemble that aggregates neighboring feature maps. The module is simple yet efficient, suggesting that \emph{exploiting spatial consistency is important}.
This novel perspective will shape future work in an interesting way. 

The limitation of spatial smoothing is that designing its components requires inductive bias. In other words, the optimal shape of the blur kernel is model-dependent. 
We believe this problem can be solved by introducing self-attention \citep{vaswani2017attention}. Self-attentions for computer vision \citep{dosovitskiy2020image}, also known as Vision Transformers, can be deemed as trainable importance-weighted ensembles of feature maps. 
Therefore, using self-attentions to generalize spatial smoothing would be a promising work (e.g., \citet{park2021vision}) because it not only expands our work, but also helps deepen our understanding of self-attentions.

%% file: appendix/experimental-setup.tex
\section{Experimental Setup and Datasets}\label{sec:conf}

We obtain the main experimental results with the Intel Xeon W-2123 Processor, 32GB memory, and a single GeForce RTX 2080 Ti for CIFAR \citep{krizhevsky2009learning} and CamVid \citep{brostow2008segmentation}. For ImageNet \citep{ILSVRC15}, we use AMD Ryzen Threadripper 3960X 24-Core Processor, 256GB memory, and four GeForce RTX 2080 Ti. 
We conduct ablation studies with four Intel Intel Broadwell CPUs, 15GB memory, and a single NVIDIA T4. 
Models are implemented in PyTorch \citep{paszke2019pytorch}. 
The detailed configurations of image classification and semantic segmentation are as follows.

\subsection{Image Classification}

\input{resources/fig-dropoutrate}

We use VGG \citep{simonyan2014very}, ResNet \citep{he2016deep}, pre-activation ResNet \citep{he2016deep}, and ResNeXt \citep{xie2017aggregated} in image classification. According to the structure suggested by \citet{zagoruyko2016wide}, each block of Bayesian NNs contains one MC dropout layer.

NNs are trained using categorical cross-entropy loss and SGD optimizer with initial learning rate of 0.1, momentum of 0.9, and weight decay of $5\times10^{-4}$. We also use multi-step learning rate scheduler with milestones at 60, 130, and 160, and gamma of 0.2 on CIFAR, and with milestones at 30, 60, and 80, and gamma of 0.2 on ImageNet. We train NNs for 200 epochs with batch size of 128 on CIFAR, and for 90 epochs with batch size of 256 on ImageNet. We start training with gradual warmup \citep{goyal2017accurate} for 1 epoch on CIFAR. 
Basic data augmentations, namely random cropping and horizontal flipping, are used. One exception is the training of ResNeXt on ImageNet. In this case, we use the batch size of 128 and learning rate of 0.05 because of memory limitation.

We use hyperparameters that minimizes NLL of ResNet: $\tau = 10$, and MC dropout rate of 30\% for CIFAR and 5\% for ImageNet. We use $\vert \bm{k} \vert = 2$ for the sake of implementation simplicity.
For fair comparison, models with and without spatial smoothing share hyperparameters such as MC dropout rate. However, \cref{fig:dropoutrate} shows that spatial smoothing improves predictive performance of ResNet-18 at all dropout rates on CIFAR-100. The default ensemble size of MC dropout is 50. We report averages of three evaluations, and error bars in figures represent min and max values. 
Standard deviations are omitted from tables for better visualization. See source code 
for other details.

\subsection{Semantic Segmentation}

We use U-Net \citep{ronneberger2015u} in semantic segmentation. Following Bayesian SegNet \citep{kendall2017bayesian}, Bayesian U-Net contains six MC dropout layers. We add spatial smoothing before each subsampling layer in U-Net encoder. We use 5 previous predictions and decay rate of $e^{-0.8} \simeq 45\%$ per frame for temporal smoothing.

CamVid consists of 720$\times$960 pixels road scene video sequences. We resize the image bilinearly to 360$\times$480 pixels. We use a list reduced to 11 labels by following previous works, e.g. \citep{kendall2017uncertainties}.

NNs are trained using categorical cross-entropy loss and Adam optimizer with initial learning rate of $0.001$ and $\beta_1$ of 0.9, and $\beta_2$ of 0.999. We train NN for 130 epoch with batch size of 3. The learning rate decreases to $0.0002$ at the 100 epoch. Random cropping and horizontal flipping are used for data augmentation. Median frequency balancing is used to mitigate dataset imbalance. Other details follow \citet{park2019vqbnn}.

\subsection{Hessian Max Eigenvalue Spectrum}\label{sec:hmes}

We investigate Hessians to evaluate the smoothness of the loss landscapes quantitatively. In particular, we calculate Hessian eigenvalue spectrum \citep{ghorbani2019investigation}---distributions of Hessian eigenvalues---to show how spatial smoothing helps NN optimization. The most widely known method to obtain the Hessian eigenvalues is the stochastic Lanczos quadrature algorithm. This algorithm finds representative Hessian eigenvalues for full batch. However, it requires a lot of memory and computing resources, so it is not feasible for many practical NNs.

In order to quantitatively evaluate loss landscapes, an efficient Hessian eigenvalue investigation method is needed.
In the training phase, we calculate the mean gradients with respect to mini-batches, rather than the entire dataset. Therefore, it may be reasonable to investigate the properties of the Hessian ``mini-batch-wisely''. Among those Hessians, large Hessian eigenvalues dominates NN training \citep{ghorbani2019investigation}. Based on these insights, we propose an efficient method, \emph{Hessian max eigenvalue spectrum}, that evaluates the distribution of ``the maximum Hessian eigenvalues for mini-batchs''. To obtain Hessian max eigenvalue spectrum, we use power iteration (\texttt{PowerIter}) to produce only the largest (or top-$k$) eigenvalue of the Hessian. Then, we visualize the spectrum by aggregating the largest eigenvalues for mini-batches. For example, we gather top-1 Hessian eigenvalues for \cref{fig:loss-landscape:hmes}. We summarize the algorithm in \cref{alg:hmes}.

Note that Hessian must be calculated for ``regularized losses'' on ``augmented datasets'', since NN training optimizes NLL + $\ell_2$ regularization on augmented datasets---not NLL on clean datasets; measuring Hessian eigenvalues on clean dataset would give incorrect results.

\begin{algorithm}[tb]
\caption{\, Hessian max eigenvalue spectrum}\label{alg:hmes}
\begin{algorithmic}[1]
\REQUIRE training dataset $\mathcal{D}$, mini-batch size $\vert \mathcal{B} \vert$, number of eigenvalues per mini-batch $k$, loss with regularizations (e.g., $\ell_2$ regularization) $\mathcal{L}(\cdot)$, NN weight $\bm{w}$, data augmentation $g(\cdot)$ 
\ENSURE Hessian max eigenvalue spectrum $\mathcal{H} = \{ \lambda^{(1)}_{1}, \lambda^{(1)}_{2}, \cdots, \lambda^{(2)}_1, \lambda^{(2)}_2, \cdots \}$ where $\lambda^{(i)}_{j}$ is the $j^\text{th}$ largest Hessian eigenvalues for the $i^\text{th}$ mini-batch
\STATE $\mathcal{H} = \{ \}$
\FOR{$i^\text{th}$ mini-batch $\mathcal{B}^{(i)}$ \textbf{in} $\mathcal{D}$}
\STATE 
${ \{ \lambda^{(i)}_{1}, \cdots, \lambda^{(i)}_{k} \} \gets } $
$\hspace*{3em} { \texttt{PowerIter}\left(k, \text{Hessian of } \mathcal{L}( \bm{w}, g(\mathcal{B}^{(i)}) ) \right) }$
\STATE $\mathcal{H} \gets \mathcal{H} \, \bigcup \, \{ \lambda^{(i)}_{1}, \cdots, \lambda^{(i)}_{k} \}$
\ENDFOR
\end{algorithmic}
\end{algorithm}

This algorithm is easy to implement and requires significantly less memory and computational cost, compared with stochastic Lanczos quadrature algorithm with respect to entire dataset. With this method, we can investigate the Hessian of large NNs, which would require a lot of GPU memory. In this paper, we use both Hessian eigenvalue spectra using stochastic Lanczos quadrature algorithm and Hessian max eigenvalue spectrum using power iteration implemented by \citet{yao2020pyhessian}. Compare \cref{fig:loss-landscape:hes} and \cref{fig:loss-landscape:hmes}.

%% file: resources/fig-dropoutrate.tex
\begin{figure*}
\vskip 0.2in
\begin{center}

\begin{subfigure}[b]{0.32\textwidth}
\centering
\includegraphics[width=0.95\textwidth]{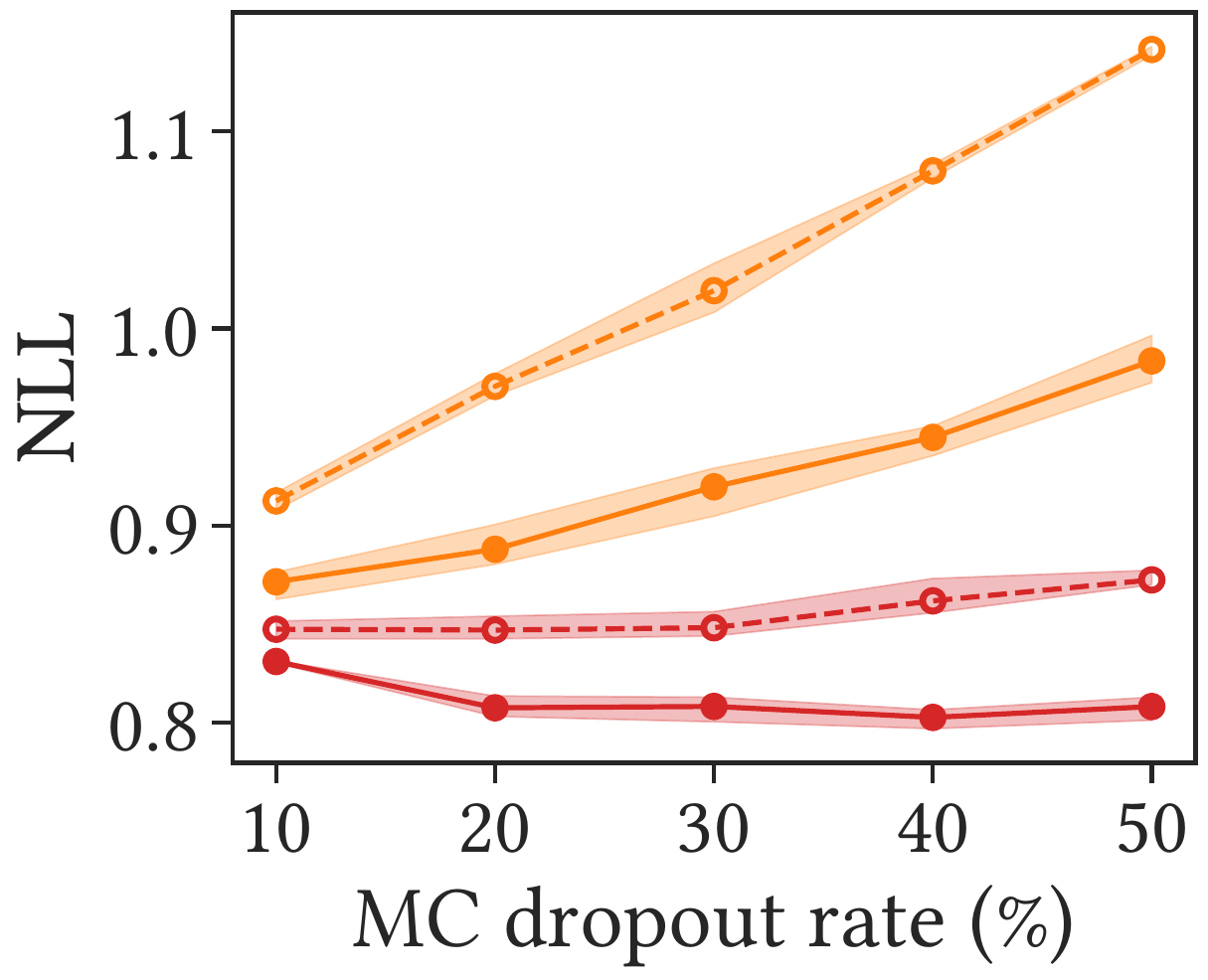}
\end{subfigure}
\hspace{1pt}
\begin{subfigure}[b]{0.32\textwidth}
\centering
\includegraphics[width=0.95\textwidth]{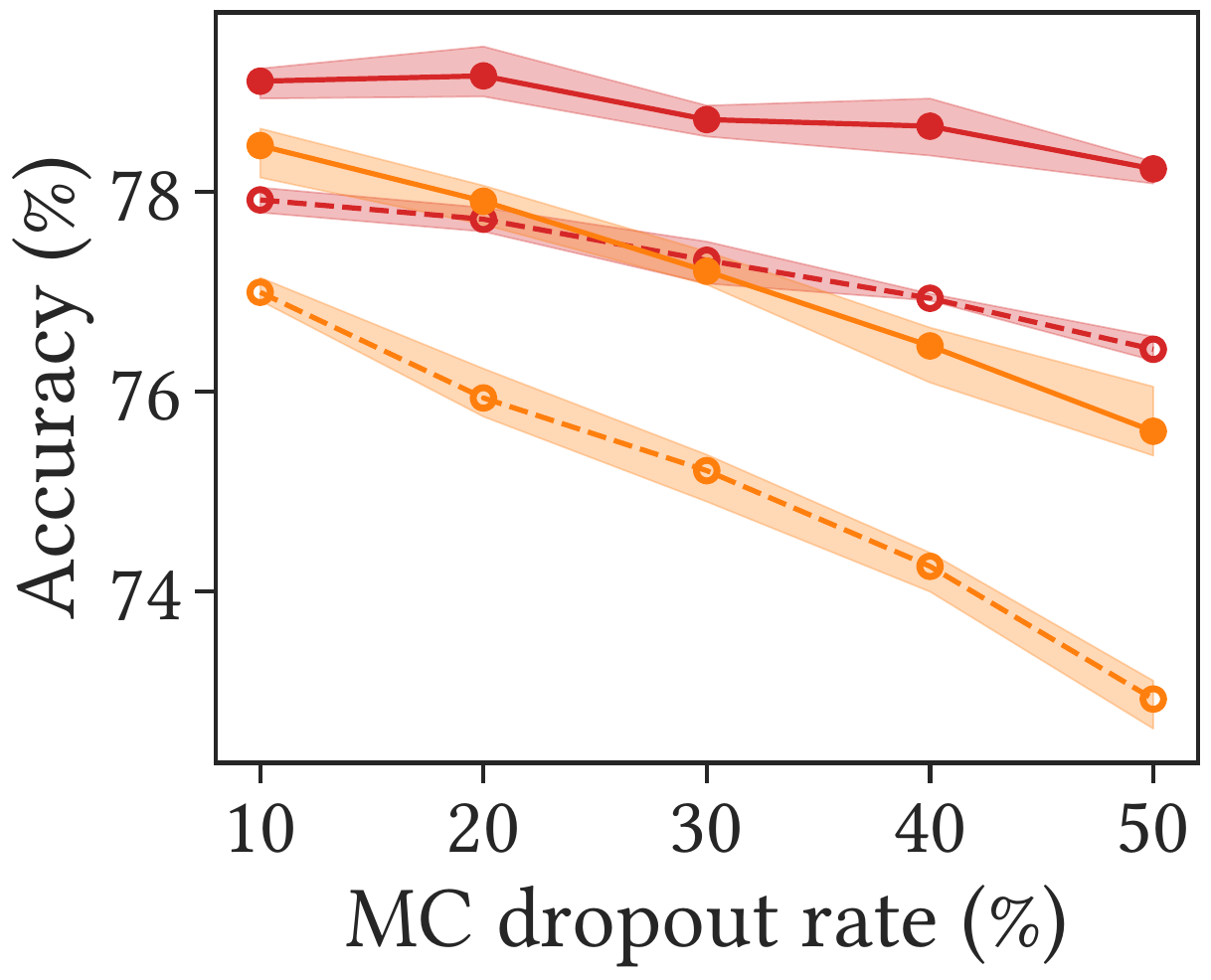}
\end{subfigure}
\hspace{1pt}
\begin{subfigure}[b]{0.32\textwidth}
\centering
\includegraphics[width=0.95\textwidth]{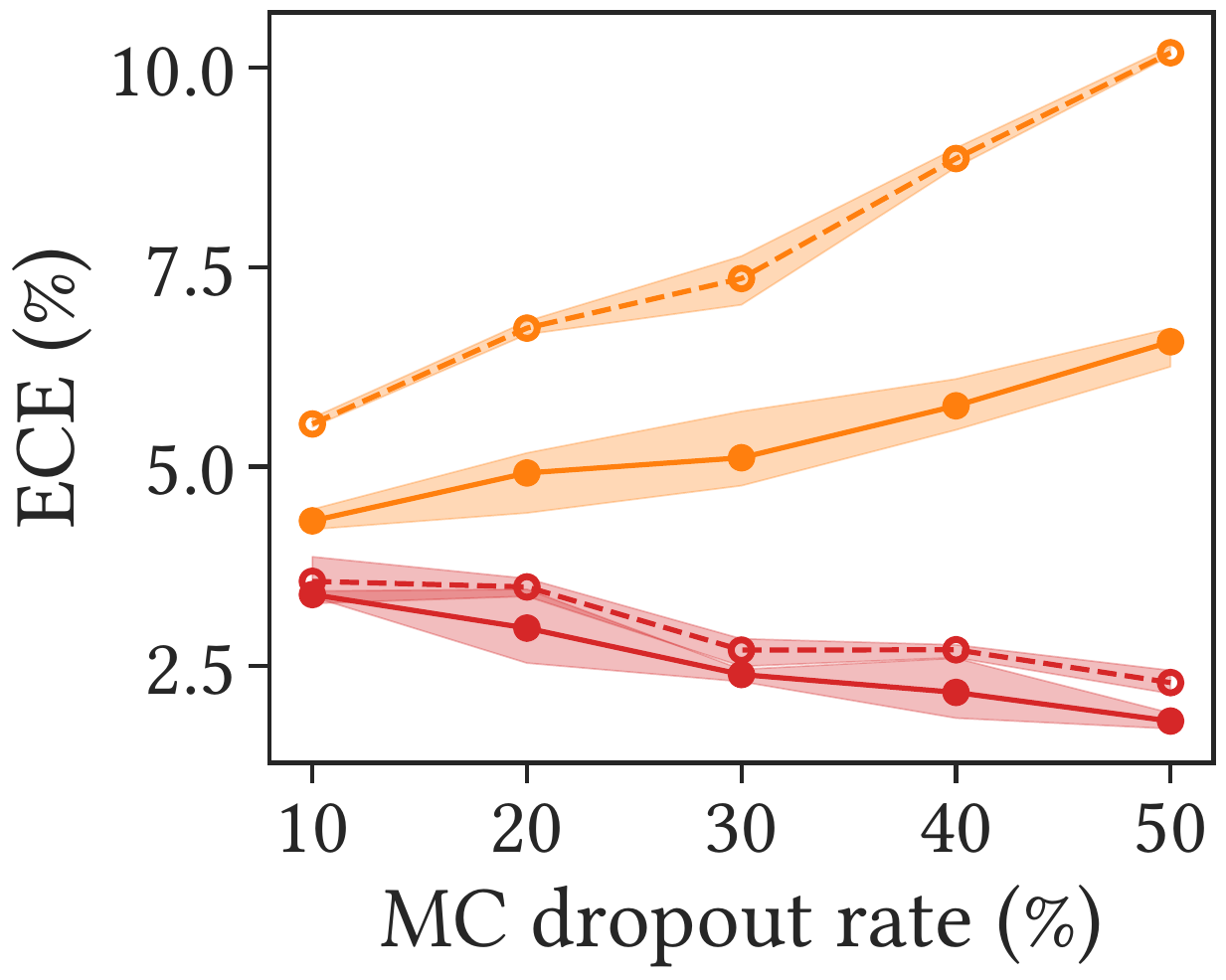}
\end{subfigure}

\end{center}

\vspace{3pt}
\centering
\includegraphics[width=0.95\textwidth]{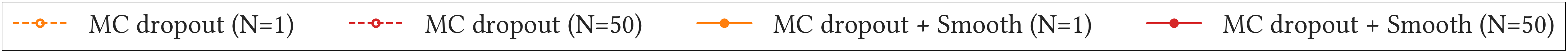}

\caption{
\textbf{Spatial smoothing improves predictive performance at all dropout rates}.
As the dropout rate increases, both accuracy and ECE decrease. The performance is optimized when accuracy and uncertainty are balanced.
}
\label{fig:dropoutrate}
\vskip -0.0in
\end{figure*}

%% file: appendix/ablation-study.tex
\section{Ablation Study}\label{sec:ablation}

\input{resources/fig-temperature}

Probabilistic spatial smoothing proposed in this paper consists of two components: \texttt{Prob} and \texttt{Blur}. This section explores several candidates for each component and their properties.

\subsection{\texttt{Prob}: Feature Maps to Probabilities}
\label{sec:ablation:prob}

\input{resources/tab-prob}

We define \texttt{Prob} as a composition of an upper-bounded function and \texttt{ReLU}, a function that imposes the lower bound of zero. There are several widely used upper-bounded functions: $\texttt{tanh}_{\tau}(\bm{x}) = \tau \, \texttt{tanh} (\bm{x} / \tau)$, $\texttt{ReLU6}(\bm{x}) = \texttt{max}(\texttt{min}(\bm{x}, 6), 0)$, and constant scaling which is $\bm{x} / \tau$. 

\Cref{tab:prob} shows the predictive performance improvement by \texttt{Prob} with various upper-bounded functions on CIFAR-100. In this experiment, we use models with MC dropout, and $\tau = 5$ for constant scaling.
The results indicate that upper-bounded functions with \texttt{ReLU} tend to improve accuracy and uncertainty at the same time. In addition, they show that \texttt{Prob} and \texttt{Blur} are complementary; the best results are obtained when using both \texttt{Prob} and \texttt{Blur}. 
For the main experiments, we use the composition of $\texttt{tanh}_{\tau}$ and \texttt{ReLU} as \texttt{Prob} although constant scaling outperforms in some cases. This is because the hyperparameter of constant scaling is highly dependent on dataset and model.

\paragraph{Temperature. }

The characteristics of temperature-scaled \texttt{tanh} depends on $\tau$. 
This $\texttt{tanh}_{\tau}$ has a couple of useful properties:
$\texttt{tanh}_{\tau}$ has an upper bound of $\tau$, and the first derivative of $\texttt{tanh}_{\tau}$ at $x = 0$ does not depend on $\tau$.

\Cref{fig:temperature} shows the predictive performance of ResNet-18 with MC dropout and spatial smoothing for the temperature on CIFAR-100. In this figure, \emph{the accuracy increases as the temperature increases}. In contrast, in terms of ECE, \emph{NN predicts more underconfident results as $\tau$ decreases}. NLL, a metric representing both accuracy and uncertainty, is minimized when the accuracy and the uncertainty are balanced. \emph{In conclusion, we set the default value of $\tau$ to 10.}

It is a misinterpretation that the result is overconfident at low $\tau$ because ECE is high. By definition, ECE relies on the absolute value of the difference between confidence and accuracy. In this example, at low $\tau$, the accuracy is greater than the confidence, which leads to a high ECE. Moreover, at $\tau = 0.2$, ECE with $N = 50$ is greater than that with $N = 1$, which means that the result is severely underconfident.

\subsection{\texttt{Blur}: Averaging Neighboring Probabilities}
\label{sec:ablation:blur}

\input{resources/tab-blur}

\texttt{Blur} is a depth-wise convolution with a normalized kernel. 
The kernel given by \cref{eq:blur} is derived from various $\bm{k}$s such as $\bm{k} \in \{ \left( 1 \right), \left( 1, 1 \right),  \left( 1, 2, 1 \right), \left( 1, 4, 6, 4, 1 \right), \cdots \}$.
In these examples, if $\vert \bm{k} \vert$ is 1, \texttt{Blur} is identity. If $\vert \bm{k} \vert$ is 2, \texttt{Blur} is a box blur, which is used in the main experiments. If $\vert \bm{k} \vert$ is 3 or 5, \texttt{Blur} is an approximated Gaussian blur.

\cref{tab:blur} shows predictive performance of models using spatial smoothing with the kernels on CIFAR-100. 
This results show that \emph{most kernels improve both accuracy and uncertainty}. The most effective kernel size depends on the model.

\subsection{Position of Spatial Smoothing }
\label{sec:ablation:position}

\input{resources/fig-position}

As shown in \cref{fig:fm-variance}, the magnitude of feature map uncertainty tends to increase as the depth increases.
Therefore, we expect that spatial smoothing close to the output layer will mainly drive performance improvement.

We investigate the predictive performance of models with MC dropout using only \emph{one} spatial smoothing layer. \Cref{fig:position} shows the predictive performance of ResNet-18 with one spatial smoothing after each stage on CIFAR-100. The results suggest that spatial smoothing after \texttt{s3} is the most important for improving performance. Surprisingly, spatial smoothing after \texttt{s4} is the least important. This is because GAP, the most extreme case of spatial smoothing, already exists there.

%% file: resources/fig-temperature.tex
\begin{figure*}
\vskip 0.2in
\begin{center}

\begin{subfigure}[b]{0.32\textwidth}
\centering
\includegraphics[width=0.95\textwidth]{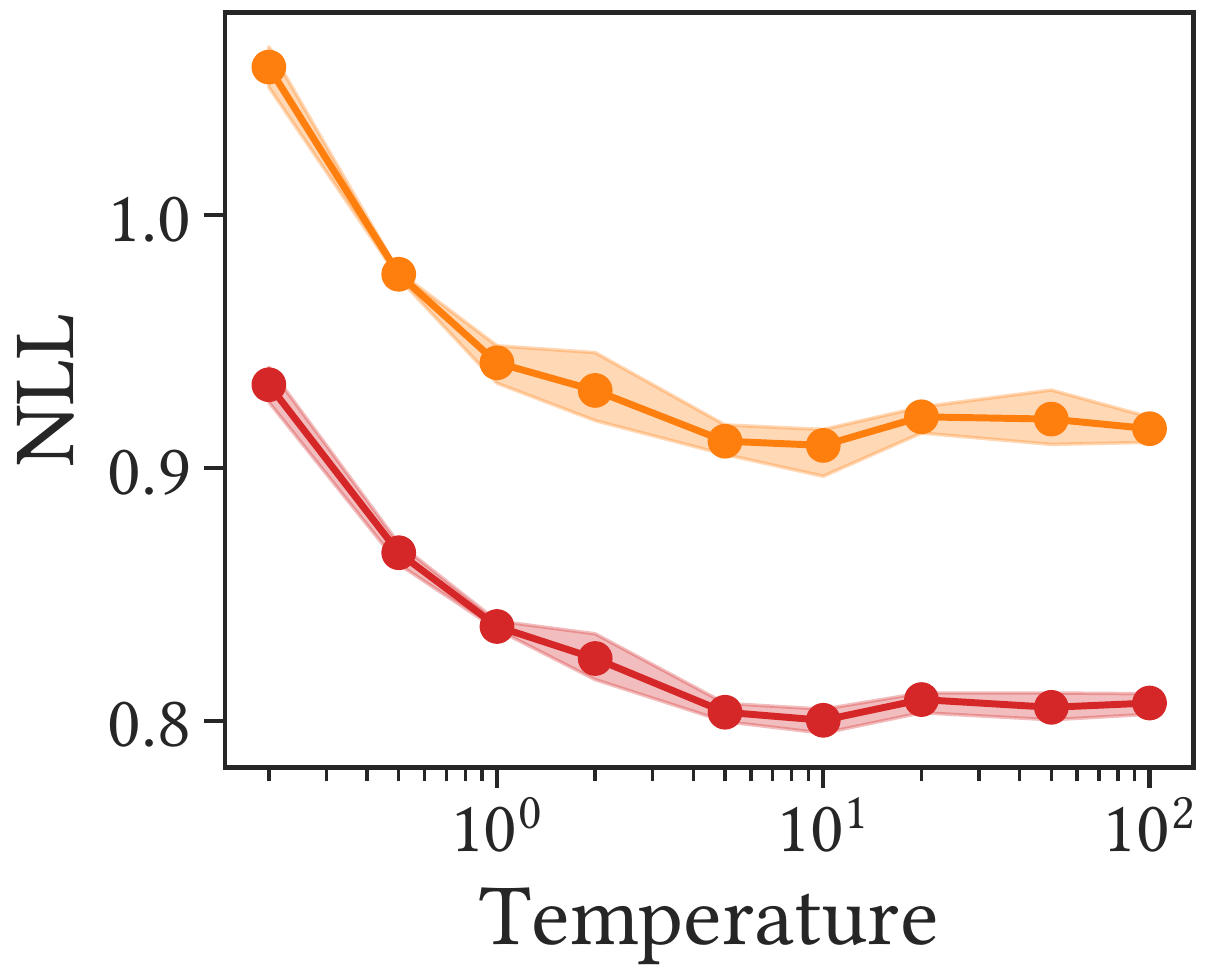}
\end{subfigure}
\hspace{1pt}
\begin{subfigure}[b]{0.32\textwidth}
\centering
\includegraphics[width=0.95\textwidth]{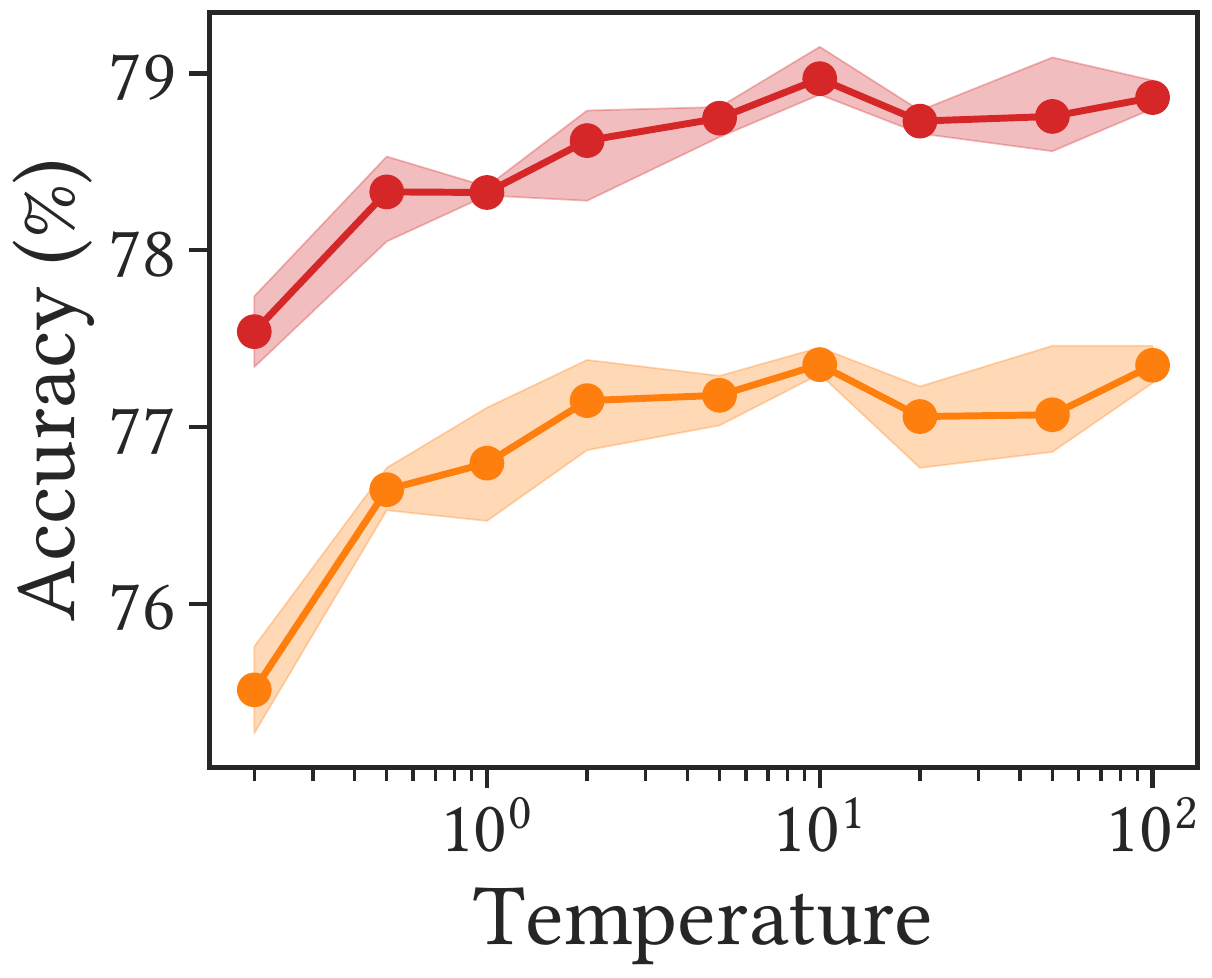}
\end{subfigure}
\hspace{1pt}
\begin{subfigure}[b]{0.32\textwidth}
\centering
\includegraphics[width=0.95\textwidth]{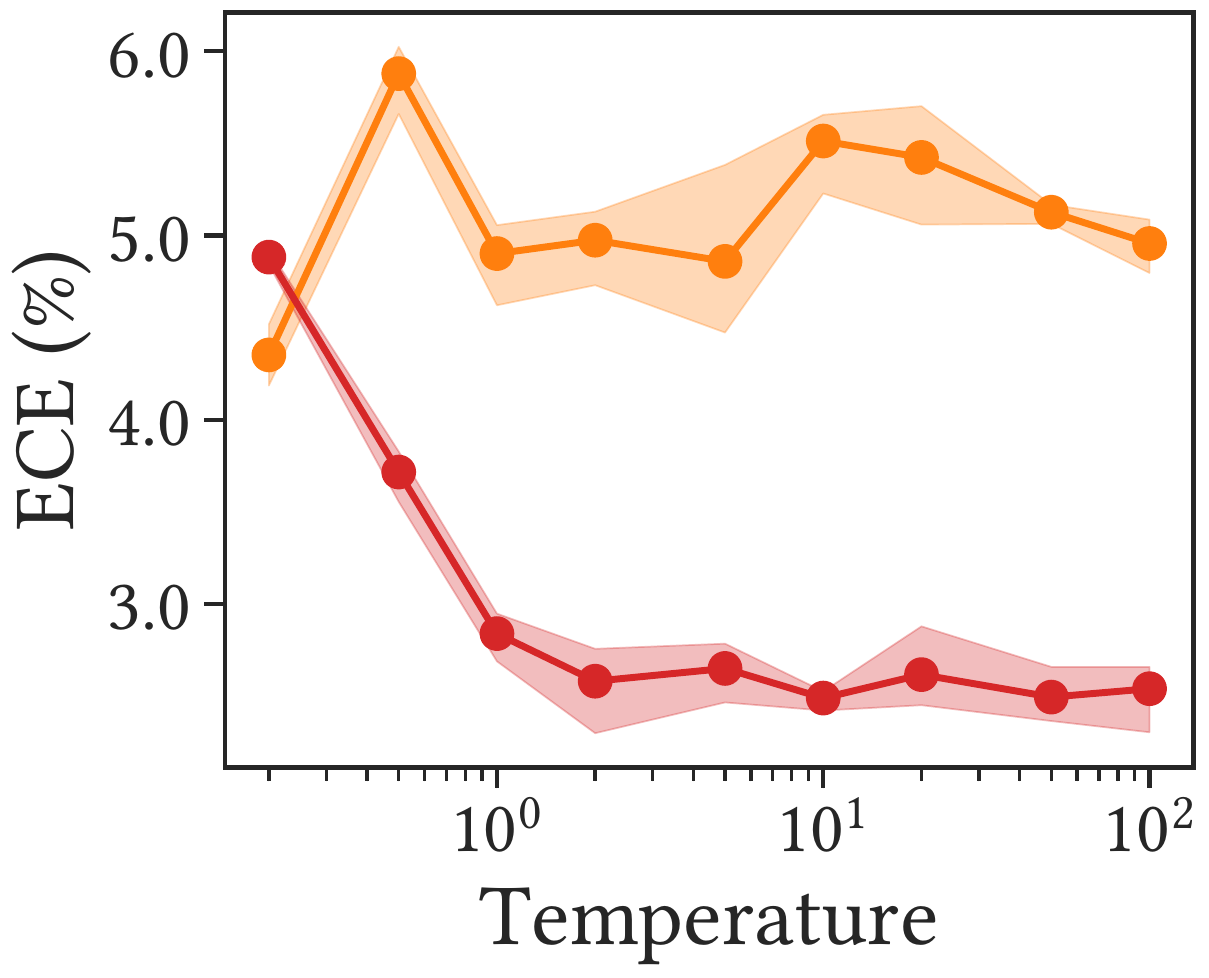}
\end{subfigure}

\end{center}

\vspace{3pt}
\centering
\includegraphics[height=0.027\textheight]{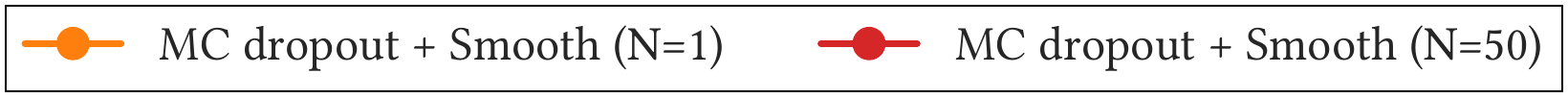}

\caption{
\textbf{The temperature controls the trade-off between accuracy and uncertainty}.
The accuracy increases as the temperature increases, but predictions become more overconfident. 
}
\label{fig:temperature}
\vskip 0.1in
\end{figure*}

%% file: resources/tab-prob.tex
\begin{table*}
\vskip 0.1in
\setlength\extrarowheight{3pt} 

\caption{
\textbf{We use \texttt{tanh} as the default for \texttt{Prob}} based on the predictive performance of MC dropout for CIFAR-100 with various \texttt{Prob}s.
}

\begin{center}
\begin{small}

  \begin{tabular}{cccccccccccc}
    \toprule

    \textsc{\thead{Model}} & \textsc{Smooth} & \textsc{NLL} & \textsc{\thead{Acc\\(\%)}} & \textsc{\thead{ECE\\(\%)}} \\
    \midrule
    \multirow{10}{*}{\thead{VGG-16}}
    & $\cdot$ & 1.133 \color{Gray}(-0.000) & 68.8 \color{Gray}(+0.0) & 3.66 \color{Gray}(+0.00) \\
    & \thead{\texttt{ReLU} $\circ$ \texttt{\ul{tanh}}} & 1.064 \color{Green}(-0.069) & 70.4 \color{Green}(+1.6) & 2.99 \color{Green}(-0.67) \\
    & \thead{\texttt{ReLU} $\circ$ \texttt{\ul{ReLU6}}} & 1.093 \color{Green}(-0.040) & 69.8 \color{Green}(+1.0) & 4.26 \color{Red}(+0.60) \\
    & \thead{\texttt{ReLU} $\circ$ \texttt{\ul{Constant}}} & \textbf{0.995 \color{Green}(-0.138)} & \textbf{72.5 \color{Green}(+3.7)} & \textbf{2.11 \color{Green}(-1.55)} \\
    \cline{2-5}
    & \thead{\texttt{Blur}} & 0.985 \color{Gray}(-0.000) & 72.4 \color{Gray}(+0.0) & 1.77 \color{Gray}(+0.00) \\
    & \thead{\texttt{Blur} $\circ$ \texttt{ReLU} $\circ$ \texttt{\ul{tanh}}} & 0.984 \color{Green}(-0.001) & 72.7 \color{Green}(+0.3) & 2.07 \color{Red}(+0.30) \\
    & \thead{\texttt{Blur} $\circ$ \texttt{ReLU} $\circ$ \texttt{\ul{ReLU6}}} & \textbf{0.982 \color{Green}(-0.003)} & 72.5 \color{Green}(+0.1) & 1.84 \color{Red}(+0.07) \\
    & \thead{\texttt{Blur} $\circ$ \texttt{ReLU} $\circ$ \texttt{\ul{Constant}}} & 0.991 \color{Red}(+0.005) & \textbf{72.9 \color{Green}(+0.5)} & \textbf{1.03 \color{Green}(-0.74)} \\
    \midrule
    \multirow{10}{*}{\thead{VGG-19}}
    & $\cdot$ & 1.215 \color{Gray}(-0.000) & 67.3 \color{Gray}(+0.0) & 6.37 \color{Gray}(+0.00) \\
    & \thead{\texttt{ReLU} $\circ$ \texttt{\ul{tanh}}} & 1.131 \color{Green}(-0.084) & 69.2 \color{Green}(+1.9) & 5.23 \color{Green}(-1.14) \\
    & \thead{\texttt{ReLU} $\circ$ \texttt{\ul{ReLU6}}} & 1.166 \color{Green}(-0.049) & 68.3 \color{Green}(+1.0) & 6.44 \color{Green}(-0.06) \\
    & \thead{\texttt{ReLU} $\circ$ \texttt{\ul{Constant}}} & \textbf{0.997 \color{Green}(-0.218)} & \textbf{72.5 \color{Green}(+5.2)} & \textbf{1.09 \color{Green}(-5.29)} \\
    \cline{2-5}
    & \thead{\texttt{Blur}} & 1.039 \color{Gray}(-0.000) & 71.1 \color{Gray}(+0.0) & 3.12 \color{Gray}(+0.00) \\
    & \thead{\texttt{Blur} $\circ$ \texttt{ReLU} $\circ$ \texttt{\ul{tanh}}} & 1.034 \color{Green}(-0.005) & 71.3 \color{Green}(+0.2) & 3.31 \color{Red}(+0.19) \\
    & \thead{\texttt{Blur} $\circ$ \texttt{ReLU} $\circ$ \texttt{\ul{ReLU6}}} & 1.038 \color{Green}(-0.002) & 71.3 \color{Green}(+0.2) & 3.84 \color{Red}(+0.72) \\
    & \thead{\texttt{Blur} $\circ$ \texttt{ReLU} $\circ$ \texttt{\ul{Constant}}} & \textbf{0.995 \color{Green}(-0.045)} & \textbf{72.3 \color{Green}(+1.2)} & \textbf{1.41 \color{Green}(-1.71)} \\
    \midrule
    \multirow{10}{*}{\thead{ResNet-18}}
    & $\cdot$ & 0.848 \color{Gray}(-0.000) & 77.3 \color{Gray}(+0.0) & 3.01 \color{Gray}(+0.00) \\
    & \thead{\texttt{ReLU} $\circ$ \texttt{\ul{tanh}}} & 0.838 \color{Green}(-0.010) & \textbf{77.7 \color{Green}(+0.4)} & 2.92 \color{Green}(-0.08) \\
    & \thead{\texttt{ReLU} $\circ$ \texttt{\ul{ReLU6}}} & 0.844 \color{Green}(-0.004) & 77.4 \color{Green}(+0.1) & 2.74 \color{Green}(-0.27) \\
    & \thead{\texttt{ReLU} $\circ$ \texttt{\ul{Constant}}} & \textbf{0.825 \color{Green}(-0.023)} & 77.7 \color{Green}(+0.4) & \textbf{1.87 \color{Green}(-1.14)} \\
    \cline{2-5}
    & \thead{\texttt{Blur}} & 0.806 \color{Gray}(-0.000) & 78.6 \color{Gray}(+0.0) & 2.56 \color{Gray}(+0.00) \\
    & \thead{\texttt{Blur} $\circ$ \texttt{ReLU} $\circ$ \texttt{\ul{tanh}}} & \textbf{0.801 \color{Green}(-0.005)} & \textbf{78.9 \color{Green}(+0.3)} & 2.56 \color{Green}(-0.01) \\
    & \thead{\texttt{Blur} $\circ$ \texttt{ReLU} $\circ$ \texttt{\ul{ReLU6}}} & 0.805 \color{Green}(-0.001) & 78.9 \color{Green}(+0.2) & 2.59 \color{Red}(+0.03) \\
    & \thead{\texttt{Blur} $\circ$ \texttt{ReLU} $\circ$ \texttt{\ul{Constant}}} & 0.811 \color{Red}(+0.005) & 78.5 \color{Red}(-0.2) & \textbf{1.84 \color{Green}(-0.72)} \\
    \midrule
    \multirow{10}{*}{ResNet-50}
    & $\cdot$ & 0.822 \color{Gray}(-0.000) & 79.1 \color{Gray}(+0.0) & 6.63 \color{Gray}(+0.00) \\
    & \thead{\texttt{ReLU} $\circ$ \texttt{\ul{tanh}}} & 0.812 \color{Green}(-0.010) & 79.3 \color{Green}(+0.2) & 6.74 \color{Red}(+0.11) \\
    & \thead{\texttt{ReLU} $\circ$ \texttt{\ul{ReLU6}}} & 0.799 \color{Green}(-0.023) & 79.4 \color{Green}(+0.3) & 6.71 \color{Red}(+0.08) \\
    & \thead{\texttt{ReLU} $\circ$ \texttt{\ul{Constant}}} & \textbf{0.788 \color{Green}(-0.034)} & \textbf{79.6 \color{Green}(+0.5)} & \textbf{5.22 \color{Green}(-1.41)} \\
    \cline{2-5}
    & \thead{\texttt{Blur}} & 0.798 \color{Gray}(-0.000) & 80.0 \color{Gray}(+0.0) & 7.21 \color{Gray}(+0.00) \\
    & \thead{\texttt{Blur} $\circ$ \texttt{ReLU} $\circ$ \texttt{\ul{tanh}}} & 0.800 \color{Red}(+0.002) & 80.1 \color{Green}(+0.1) & 7.25 \color{Red}(+0.04) \\
    & \thead{\texttt{Blur} $\circ$ \texttt{ReLU} $\circ$ \texttt{\ul{ReLU6}}} & 0.800 \color{Red}(+0.002) & 80.2 \color{Green}(+0.2) & 7.30 \color{Red}(+0.09) \\
    & \thead{\texttt{Blur} $\circ$ \texttt{ReLU} $\circ$ \texttt{\ul{Constant}}} & \textbf{0.779 \color{Green}(-0.019)} & \textbf{80.4 \color{Green}(+0.4)} & \textbf{5.81 \color{Green}(-1.40)} \\
    \bottomrule
  \end{tabular}

\end{small}
\end{center}
\vskip 0.1in

\label{tab:prob}
\end{table*}

%% file: resources/tab-blur.tex
\begin{table*}
\vskip 0.1in
\setlength\extrarowheight{3pt} 

\caption{
\textbf{The optimal shape of the blur kernel is model-dependent}.
We measure the predictive performance of MC dropout using spatial smoothing with various size of \texttt{Blur} kernels on CIFAR-100.
}

\begin{center}
\begin{small}

  \begin{tabular}{cccccccccccc}
    \toprule
    \textsc{\thead{Model}} & $\vert \bm{k} \vert$ & \textsc{NLL} & \textsc{\thead{Acc\\(\%)}} & \textsc{\thead{ECE\\(\%)}} \\
    \midrule
    \multirow{4}{*}{\thead{VGG-16}}
    & 1 & 1.087 \color{Gray}(-0.000) & 69.8 \color{Gray}(+0.0) & 3.43 \color{Gray}(-0.00) \\
    & 2 & 1.034 \color{Green}(-0.053) & 71.4 \color{Green}(+1.6) & \textbf{1.06 \color{Green}(-2.37)} \\
    & 3 & \textbf{0.986 \color{Green}(-0.101)} & \textbf{72.7 \color{Green}(+2.9)} & 1.03 \color{Green}(-2.40) \\
    & 5 & 1.018 \color{Green}(-0.069) & 72.0 \color{Green}(+2.2) & 1.32 \color{Green}(-2.11) \\
    \midrule
    \multirow{4}{*}{\thead{VGG-19}}
    & 1 & 1.096 \color{Gray}(-0.000) & 69.8 \color{Gray}(+0.0) & 4.74 \color{Gray}(-0.00) \\
    & 2 & 1.071 \color{Green}(-0.025) & 70.4 \color{Green}(+0.6) & \textbf{2.15 \color{Green}(-2.59)} \\
    & 3 & \textbf{1.026 \color{Green}(-0.070)} & \textbf{71.9 \color{Green}(+2.1)} & 2.56 \color{Green}(-2.18) \\
    & 5 & 1.032 \color{Green}(-0.064) & 71.6 \color{Green}(+1.8) & 2.16 \color{Green}(-2.58) \\
    \midrule
    \multirow{4}{*}{\thead{ResNet-18}}
    & 1 & 0.840 \color{Gray}(-0.000) & 77.6 \color{Gray}(+0.0) & 2.63 \color{Gray}(-0.00) \\
    & 2 & \textbf{0.801 \color{Green}(-0.039)} & \textbf{78.9 \color{Green}(+1.4)} & \textbf{2.56 \color{Green}(-0.07)} \\
    & 3 & 0.822 \color{Green}(-0.018) & 78.7 \color{Green}(+1.1) & 2.86 \color{Green}(-0.23) \\
    & 5 & 0.837 \color{Green}(-0.003) & 78.4 \color{Green}(+0.8) & 3.05 \color{Green}(-0.42) \\
    \midrule
    \multirow{4}{*}{\thead{ResNet-50}}
    & 1 & 0.814 \color{Gray}(-0.000) & 79.5 \color{Gray}(+0.0) & \textbf{6.56 \color{Gray}(-0.00)} \\
    & 2 & 0.806 \color{Green}(-0.008) & \textbf{80.0 \color{Green}(+0.5)} & 7.35 \color{Red}(+0.79) \\
    & 3 & \textbf{0.796 \color{Green}(-0.019)} & 79.9 \color{Green}(+0.4) & 7.38 \color{Red}(+0.82) \\
    & 5 & 0.816 \color{Red}(+0.001) & 79.4 \color{Red}(-0.1) & 7.38 \color{Red}(+0.82) \\
    \bottomrule
  \end{tabular}

\end{small}
\end{center}
\vskip 0.1in

\label{tab:blur}
\end{table*}

%% file: resources/fig-position.tex
\begin{figure*}
\vskip 0.2in
\begin{center}

\begin{subfigure}[b]{0.32\textwidth}
\centering
\includegraphics[width=0.95\textwidth]{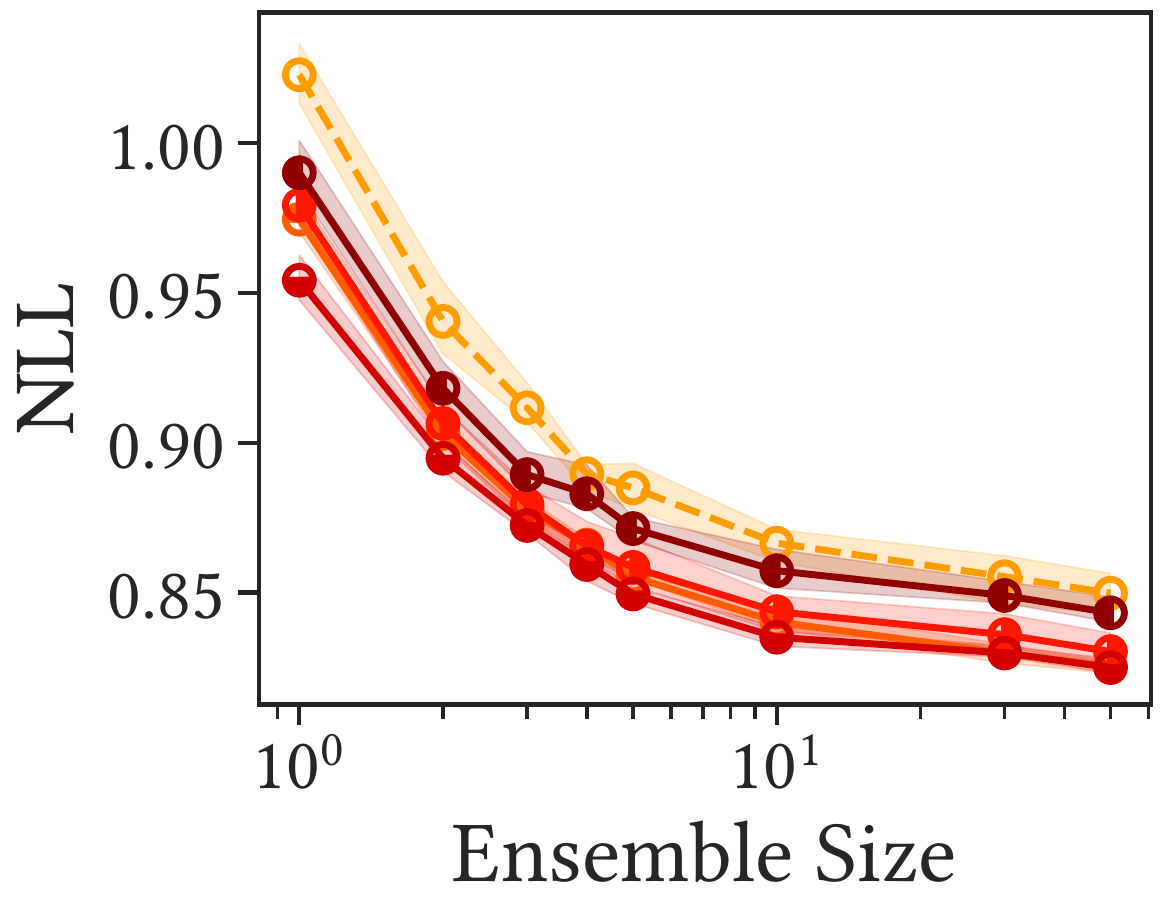}
\end{subfigure}
\hspace{1pt}
\begin{subfigure}[b]{0.32\textwidth}
\centering
\includegraphics[width=0.95\textwidth]{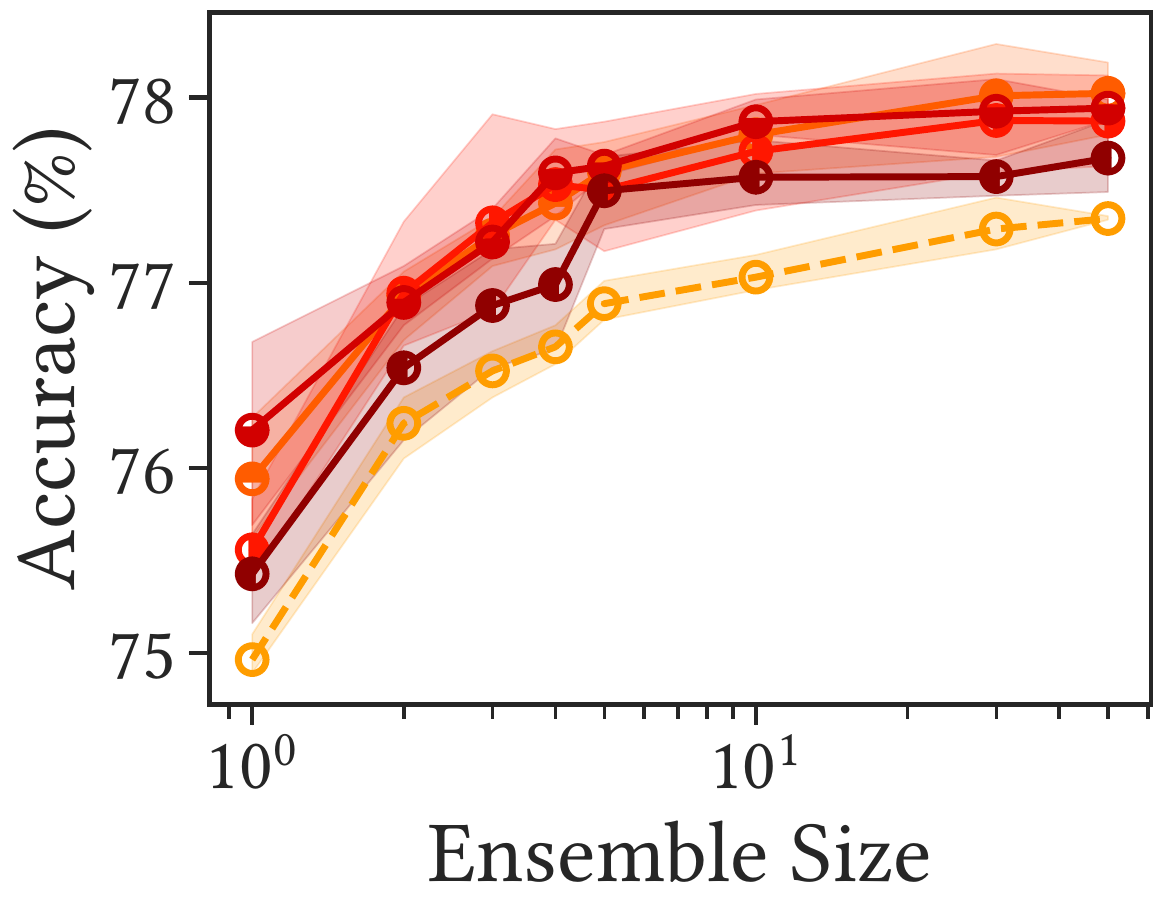}
\end{subfigure}
\hspace{1pt}
\begin{subfigure}[b]{0.32\textwidth}
\centering
\includegraphics[width=0.95\textwidth]{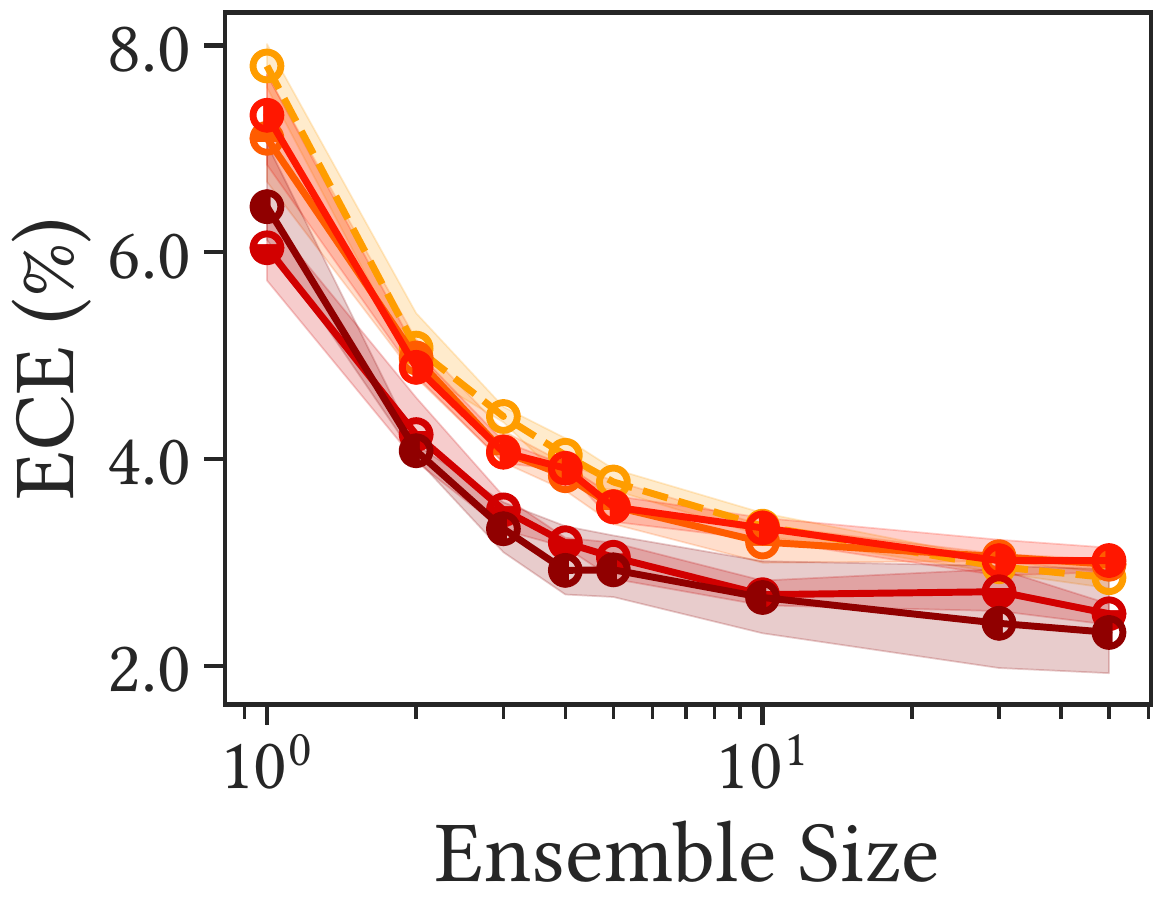}
\end{subfigure}

\end{center}

\vspace{3pt}
\centering
\includegraphics[height=0.027\textheight]{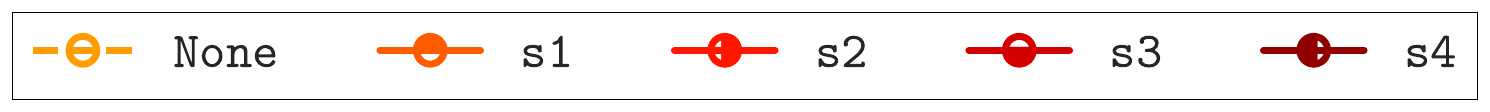}

\caption{
\textbf{Spatial smoothing close to the last layer (\texttt{s3}) significantly improves performance}.
We report predictive performance of ResNet-18 with \emph{one} spatial smoothing after each stage on CIFAR-100.  \texttt{None} indicates vanilla MC dropout.
}
\label{fig:position}
\vskip -0.0in
\end{figure*}

%% file: appendix/revisiting-gap.tex
\section{Revisiting Prior Works}
\label{sec:revisiting}

As mentioned in \cref{sec:spatial-smoothing}, prior works---namely, GAP, pre-activation, and \texttt{ReLU6}---are spacial cases of spatial smoothing. This section discusses them in detail.

\subsection{Global Average Pooling}
\label{sec:revisiting:gap}

The composition of GAP and a fully connected layer is the most popular classifier in classification tasks. The original motivation and the most widely accepted explanation for the success is that \emph{GAP classifier prevents overfitting because it uses significantly fewer parameters than MLP} \citep{lin2013network}. To disprove this claim, we measure the predictive performance of MLP, GAP, and global max pooling (GMaxP), a classifier that uses the same number of parameters as GAP, on training dataset.

\paragraph{Predictive performance. } 

\input{resources/tab-classifier}

\input{resources/fig-classifier-robustness}

\cref{tab:classifier} shows the experimental results on the training and the test dataset of CIFAR-100, suggesting that the explanation is poorly supported. On \emph{both} the training and the test dataset, most predictive performance of MLP is worse than that of GAP. It is a counter-intuitive result meaning that \emph{MLP do not overfit the training dataset}. In addition, the performance improvement by GAP is remarkable in VGG, which has irregular loss landscape. 
The predictive performance of GMaxP is better than that of MLP, but worse than that of GAP. This shows that using fewer parameters partially helps to improve predictive performance; however, it is insufficient to explain the predictive performance improvement by GAP.
Finally, global median pooling (GMedP) provides better predictive performance than GMaxP. It implies that using other noise reduction methods also helps, in part, to improve predictive performance.

\paragraph{Robustness. }

To evaluate the robustness of the classifiers, we measure the predictive performance of ResNet-18 using MC dropout with the classifiers on CIFAR-100-C. \Cref{fig:classifier:robustness} shows the experimental results suggesting that MLP is not robust against data corruption, as we would expect. In terms of accuracy, the robustness of GMaxP and GMedP is relatively comparable to that of GAP; however, in terms of uncertainty, \emph{GAP is the most robust}. These are consistent results with other spatial smoothing experiments.

\input{resources/fig-loss-landscape-sequence}

\paragraph{Loss landscape visualization. }

To understand the mechanism of GAP performance improvement, we investigate the loss landscape. \Cref{fig:loss-landscape:animated} shows the loss landscape sequences of ResNet with MC dropout. In this figure, each sequence shares the bases, but they fluctuate due to the randomness of the MC dropout. \Cref{fig:loss-landscape:animated:mlp} is the loss landscape of the model using MLP classifier instead of GAP classifier. This loss landscape is chaotic and irregular, resulting in hindering and destabilizing NN optimization. \cref{fig:loss-landscape:animated:gap} is loss landscape sequence of ResNet with GAP classifier. Since GAP ensembles all of the feature map points at the last stage, it flattens and stabilizes the loss landscape. Likewise, as shown in \cref{fig:loss-landscape:animated:smooth}, spatial smoothing layers at the end of all stages also flattens and stabilizes the loss landscape.

\paragraph{Hessian eigenvalue spectra. }

\begin{figure}[ht]
\centering

\vskip 0.10in

\raisebox{0pt}[\dimexpr\height-0.6\baselineskip\relax]{
\includegraphics[width=0.32\textwidth]{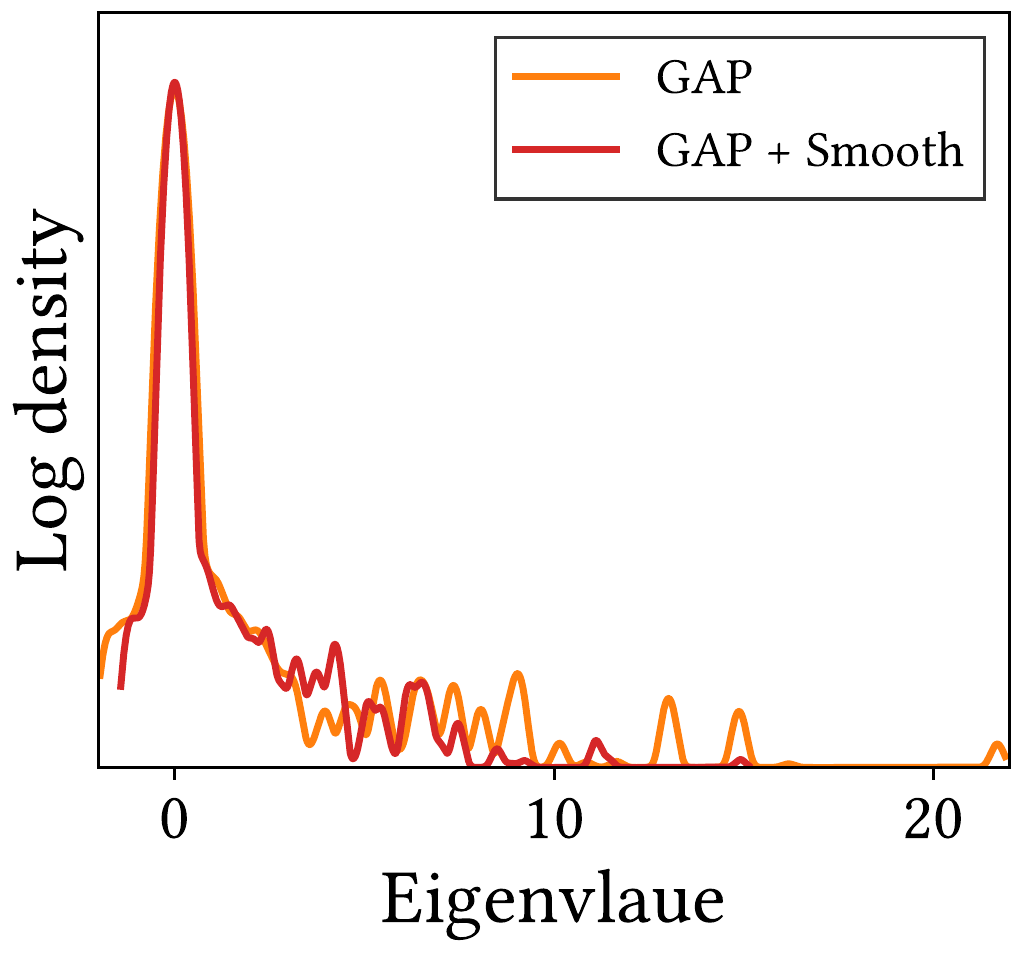}
}

\caption{
\textbf{Spatial smoothing suppress eigenvalue outliers}. 
We provide Hessian eigenvalue spectra of ResNet-18 with MC dropout on CIFAR-100. Compare with \cref{fig:loss-landscape:hmes}.
}
\label{fig:loss-landscape:hes}
\vskip -0.1in
\end{figure}

\Cref{fig:loss-landscape:hmes} shows the Hessian max eigenvalue spectra of MLP classifier model and GAP classifier models with and without spatial smoothing layers. As \citet{li2017visualizing,foret2020sharpness} and \cref{sec:extended-analysis:dropout} pointed out, Hessian eigenvalue outliers disturb NN training. This figure explicitly show that the GAP and spatial smoothing reduce the magnitude of the Hessian eigenvalues and suppress the outliers, which leads to the same conclusion as the previous visualizations: GAP as well as spatial smoothing smoothen the loss landscapes.
In conclusion, \emph{averaging feature map points tends to help neural network optimization by smoothing, flattening, and stabilizing the loss landscape}. We observe a similar phenomenon for deterministic NNs. We also provide the Hessian eigenvalue spectrum as shown in \cref{fig:loss-landscape:hes}, and it leads to the same conclusion.

In these experiments, we use MLP incorporating dropout layers with a rate of 50\% as the classifier. Since the dropout is one of the factors that makes MLP underfit the training dataset, we also evaluate MLP without dropouts. Nevertheless, the results still shows that the predictive performance of MLP is worse than that of GAP on the training dataset. Moreover, it severely degrades predictive performance of ResNet on the test dataset.

\subsection{Pre-activation}
\label{sec:revisiting:preact}

\citet{he2016identity} experimentally showed that the pre-activation arrangement, in which the activation $\texttt{ReLU} \circ \texttt{BatchNorm}$ is placed before the convolution, improves the accuracy of ResNet. Since $\gamma$s of most \texttt{BatchNorm}s in CNNs are near-zero \citep{frankle2020training}, \texttt{BatchNorm}s reduce the magnitude of feature maps. 
Constant scaling is a non-trainable \texttt{BatchNorm} with no bias, and it also reduces the magnitude of feature map. 
In \cref{tab:prob}, we show that constant scaling improves predictive performance.
Considering the similarity between \texttt{Prob} with constant scaling and conventional activation, i.e., the similarity between $\texttt{ReLU} \circ \texttt{ConstantScaling}$ and $\texttt{ReLU} \circ \texttt{BatchNorm}$, we find that the pre-activation arrangement improves uncertainty as well as accuracy, because convolutions act as a \texttt{Blur}.

To demonstrate this, we change the post-activation of all layers to pre-activation, and measure the predictive performance. 
\cref{tab:preact} shows the predictive performance of various models with pre-activation. The results suggests that pre-activation improves both accuracy and uncertainty in most cases. For VGG-19, pre-activation significantly degrades accuracy but improves NLL.
In conclusion, they imply that pre-activation is a special case of spatial smoothing. 

\citet{santurkar2018does} argued that \texttt{BatchNorm} helps in optimization by flattening the loss landscape. Likewise, we show that spatial smoothing flattens and smoothens the loss landscapes. It will be interesting to rigorously investigate if \texttt{BatchNorm} helps in ensembling feature maps.

\subsection{ReLU6}
\label{sec:revisiting:relu6}

\texttt{ReLU6} was empirically introduced to improve predictive performance \citep{krizhevsky2010convolutional}. \citet{sandler2018mobilenetv2} used ``\texttt{ReLU6} as the nonlinearity because of its robustness when used with low-precision computation''. In \cref{tab:prob}, we show that  \texttt{ReLU6}s at the end of stages helps to ensemble spatial information by transforming the feature map to Bernoulli distributions.
Since spatial smoothing improves robustness against data corruption, it seems reasonable that \texttt{ReLU6} is robust to low-precision computation. A more abundant investigation is promising future works.

We measure the predictive performance of NNs using all activations as \texttt{ReLU6} instead of \texttt{ReLU}. 
However, in contrast to the results in \cref{tab:prob}, the results are not consistent. We speculate that the reason is that a lot of \texttt{ReLU6}s overly regularize NNs.

\input{resources/tab-preact}

%% file: resources/tab-classifier.tex
\begin{table*}
\vskip 0.1in
\setlength\extrarowheight{2pt} 

\caption{
\textbf{MLP classifier does not overfit training dataset}, i.e., GAP does not regularize NNs.
We provide predictive performance of MC dropout with various classifiers on CIFAR-100. \textsc{Err} is error.
}

\begin{center}
\begin{small}

  \begin{tabular}{cccccccccccc}
    \toprule

    \multirow{3}{*}{\textsc{Model}} & \multirow{3}{*}{\textsc{Classifier}} & \multicolumn{3}{c}{\textsc{Train}} && \multicolumn{3}{c}{\textsc{Test}} \\
    \cline{3-5} \cline{7-9} 
    & & \thead{\textsc{NLL}} & \thead{\textsc{Err}\\(\%)} & \thead{\textsc{ECE}\\(\%)} && \thead{\textsc{NLL}} & \thead{\textsc{Acc}\\(\%)} & \thead{\textsc{ECE}\\(\%)} \\
    \midrule
    \multirow{4}{*}{VGG-16}
    & GAP & 0.0852 & \textbf{0.461} & 6.75 && \textbf{1.030} & \textbf{72.3} & \textbf{3.24} \\
    & MLP & 0.5492 & 13.1 & 13.8 && 1.133 & 68.8 & 3.66 \\
    & GMaxP & \textbf{0.0846} & 0.470 & \textbf{6.67} && 1.050 & 72.2 & 3.60 \\
    & GMedP & 0.0867 & 0.501 & 6.80 && 1.042 & 72.2 & 3.35 \\
    \midrule
    \multirow{4}{*}{VGG-19}
    & GAP & \textbf{0.1825} & \textbf{2.50} & \textbf{10.4} && \textbf{1.035} & \textbf{71.9} & \textbf{1.46} \\
    & MLP & 0.7144 & 17.7 & 14.8 && 1.215 & 67.3 & 6.37 \\
    & GMaxP & 0.1939 & 2.85 & 10.6 && 1.063 & 71.5 & 2.10 \\
    & GMedP & 0.1938 & 2.80 & 10.6 && 1.051 & 71.7 & 1.70 \\
    \midrule
    \multirow{4}{*}{ResNet-18}
    & GAP & 0.0124 & 0.0287 & \textbf{1.19} && \textbf{0.841} & \textbf{77.5} & \textbf{2.92} \\
    & MLP & \textbf{0.0076} & 0.0347 & 7.22 && 1.040 & 74.8 & 9.55 \\
    & GMaxP & 0.0113 & \textbf{0.0233} & 1.41 && 0.905 & 76.3 & 5.23 \\
    & GMedP & 0.0156 & 0.0347 & 1.46 && 0.889 & 76.4 & 5.03 \\
    \midrule
    \multirow{4}{*}{ResNet-50}
    & GAP & \textbf{0.0061} & \textbf{0.0220} & 0.48 && \textbf{0.822} & \textbf{79.1} & 6.63 \\
    & MLP & 0.0071 & 0.0370 & 8.53 && 1.029 & 76.9 & 11.8 \\
    & GMaxP & 0.0074 & 0.0313 & 1.09 && 0.887 & 77.2 & \textbf{5.67} \\
    & GMedP & 0.0053 & 0.0287 & \textbf{0.47} && 0.849 & 78.5 & 6.29 \\
    \bottomrule
  \end{tabular}

\end{small}
\end{center}
\vskip 0.1in

\label{tab:classifier}
\end{table*}

%% file: resources/fig-classifier-robustness.tex
\begin{figure*}
\centering

\centering
\centering
\includegraphics[width=0.30\textwidth]{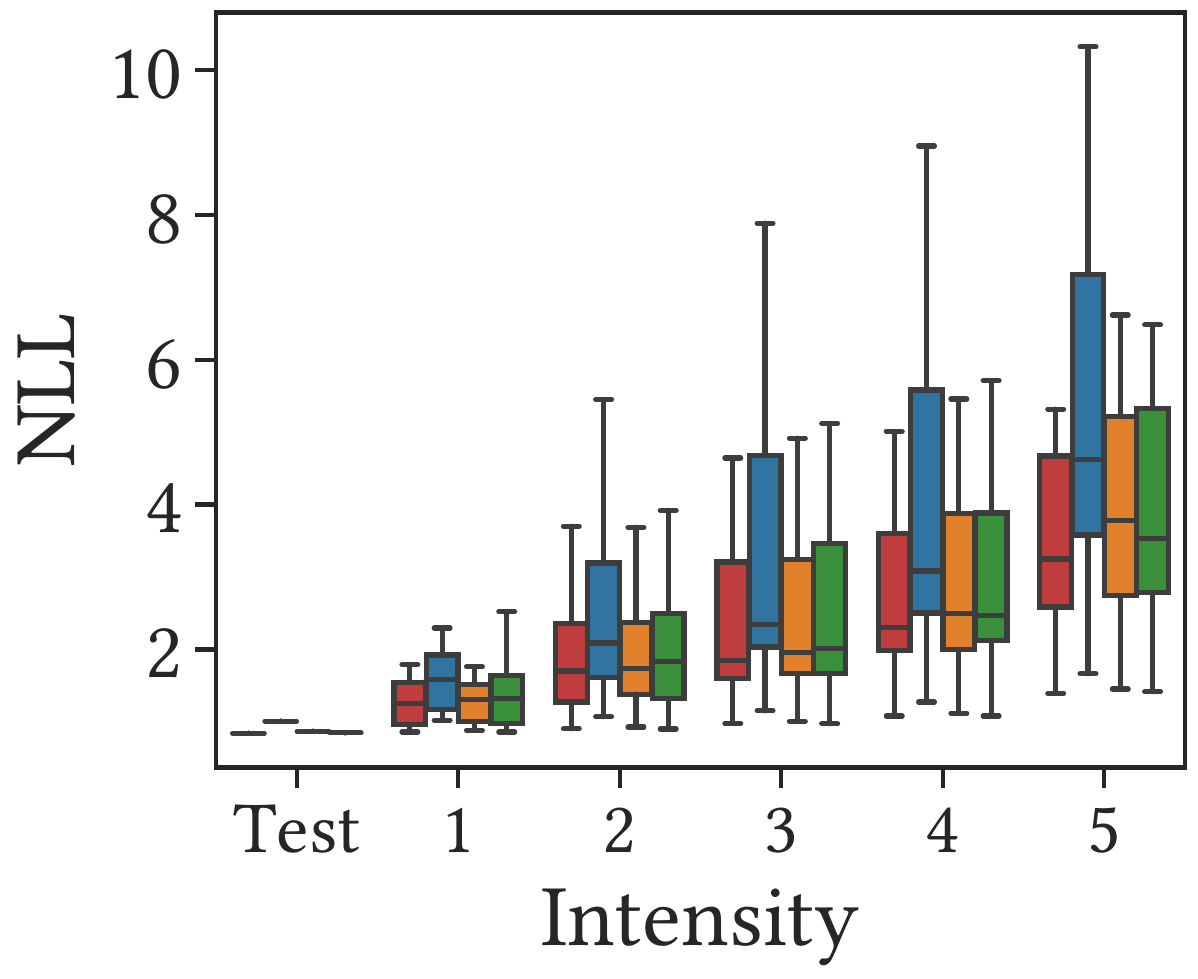}
\hspace{1pt}
\includegraphics[width=0.30\textwidth]{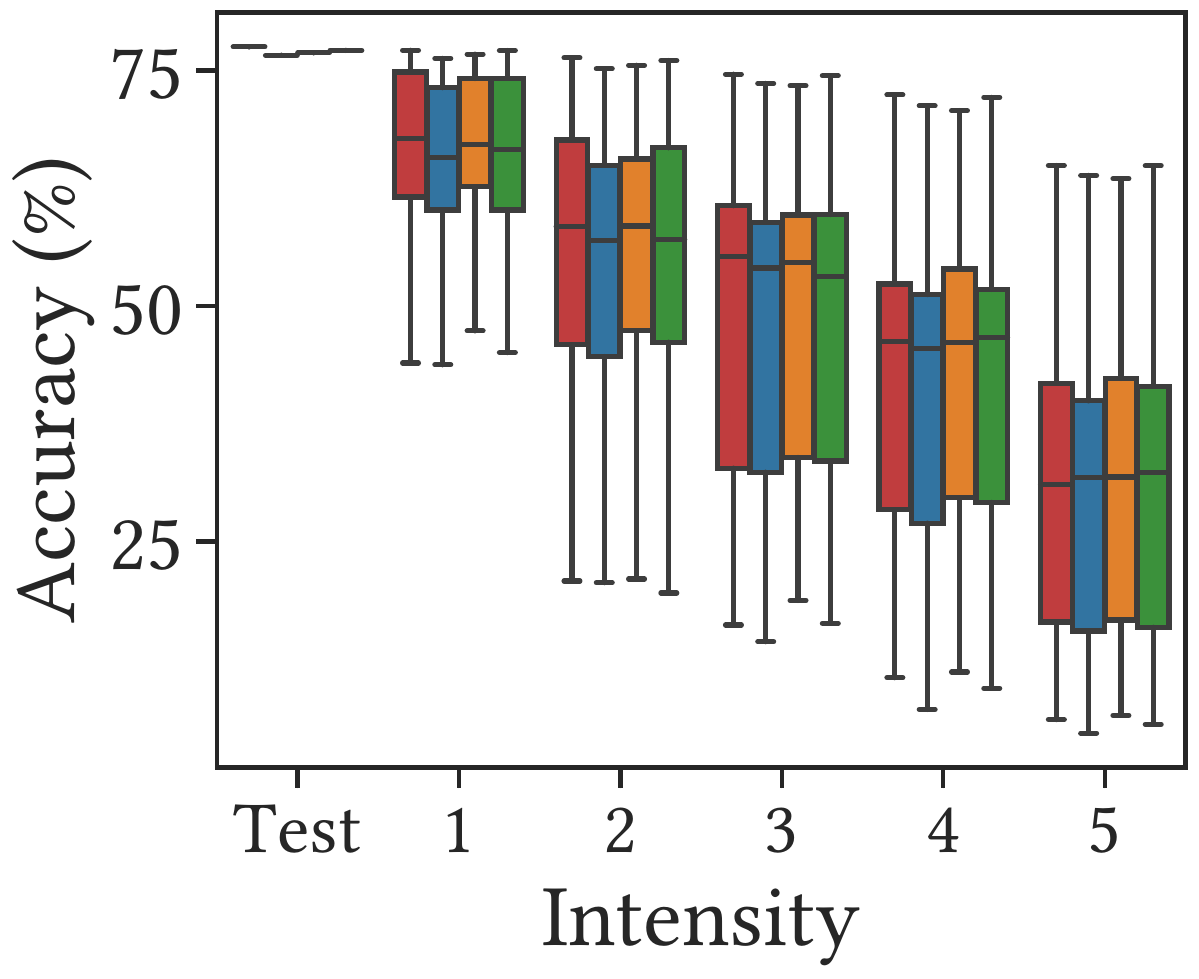}
\hspace{1pt}
\includegraphics[width=0.30\textwidth]{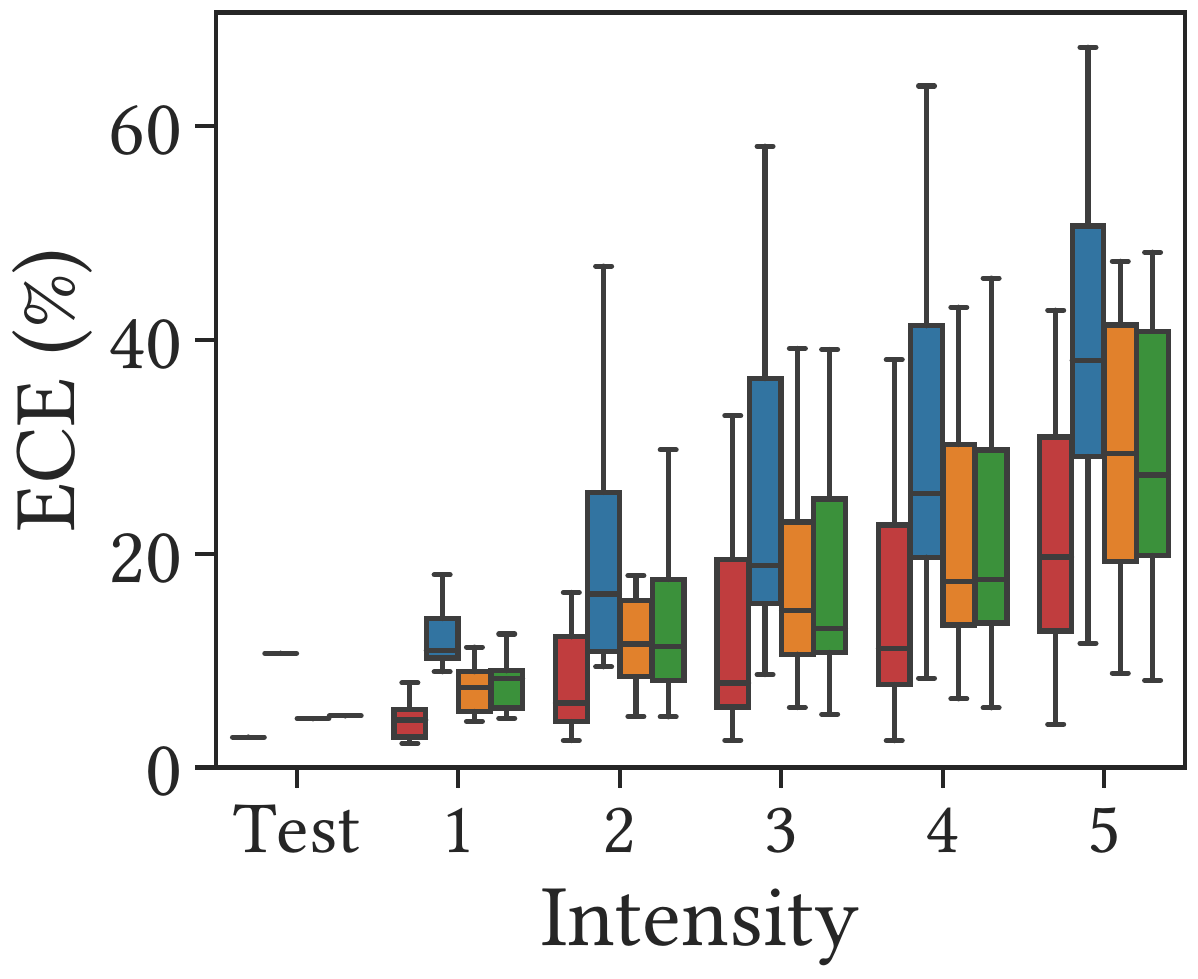}

\vspace{3pt}
\centering
\includegraphics[height=0.025\textheight]{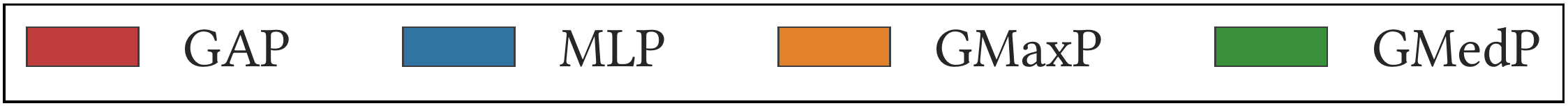}

\vspace{5pt}

\includegraphics[width=0.30\textwidth]{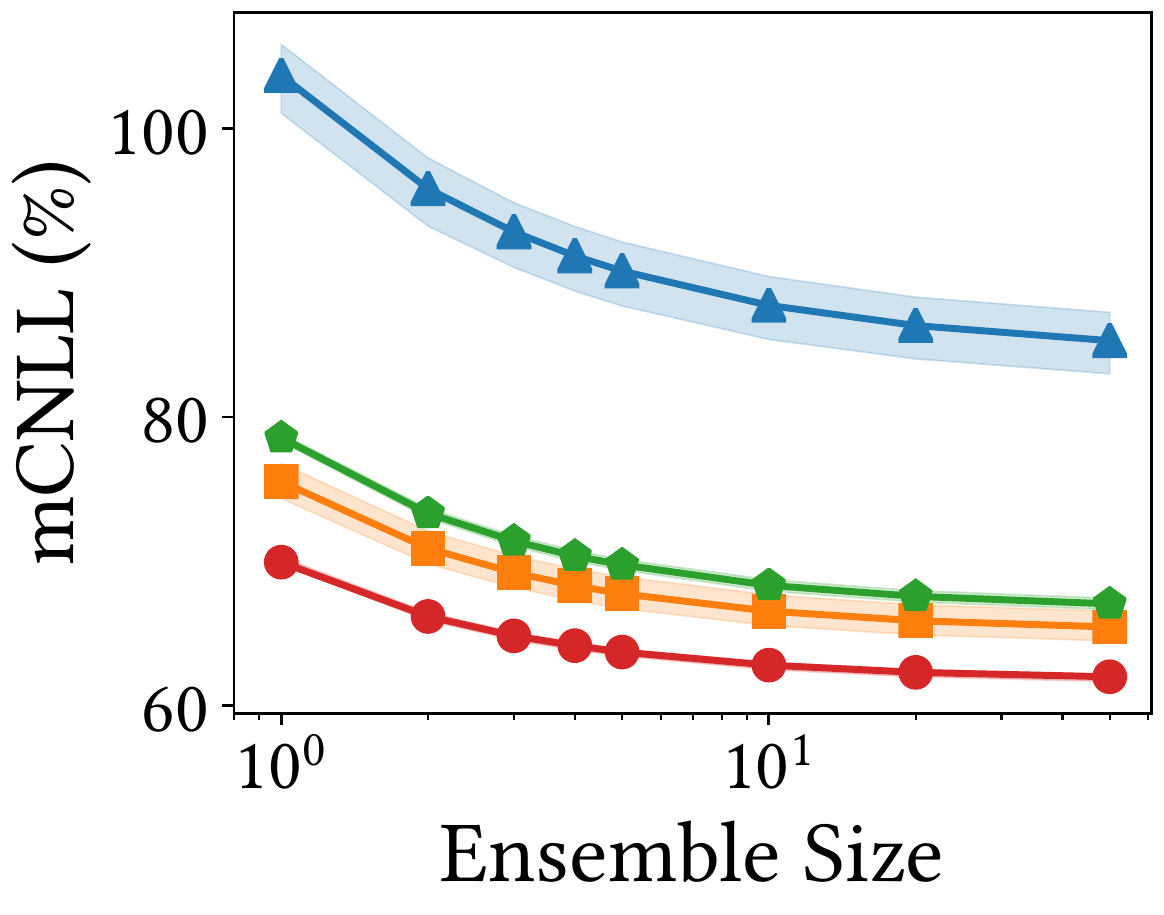}
\hspace{1pt}
\includegraphics[width=0.30\textwidth]{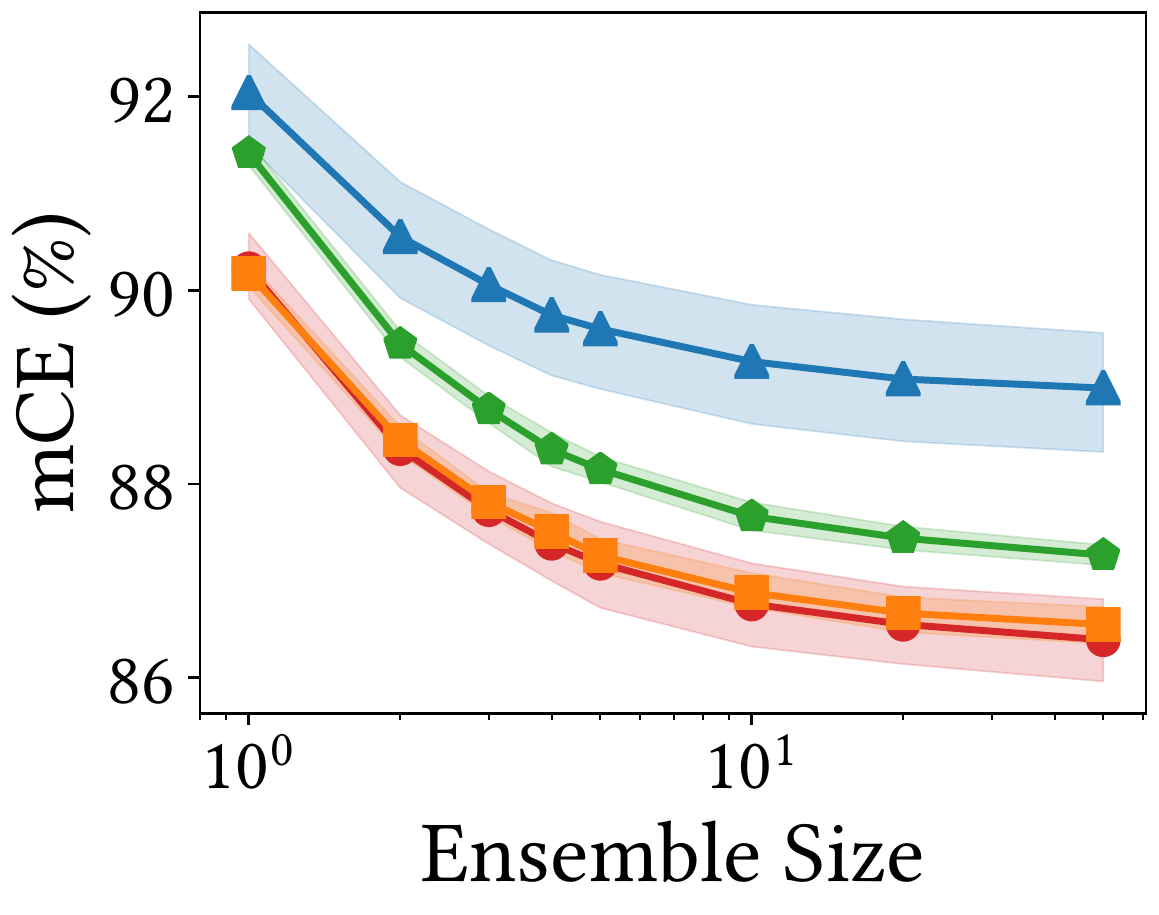}
\hspace{1pt}
\includegraphics[width=0.30\textwidth]{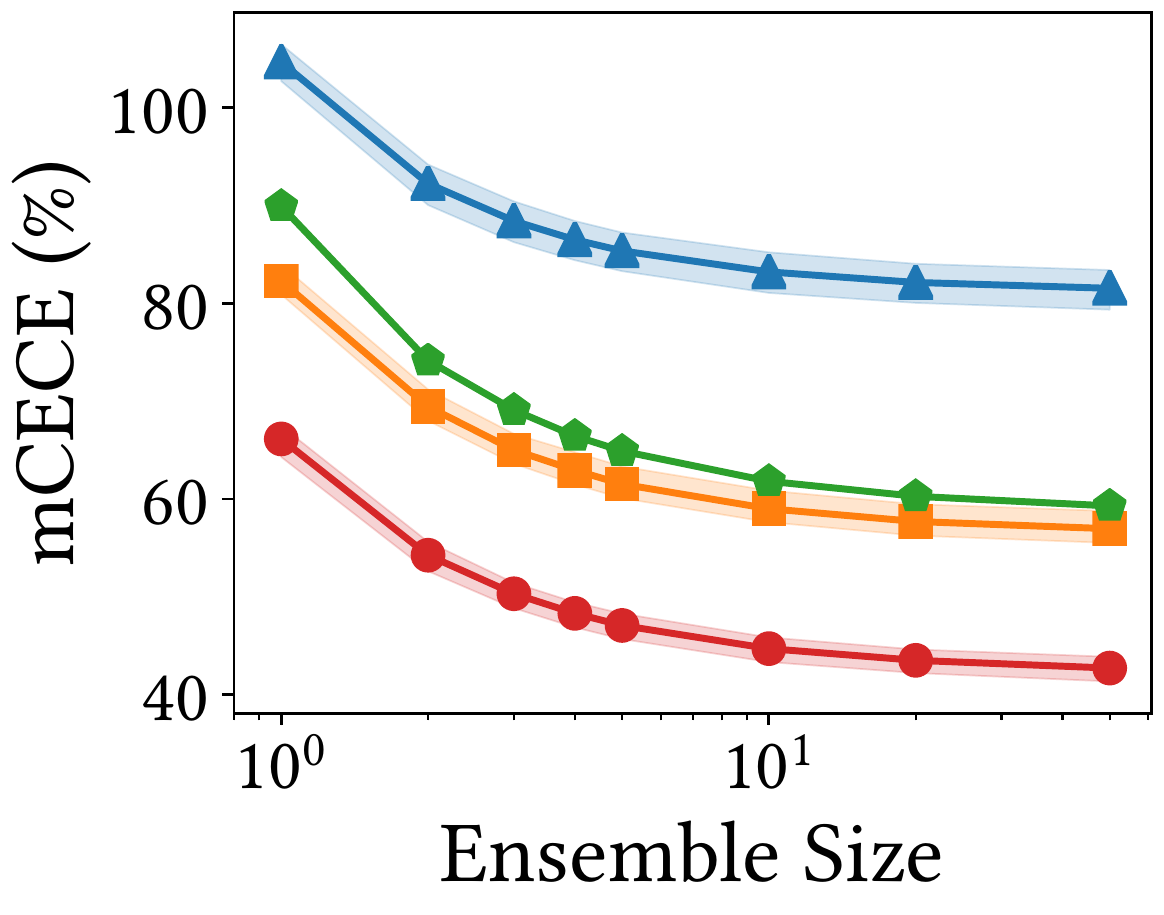}

\vspace{3pt}
\centering
\includegraphics[height=0.025\textheight]{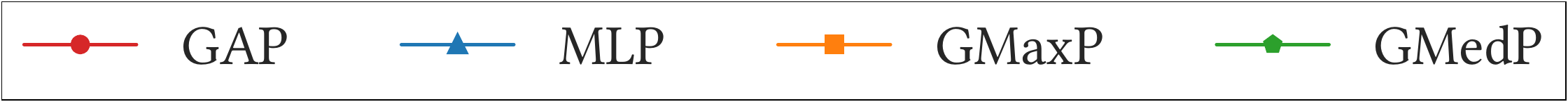}

\caption{
\textbf{GAP classifier improves not only the predictive performance on clean dataset but also the robustness}.
We measure the predictive performance of ResNet-18 using MC dropout with classifiers on CIFAR-100-C. 
}
\label{fig:classifier:robustness}
\end{figure*}

%% file: resources/fig-loss-landscape-sequence.tex
\begin{figure*}
\vskip 0.05in
\begin{center}

\begin{subfigure}[b]{\textwidth}
\centering
\includegraphics[width=0.17\textwidth]{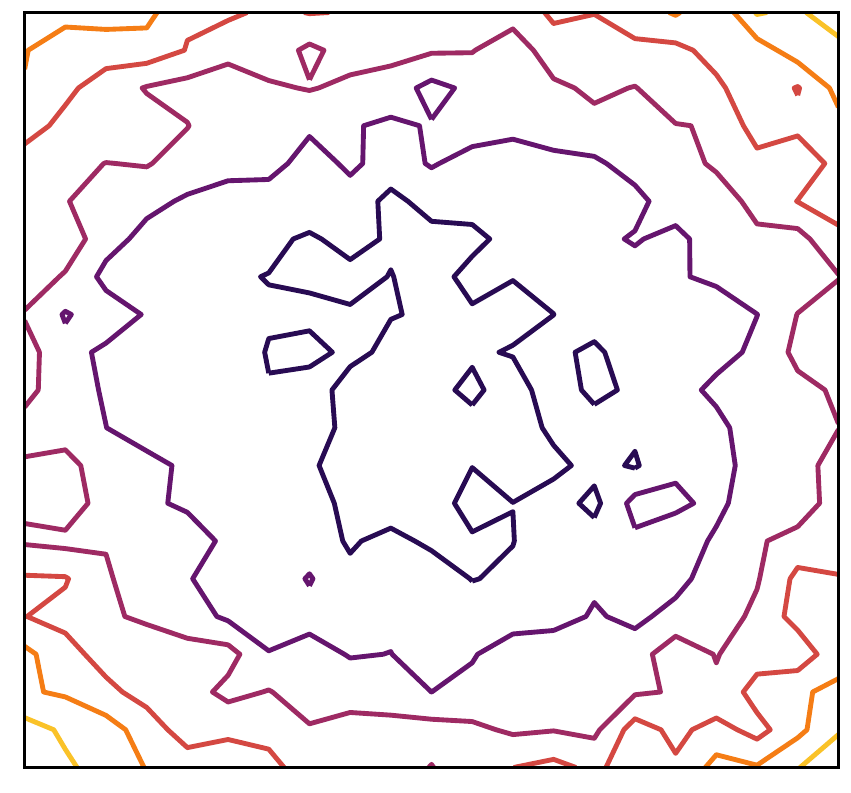}
\includegraphics[width=0.17\textwidth]{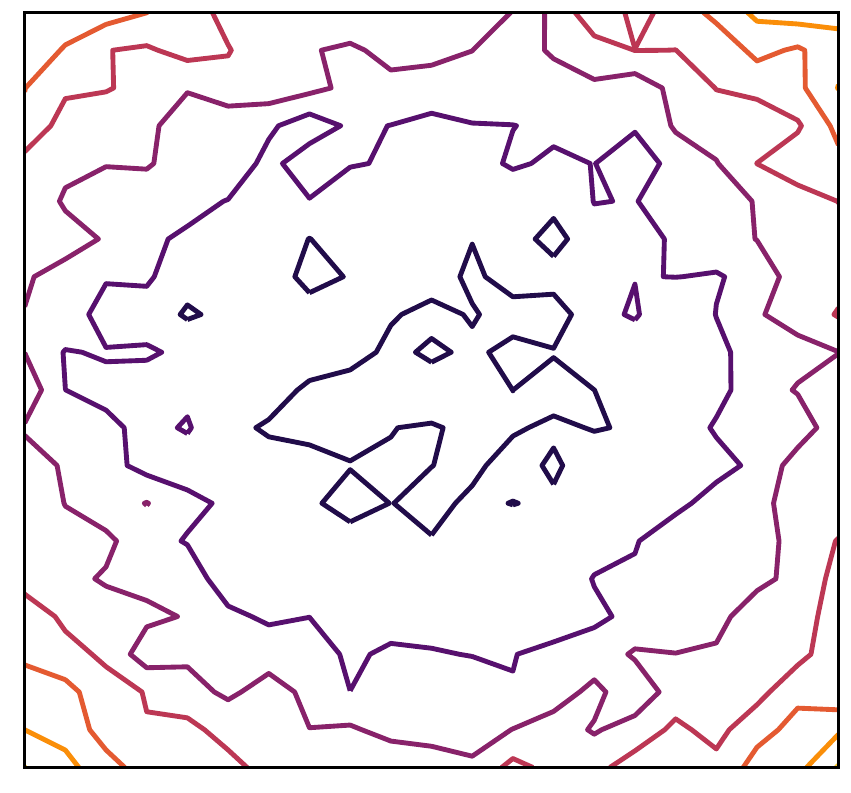}
\includegraphics[width=0.17\textwidth]{resources/figures/losssurface/animated/resnet_mcdo_18_20210420_170127_2.pdf}
\includegraphics[width=0.17\textwidth]{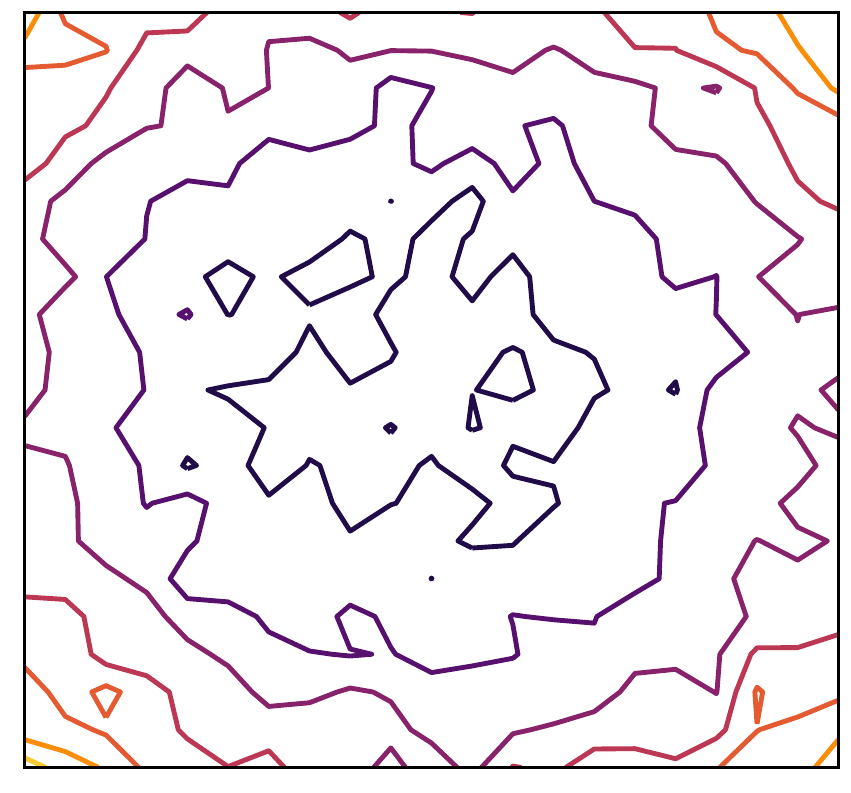}
\includegraphics[width=0.17\textwidth]{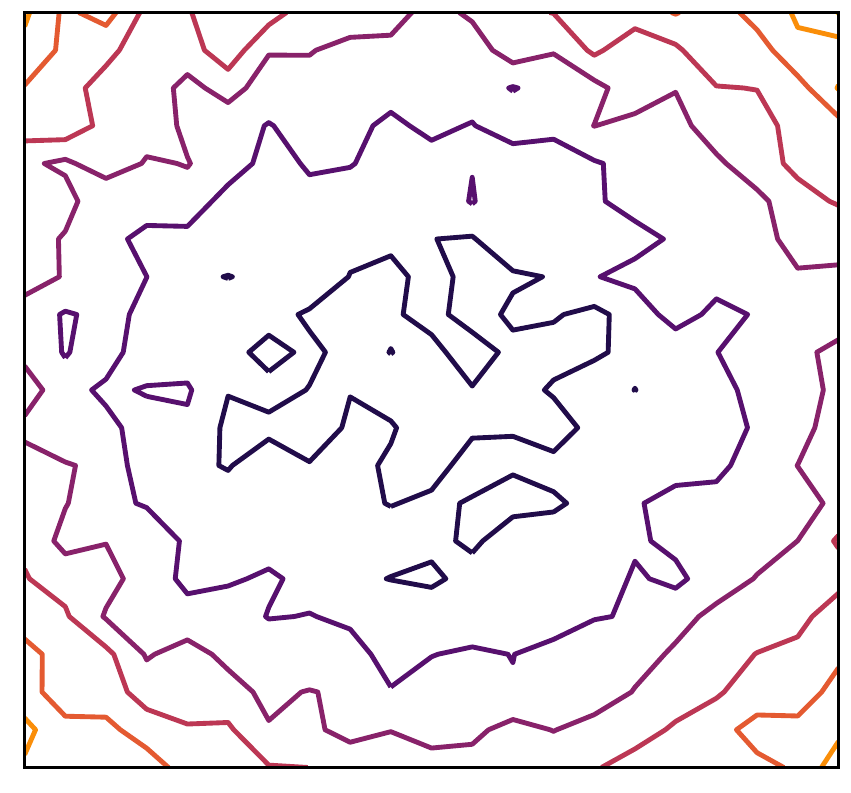}
\caption{MLP classifier}
\label{fig:loss-landscape:animated:mlp}
\end{subfigure}

\vspace{5px}

\begin{subfigure}[b]{\textwidth}
\centering
\includegraphics[width=0.17\textwidth]{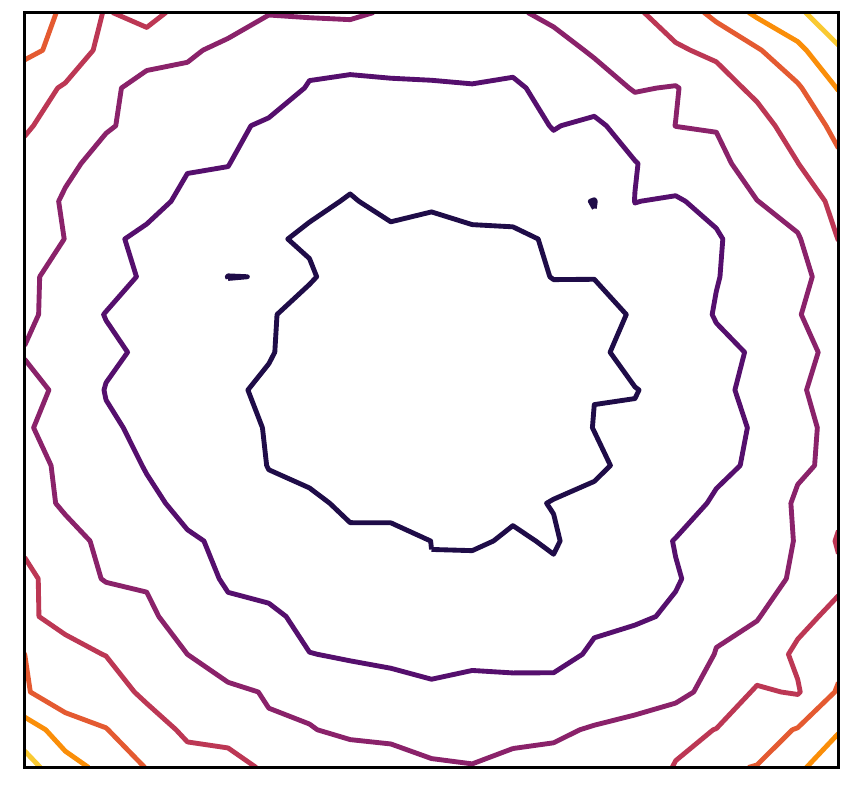}
\includegraphics[width=0.17\textwidth]{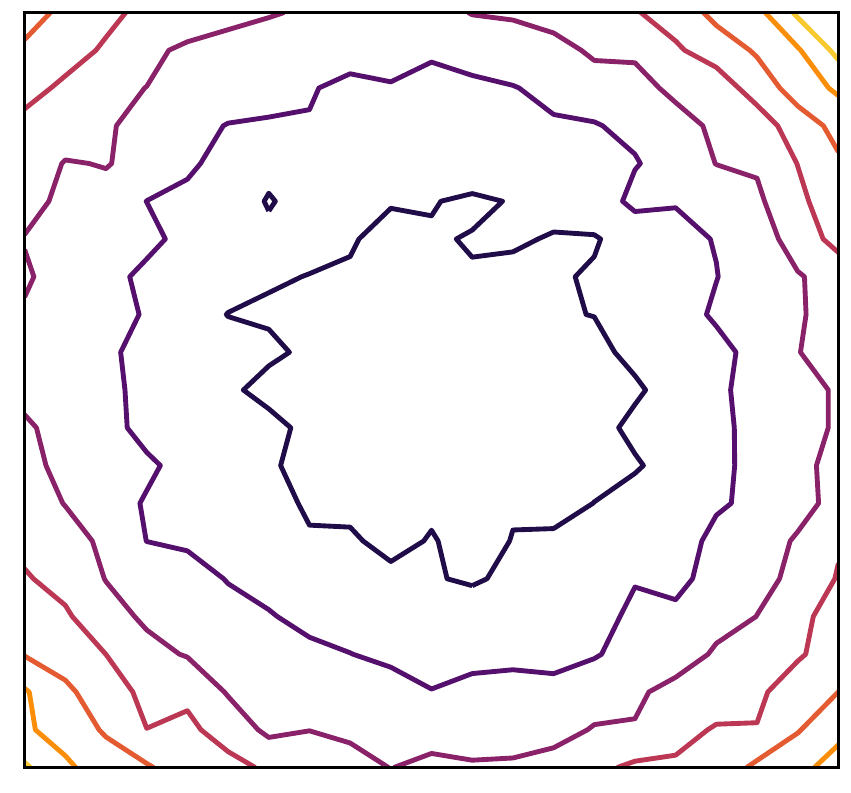}
\includegraphics[width=0.17\textwidth]{resources/figures/losssurface/animated/resnet_mcdo_18_20210301_170248_2.pdf}
\includegraphics[width=0.17\textwidth]{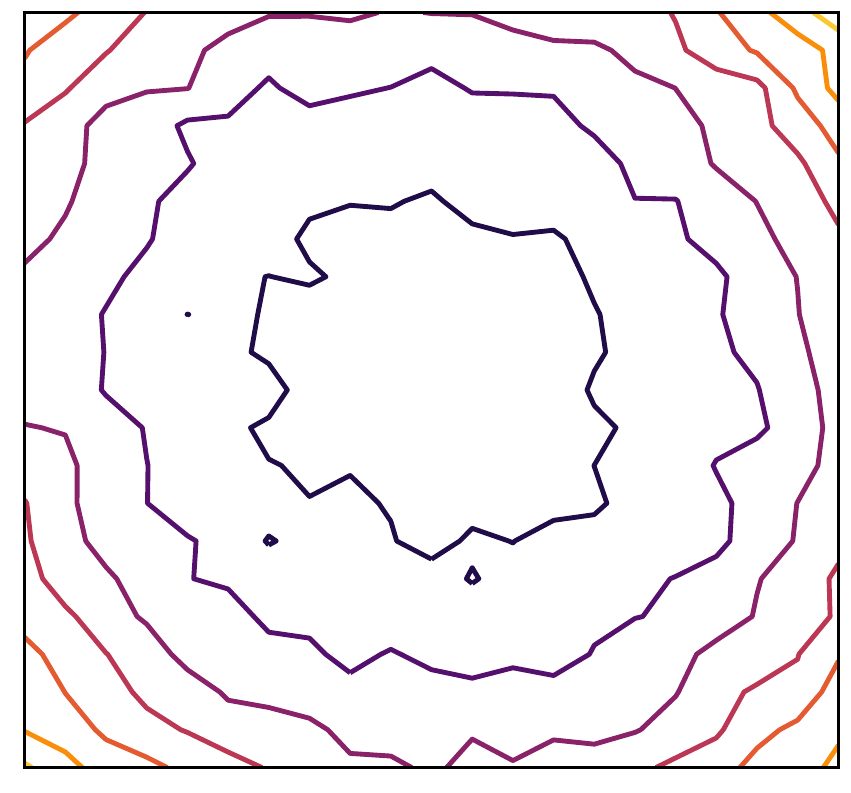}
\includegraphics[width=0.17\textwidth]{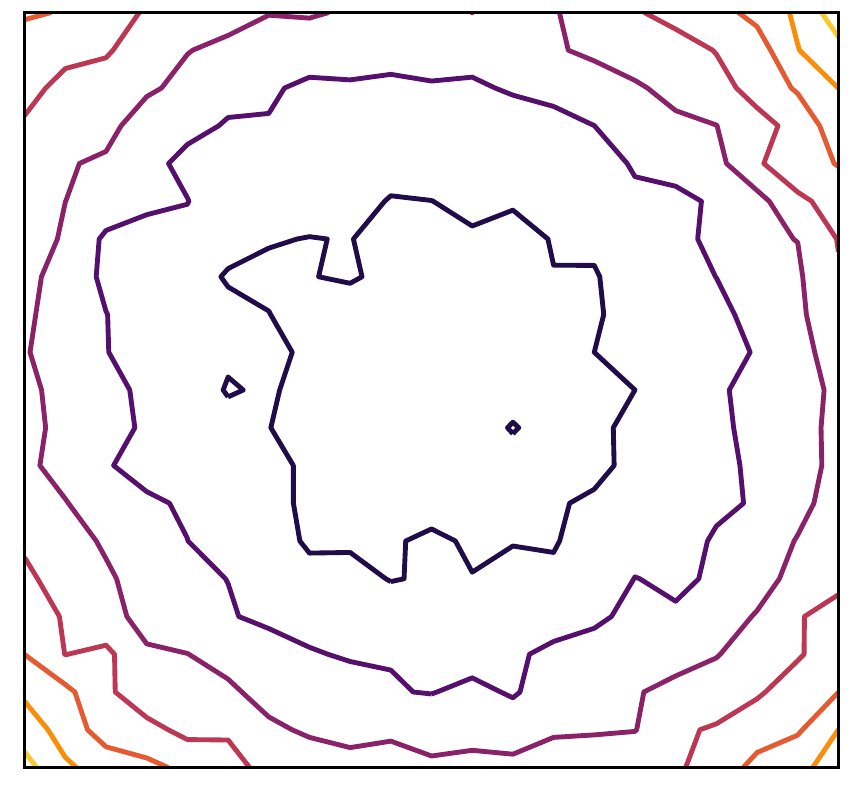}
\caption{GAP classifier}
\label{fig:loss-landscape:animated:gap}
\end{subfigure}

\vspace{5px}

\begin{subfigure}[b]{\textwidth}
\centering
\includegraphics[width=0.17\textwidth]{resources/figures/losssurface/animated/resnet_mcdo_smoothing_18_20210218_130655_0.pdf}
\includegraphics[width=0.17\textwidth]{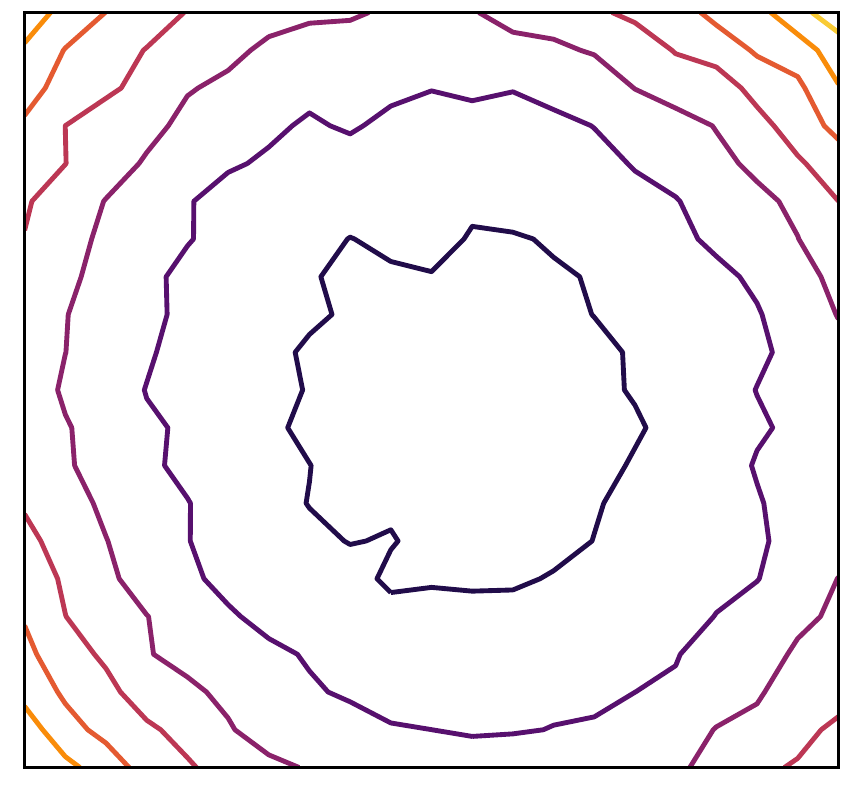}
\includegraphics[width=0.17\textwidth]{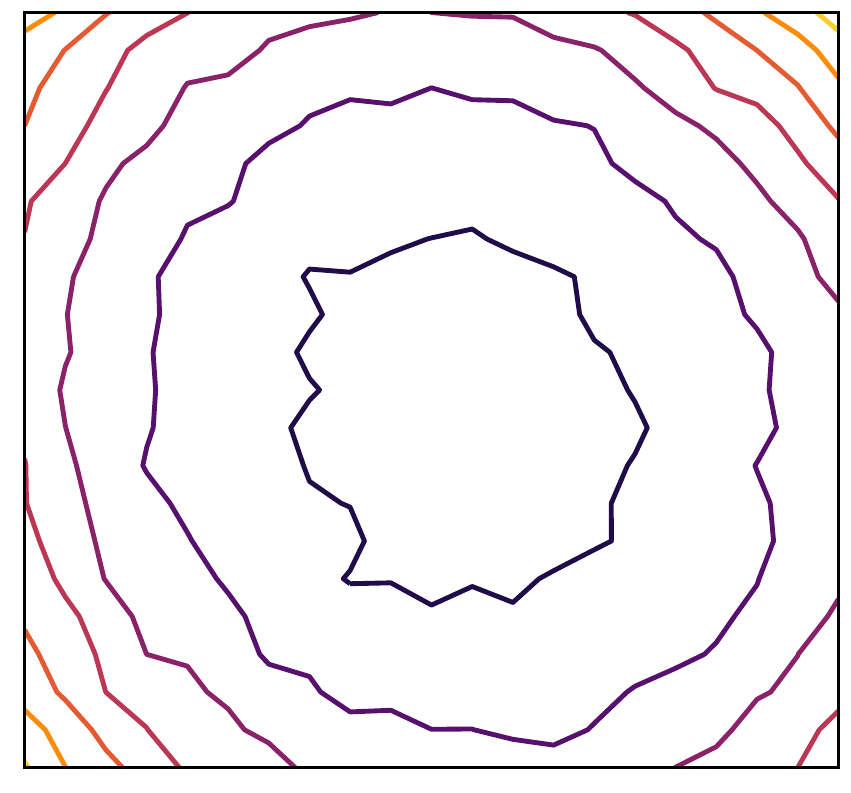}
\includegraphics[width=0.17\textwidth]{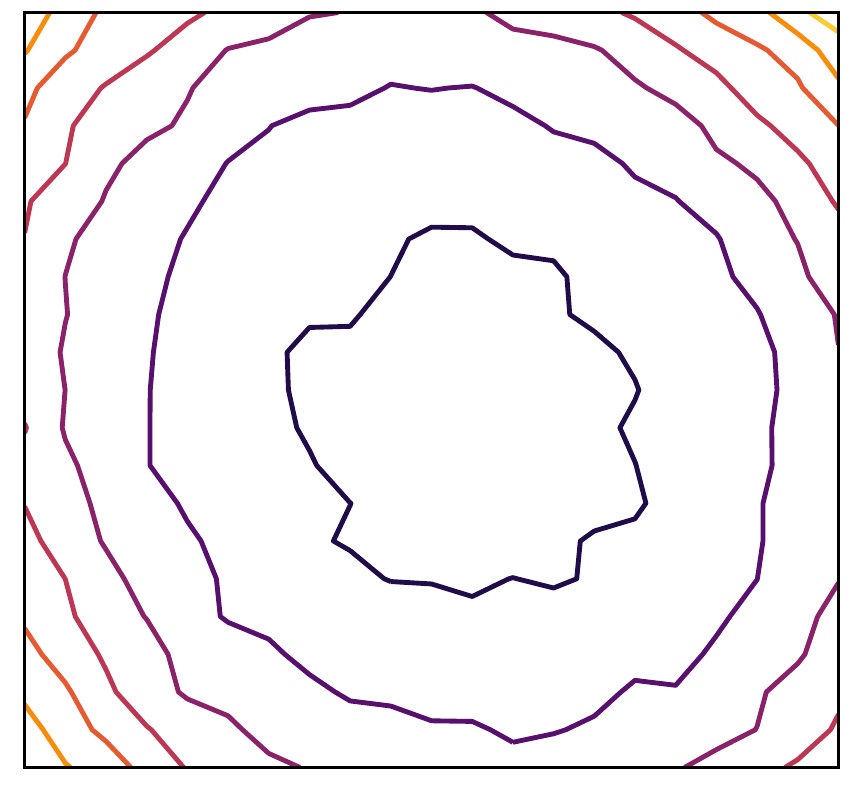}
\includegraphics[width=0.17\textwidth]{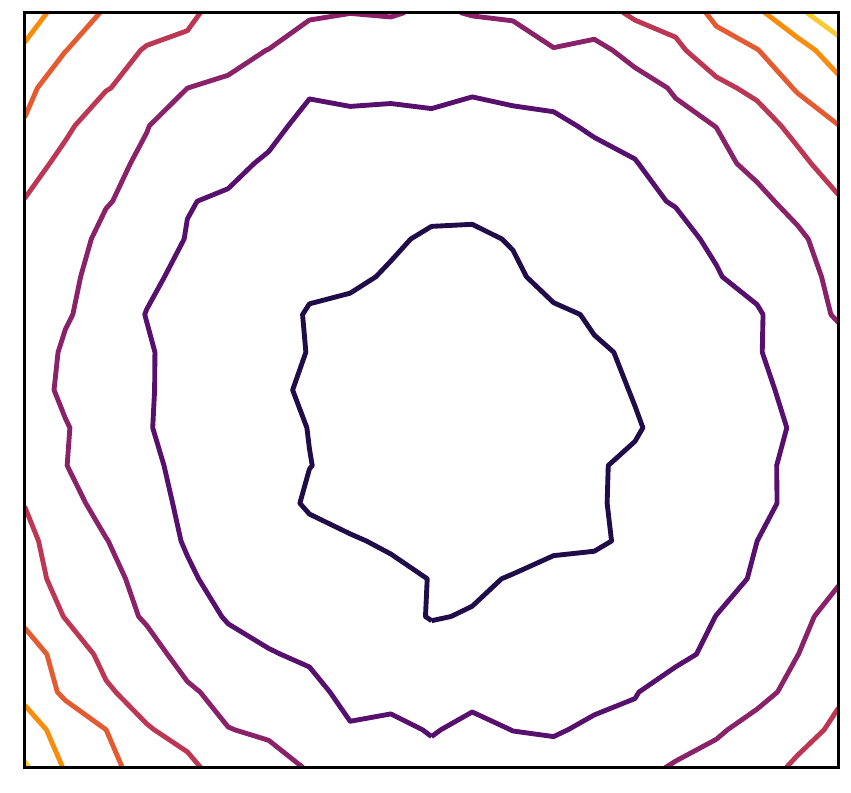}
\caption{GAP classifier + \texttt{Smooth}}
\label{fig:loss-landscape:animated:smooth}
\end{subfigure}

\end{center}

\vskip -0.05in

\caption{
\textbf{GAP and spatial smoothing smoothen the loss landscapes}.
We visualize the loss landscape sequences of ResNet-18 with MC dropout on CIFAR-100. Although each sequence shares the bases, it fluctuates due to the randomness.
}
\label{fig:loss-landscape:animated}
\vskip -0.1in
\end{figure*}

%% file: resources/tab-preact.tex
\begin{table*}
\vskip 0.1in
\setlength\extrarowheight{3pt} 

\caption{
\textbf{Pre-activation arrangement improves uncertainty as well as accuracy}.
We measure the predictive performance of models with pre-activation arrangement on CIFAR-100.
}

\begin{center}
\begin{small}

  \begin{tabular}{cccccccccccc}

    \toprule

    \textsc{Model} & \textsc{MC dropout} & \textsc{Pre-act} & \textsc{NLL} & \thead{\textsc{Acc}\\(\%)} & \thead{\textsc{ECE}\\(\%)} \\
    \midrule
    \multirow{4}{*}{\thead{VGG-16}}
    & $\cdot$ & $\cdot$ & 2.047 \color{Gray}(-0.000) & 71.6 \color{Gray}(+0.0) & 19.2 \color{Gray}(-0.0) \\
    & $\cdot$ & \checkmark & 1.827 \color{Green}(-0.219) & \textbf{72.5 \color{Green}(+0.9)} & 19.8 \color{Red}(+0.6) \\
    \cline{2-6}
    & \checkmark & $\cdot$ & 1.133 \color{Gray}(-0.000) & 68.8 \color{Gray}(+0.0) & 3.66 \color{Gray}(-0.00) \\
    & \checkmark & \checkmark & \textbf{1.036 \color{Green}(-0.096)} & 71.7 \color{Green}(+2.9) & \textbf{3.55 \color{Green}(-0.11)} \\
    \midrule
    \multirow{4}{*}{\thead{VGG-19}}
    & $\cdot$ & $\cdot$ & 2.016 \color{Gray}(-0.000) & 67.6 \color{Gray}(+0.0) & 21.2 \color{Gray}(-0.0) \\
    & $\cdot$ & \checkmark & 1.799 \color{Green}(-0.217) & 64.4 \color{Red}(-3.2) & 17.2 \color{Green}(-4.0) \\
    \cline{2-6}
    & \checkmark & $\cdot$ & 1.215 \color{Gray}(-0.000) & 67.3 \color{Gray}(+0.0) & 6.37 \color{Gray}(-0.00) \\
    & \checkmark & \checkmark & \textbf{1.084 \color{Green}(-0.131)} & \textbf{70.1 \color{Green}(+3.7)} & \textbf{4.23 \color{Green}(-2.14)} \\
    \midrule
    \multirow{4}{*}{\thead{ResNet-18}}
    & $\cdot$ & $\cdot$ & 0.983 \color{Gray}(-0.000) & 77.1 \color{Gray}(+0.0) & 7.75 \color{Gray}(-0.00) \\
    & $\cdot$ & \checkmark & 0.934 \color{Green}(-0.049) & 77.6 \color{Green}(+0.5) & 8.04 \color{Red}(+0.29) \\
    \cline{2-6}
    & \checkmark & $\cdot$ & 0.937 \color{Gray}(-0.000) & 76.9 \color{Gray}(+0.0) & \textbf{5.11 \color{Gray}(-0.00)} \\
    & \checkmark & \checkmark & \textbf{0.872 \color{Green}(-0.065)} & \textbf{77.6 \color{Green}(+0.7)} & 5.53 \color{Red}(+0.42) \\
    \midrule
    \multirow{4}{*}{\thead{ResNet-50}}
    & $\cdot$ & $\cdot$ & 0.880 \color{Gray}(-0.000) & 79.0 \color{Gray}(+0.0) & 8.35 \color{Gray}(-0.00) \\
    & $\cdot$ & \checkmark & 0.870 \color{Green}(-0.010) & 79.4 \color{Green}(+0.4) & 8.27 \color{Green}(-0.08) \\
    \cline{2-6}
    & \checkmark & $\cdot$ & 0.831 \color{Gray}(-0.000) & 78.6 \color{Gray}(+0.0) & \textbf{6.06 \color{Gray}(-0.00)} \\
    & \checkmark & \checkmark & \textbf{0.819 \color{Green}(-0.012)} & \textbf{79.5 \color{Green}(+0.9)} & 6.29 \color{Red}(+0.23) \\
    \bottomrule
  \end{tabular}

\end{small}
\end{center}
\vskip 0.1in

\label{tab:preact}
\end{table*}

%% file: appendix/extended-analysis.tex
\section{Extended Analysis of How Spatial Smoothing Works}\label{sec:extended-analysis}

This section provides further explanation of the analysis in \cref{sec:spatial-smoothing:how}.

\subsection{Neighboring Feature Maps in CNNs Are Similar}

Although our work is based on the assumption that images are spatially consistent, we provide one explanation of the spatial consistency of feature maps: even if input images are spatially inconsistent, feature maps are consistent.

Consider a single-layer CNN with one channel: 
\begin{align}
y_i 
&= \left[ \bm{w} * \bm{x} \right]_{i} = \sum_{l=1}^{k} w_{l} x_{i - l + 1}
\end{align}
where $*$ is convolution with a kernel of size $k$, $\bm{y}$ is feature map output, $\bm{w}$ is kernel weight, and $\bm{x}$ is input \emph{random variable}. Then, the covariance of two neighboring points is:
\begin{align}
\text{Cov}(y_{i}, y_{i + 1}) 
&= \text{Cov} (\sum_{l=1}^{k} w_{l} x_{i - l + 1}, \, \sum_{m=1}^{k} w_{m} x_{i - m + 2}) \\
&= \sum_{l=1}^{k} \sum_{m=1}^{k} w_{l} w_{m} \, \text{Cov} (x_{i - l + 1}, x_{i - m + 2}) \\
&= \sum_{l=1}^{k - 1} w_{l} w_{l + 1} \, \sigma^{2} (x_{i - l + 2}) + \cdots \label{eq:consistency:proof}
\end{align}
where $\sigma^{2}(x_{i - l + 1})$ is the variance of $x_{i - l + 1}$. Therefore, $\text{Cov}(\bm{y}_{i}, \bm{y}_{i + 1})$ is non-zero for randomly initialized weights. If $\bm{x}$ is \emph{iid}, i.e., $\text{Cov}(x_i, x_j) = \delta_{ij} \sigma^{2}(x_i) $ where $\delta_{ij}$ is the Kronecker delta, the remainders in \cref{eq:consistency:proof} vanish.

\begin{figure*}

\centering

\vskip 0.1in

\begin{subfigure}[b]{0.31\textwidth}
\centering
\includegraphics[width=0.83\textwidth]{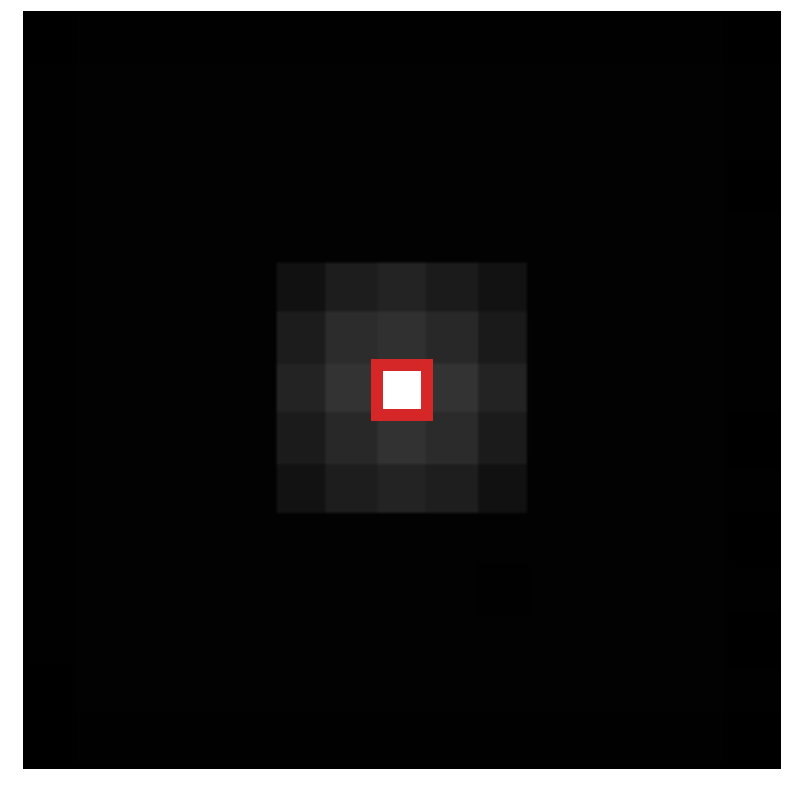}
\caption{single-layer CNN}
\label{fig:consistency:1-layer}
\end{subfigure}
\hspace{8pt}
\centering
\begin{subfigure}[b]{0.31\textwidth}
\centering
\includegraphics[width=0.83\textwidth]{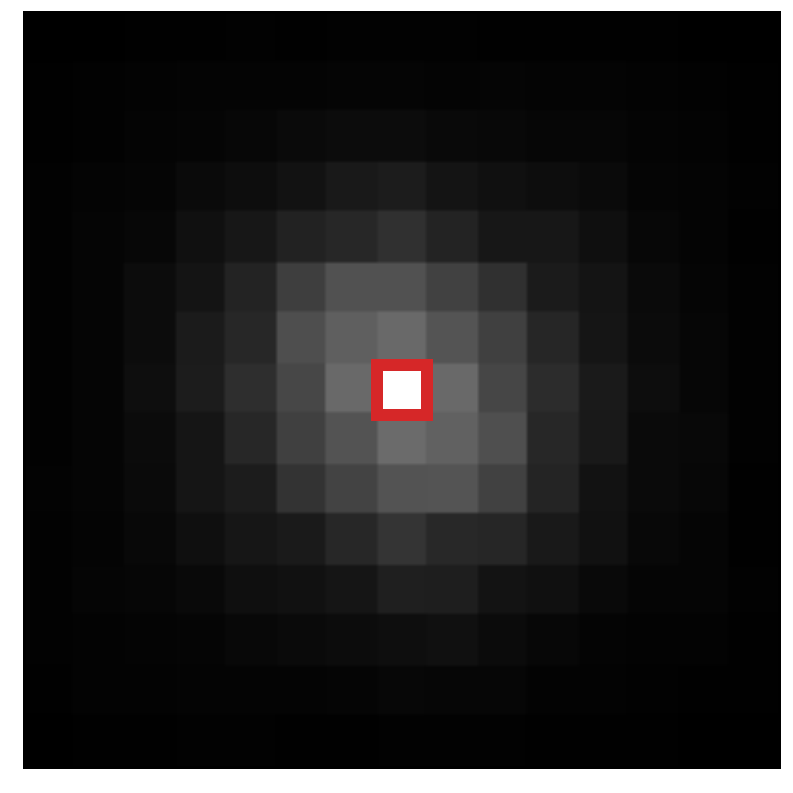}
\caption{five-layer CNN with \texttt{ReLU}}
\label{fig:consistency:5-layer}
\end{subfigure}
     
\vspace{-2pt}
\caption{
\textbf{Neighboring feature map points in CNNs are similar, even if input values are \emph{iid}}. We provide covariances of feature map points with respect to the center feature map (in the red square). Input values are Gaussian random noise. 
\emph{Left: }
A single convolutional layer correlates the target feature map with another feature map  that is 3 pixels away, since the kernel size is 3$\times$3.
\emph{Right: }
A deep CNN more strongly correlates neighboring feature maps.
}
\label{fig:training-phase-ensemble}

\end{figure*}

For example, the covariance of two neighboring feature map points in a CNN with a kernel size of 3 is non-zero, i.e., 
\begin{align}
	\text{Cov}(y_{1}, y_{2}) = w_1 w_2 \, \sigma^2(x_2) + w_2 w_3 \, \sigma^2(x_3)
\end{align}
since the terms for $i \neq j$ vanish and only the terms for $i = j$ remain. Therefore, the neighboring feature maps $y_{1}$ and $y_{2}$ are correlated.

\paragraph{Experiment. }

To demonstrate the spatial consistency of feature maps empirically, we provide feature map covariances of randomly initialized single-layer CNN and five-layer CNN with \texttt{ReLU} nonlinearity. In this experiment, the input values are Gaussian random noises. As shown in \cref{fig:consistency:1-layer}, one convolutional layer correlates neighboring feature map points. \cref{fig:consistency:5-layer} shows that multiple convolutional layers correlate one feature map with distant feature maps. Moreover, the feature maps in deep CNNs have a stronger relationship with neighboring feature maps.

\subsection{Ensembles Filter High-Frequency Signals}\label{sec:extended-analysis:fourier}

Consider a square importance-weight matrix for an ensemble that does not change size. The sum of the matrix columns is one, and all the elements are greater than zero. Therefore, the matrix can be expressed using \texttt{Softmax}, and a \texttt{Softmax}-normalized matrix is a low-pass filter \citep{wang2021scaling}.

\paragraph{Experiment. }

Since blur filter (\texttt{Blur}) is low-pass filter, probabilistic spatial smoothing (\texttt{Prob}--\texttt{Blur}) is also low-pass filter. In \cref{fig:fourier:amplitude}, at the end of the stage 1, we show that MC dropout adds high-frequency noise to feature maps, and spatial smoothing effectively removes it. 
We observe the same phenomena at other stages. 

In addition, \cref{fig:fourier:robustness} shows that CNNs are vulnerable to high-frequency random noise. Interestingly, it also shows that CNNs are robust against noise with frequencies from $0.6 \pi$ to $0.8 \pi$, corresponding to approximately 3 pixel periods. Since the receptive fields of convolutions are 3$\times$3, the noise with a period smaller than the size is averaged out by convolutions. For the same reason, convolutions are particularly vulnerable against the noise with a frequency of $0.3 \pi$, corresponding to a period of 6 pixel.

\subsection{Randomness Sharpens the Loss Landscapes, and Ensembles Smoothen Them}\label{sec:extended-analysis:dropout}

\begin{figure*}
\centering

\vskip 0.2in

\raisebox{0pt}[\dimexpr\height-0.6\baselineskip\relax]{
\begin{subfigure}[b]{0.40\textwidth}
\centering
\includegraphics[width=0.80\textwidth]{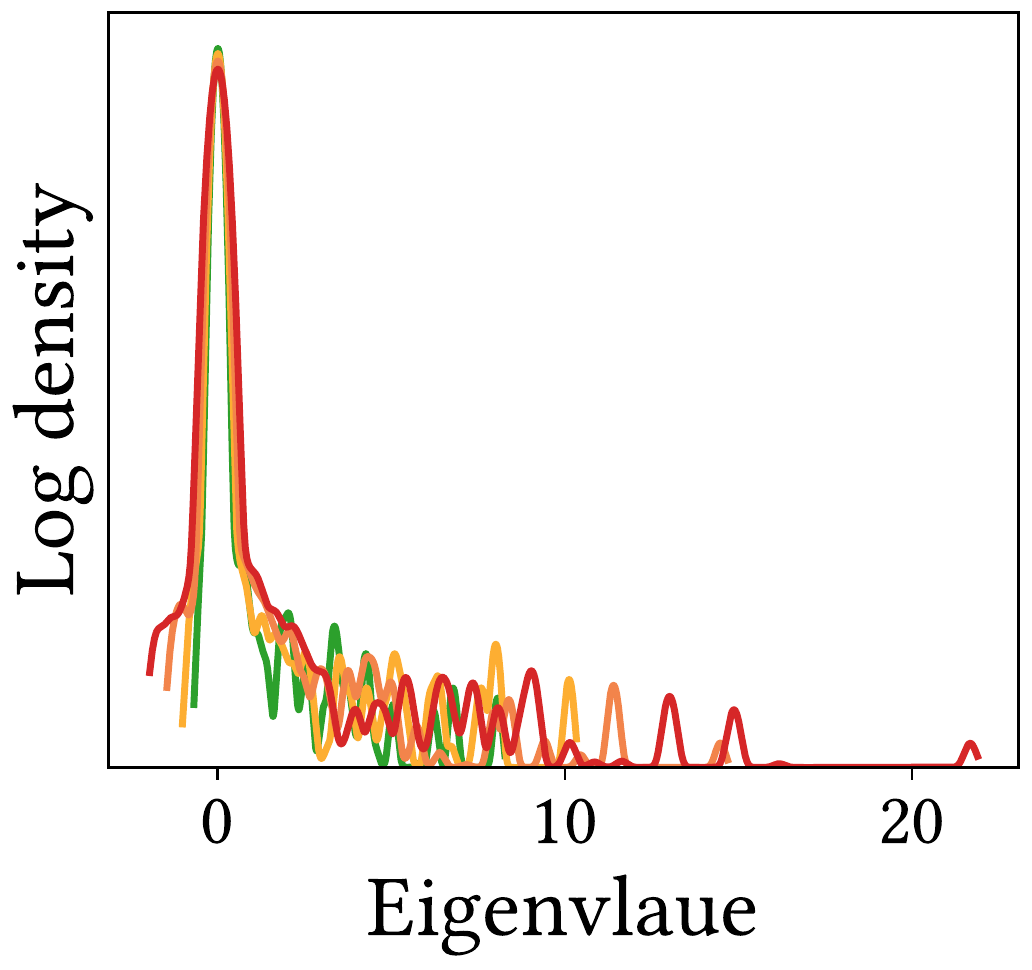}
\end{subfigure}
\begin{subfigure}[b]{0.415\textwidth}
\centering
\includegraphics[width=0.80\textwidth]{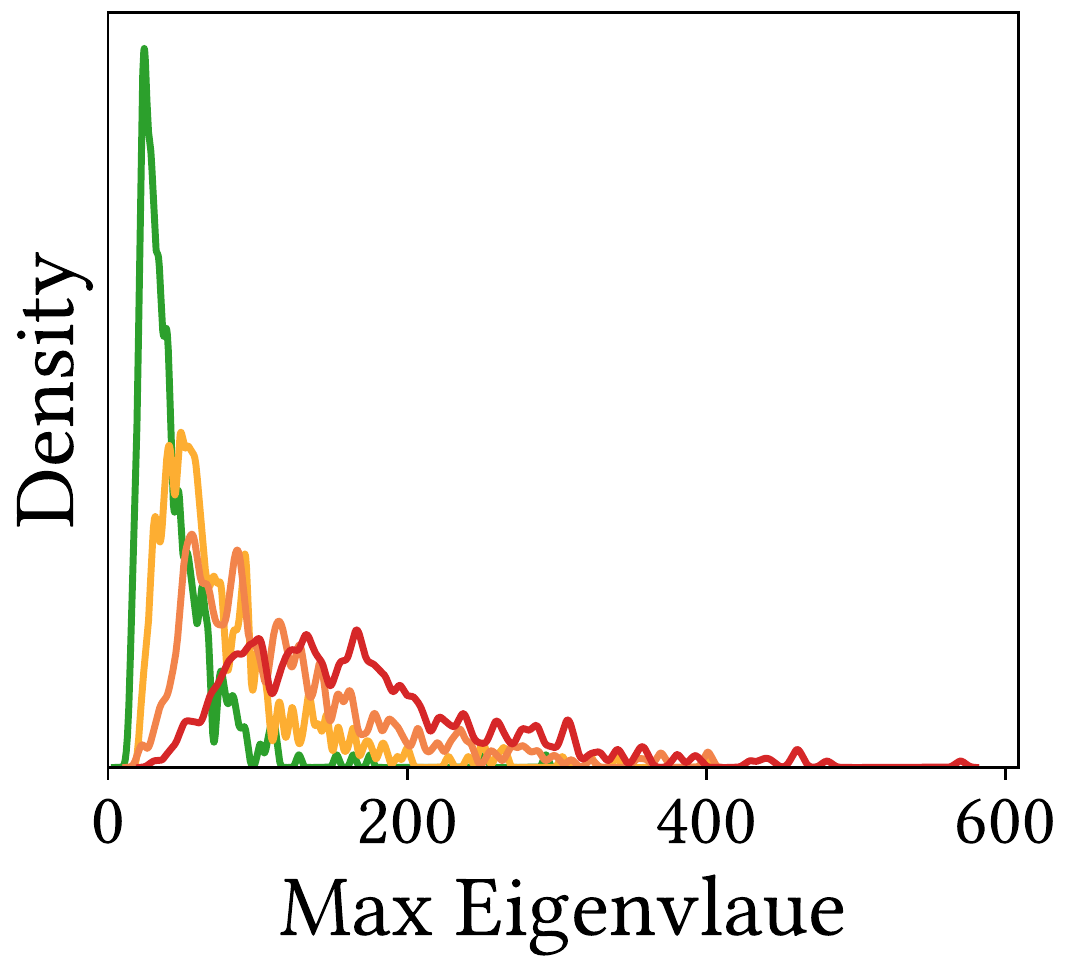}
\end{subfigure}
}

\vspace{2pt}

\centering
\includegraphics[height=0.027\textheight]{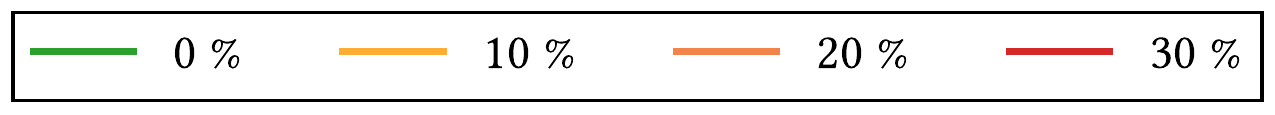}

\caption{
\textbf{Randomness due to MC dropout sharpens the loss function}.
We provide Hessian eigenvalue spectra (\emph{left}) and Hessian max eigenvalue spectra (\emph{right}) of ResNet-18 on CIFAR-100.
}
\label{fig:dropout:hes}
\end{figure*}

Ws show that the randomness of NN predictions hinder and destabilize NN training because it causes the loss landscape and its gradient to fluctuate from moment to moment. In other words, the randomness, such as dropout, sharpens the loss landscape. Since ensemble effectively reduces the randomness of predictions, it smoothens the loss landscape. Below, we prove these claims.

\paragraph{Definition of sharpness. }

We start with \citet{foret2020sharpness}'s definition of sharpness:
\begin{align}
	\text{sharpness}_{\rho} = \max_{\vert\vert \bm{\epsilon} \vert\vert \leq \rho} \mathcal{L} (\bm{w} + \bm{\epsilon}) - \mathcal{L} (\bm{w})
\label{eq:sharpness:definition}
\end{align}
where $\mathcal{L}$ is NLL loss on a training dataset, $\bm{w}$ is NN weight, $\bm{\epsilon}$ is small weight perturbation, and $\rho$ is neighborhood radius. However, they used this expression for deterministic optimization tasks, and this expression is not for random variables $\bm{\epsilon}$; the maximum value of a random variable loss $\max_{\vert\vert \bm{\epsilon} \vert\vert \leq \rho} \mathcal{L}(\bm{w} + \bm{\epsilon})$ does not appropriately represent the properties of the random variable in many cases. For example, the maximum value of a Gaussian random variable is infinity, but we cannot observe that infinity in practice. 

To address this issue, we replace ``the maximum value of the loss random variable'' with ``the expected value of sufficiently large losses'' as follows:
\begin{align}
\max_{\vert\vert \bm{\epsilon} \vert\vert \leq \rho} \mathcal{L}(\bm{w} + \bm{\epsilon}) 
\rightarrow 
\mathbb{E}\left[  \max(\mathcal{L} (\bm{w} + \bm{\epsilon}), \mathcal{L}(\bm{w})) \ \Big\vert \ \vert\vert \bm{\epsilon} \vert\vert \leq \rho  \right]
\end{align}
where $\mathbb{E}[ \ \cdot \ \big\vert \ \vert\vert \bm{\epsilon} \vert\vert \leq \rho ]$ is expected value under the constraint $\vert\vert \bm{\epsilon} \vert\vert \leq \rho$. 
Then, the \emph{expected} sharpness is:
\begin{align}
\begin{aligned}
{}&\mathbb{E}[\text{sharpness}_{\rho}] = \\
&\;\;\;\;\;\; \mathbb{E}\left[ \max(\mathcal{L} (\bm{w} + \bm{\epsilon}), \mathcal{L}(\bm{w})) - \mathcal{L} (\bm{w}) \ \Big\vert \ \vert\vert \bm{\epsilon} \vert\vert \leq \rho \right]
\end{aligned}
\end{align}
so $\mathbb{E}[\text{sharpness}_{\rho}] \geq 0$ by definition. Therefore, as the magnitude of $\bm{\epsilon}$---and dropout rate for MC dropout---increases, the sharpness increases.

This expression can be regarded as the difference between ``the large neighborhood losses'' and ``the average loss'' when $\mathcal{L}(\bm{w} + \bm{\epsilon})$ is a Gaussian random variable, i.e., $\mathbb{E}[\mathcal{L} (\bm{w} + \bm{\epsilon})] \simeq \mathbb{E}[\mathcal{L} (\bm{w})] + \mathbb{E}[\bm{\epsilon}^{T} \nabla \mathcal{L} (\bm{w})] \simeq \mathbb{E}[\mathcal{L} (\bm{w})] = \mathcal{L} (\bm{w})$. In other words, this expression connects the sharpness and instability of the loss landscapes in probabilistic NN settings:
\begin{align}
\begin{aligned}
\mathbb{E}[\text{sharpness}_{\rho}]
\simeq {}& \underbrace{\mathbb{E}\left[  \max(\mathcal{L} (\bm{w} + \bm{\epsilon}), \mathcal{L}(\bm{w})) \ \Big\vert \ \vert\vert \bm{\epsilon} \vert\vert \leq \rho  \right]}_{\text{Expected lower-bounded loss in neighborhood}} \\
{}& - \underbrace{\mathbb{E}\left[\mathcal{L} (\bm{w} + \bm{\epsilon}) \ \Big\vert \ \vert\vert \bm{\epsilon} \vert\vert \leq \rho \right]}_{\text{Expected loss}}	
\end{aligned}
\end{align}

\paragraph{Expected sharpness is proportional to the variance of loss. }

Let $\rho$ be sufficiently large ($\rho \gg 1$), i.e., we use weak constraints for weight randomness $\bm{\epsilon}$ at fixed $\bm{w}$. 
If $\mathcal{L}$ is a Gaussian random variable, the expected value of $\max(\mathcal{L} (\bm{w} + \bm{\epsilon}), \mathcal{L} (\bm{w}))$ is the expected value of Rectified Gaussian distribution:
\begin{align}
\mathbb{E} [\max(\mathcal{L} (\bm{w} + \bm{\epsilon}), \mathcal{L} (\bm{w}))] = \mathbb{E}[\mathcal{L} (\bm{w})] + c \mathbb{V}[\mathcal{L} (\bm{w} + \bm{\epsilon})]^{\sfrac{1}{2}}
\end{align}
where $c$ is a positive constant. Therefore, the expected value of sharpness is proportional to the standard deviation of the loss: 
\begin{align}
\mathbb{E}[\text{sharpness}] 
&= \mathbb{E}[\max(\mathcal{L} (\bm{w} + \bm{\epsilon}), \mathcal{L} (\bm{w})) - \mathcal{L}(\bm{w})] \\
&= \mathbb{E}[\mathcal{L} (\bm{w})] + c \mathbb{V}[\mathcal{L} (\bm{w} + \bm{\epsilon})]^{\sfrac{1}{2}} - \mathbb{E}[ \mathcal{L} (\bm{w}) ] \\
&= c \mathbb{V}[\mathcal{L} (\bm{w} + \bm{\epsilon})]^{\sfrac{1}{2}}
\end{align}
In conclusion, the expected sharpness is proportional to the variance of loss random variable.

\begin{figure*}

\centering
\begin{subfigure}[b]{0.43\textwidth}
\centering

\vskip 0.2in

\includegraphics[width=0.80\textwidth]{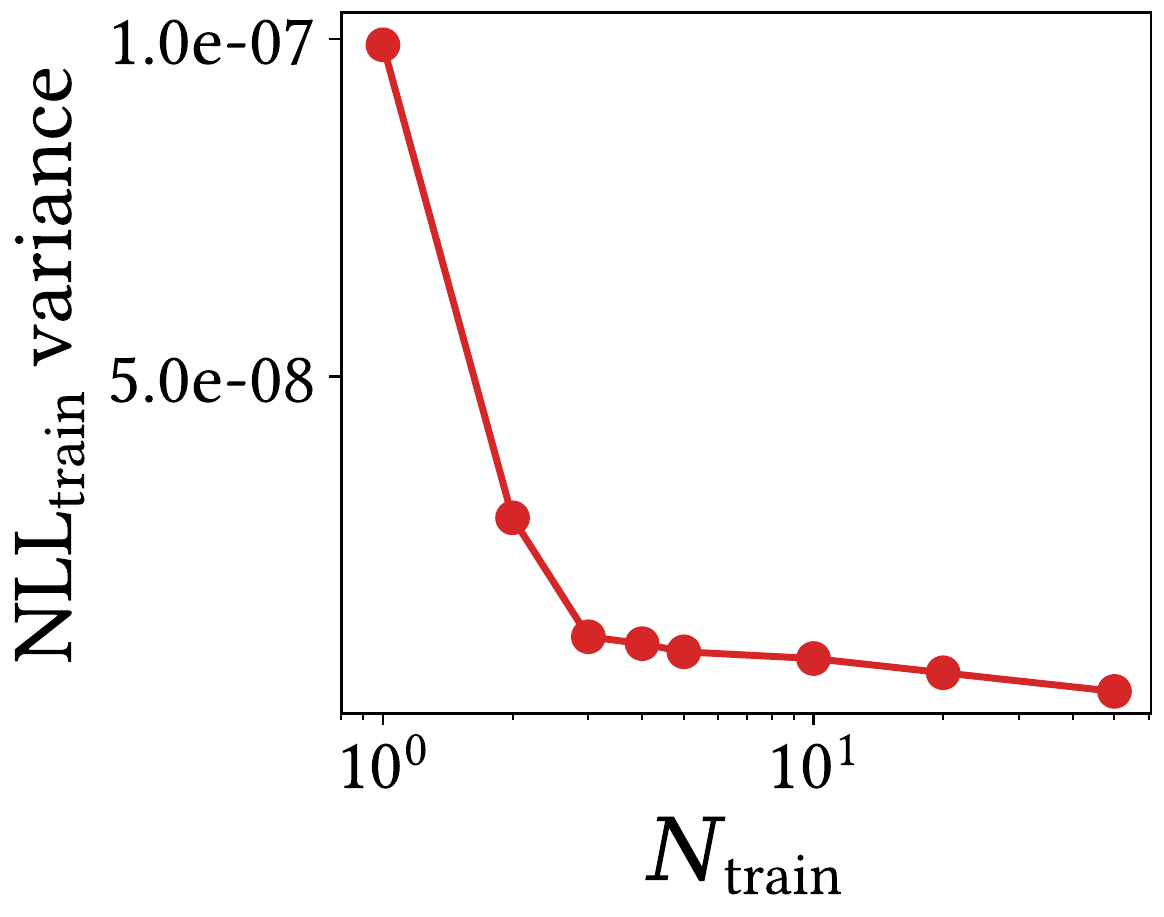}
\caption{$\mathbb{V} \left[ \mathcal{L} \right]$ for ensemble size on training dataset}
\label{fig:training-phase-ensemble:variance}
\end{subfigure}
\centering
\begin{subfigure}[b]{0.431\textwidth}
\centering
\includegraphics[width=0.80\textwidth]{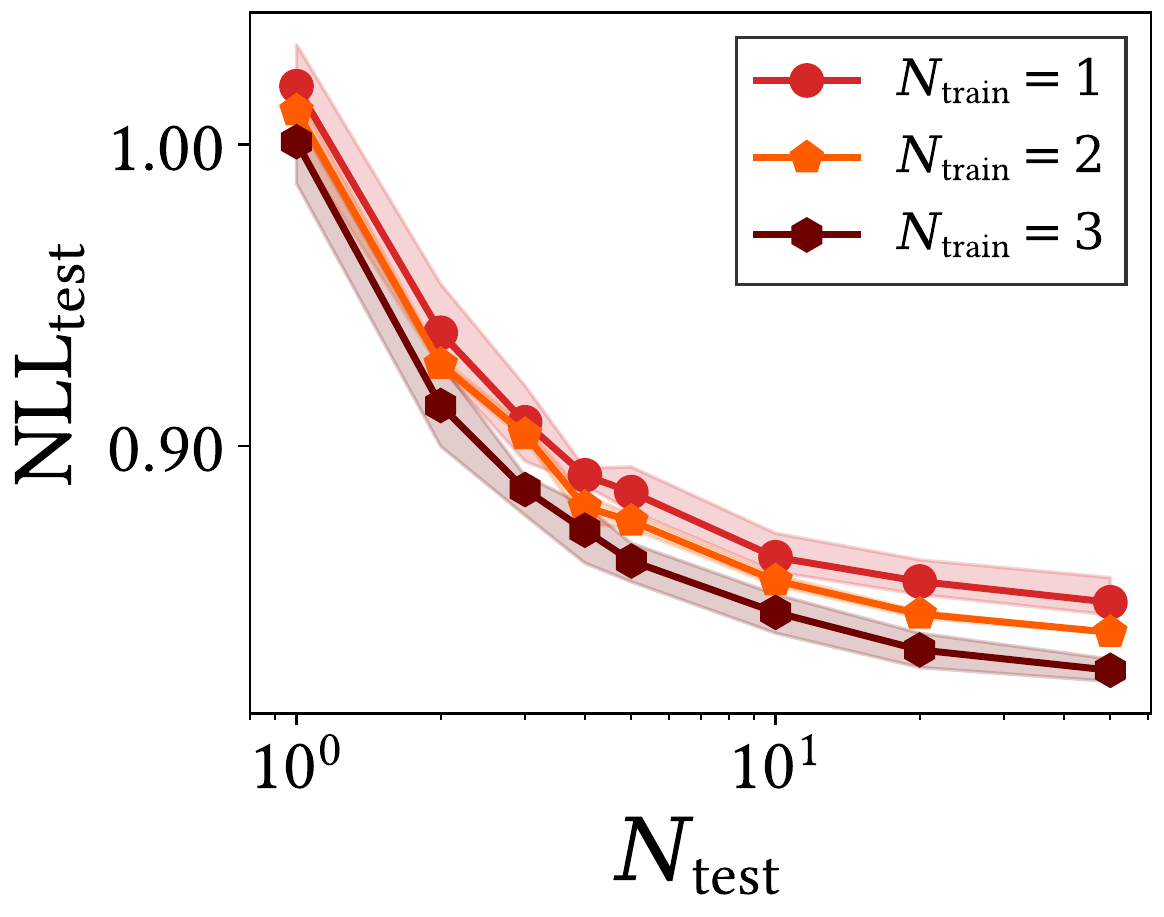}
\caption{NLL for ensemble size on test dataset}
\label{fig:training-phase-ensemble:performance}
\end{subfigure}
     
\vspace{-2pt}
\caption{
\textbf{Training phase ensemble helps NN learn strong representation.}
\emph{Left: }
The variance of NLL ($\mathbb{V} \left[ \mathcal{L} \right]$) on training dataset is inversely proportional to the ensemble size for large $N_\text{train}$. See \cref{eq:sharpness:variance}.
\emph{Right: }
Training phase ensemble improves the predictive performance on test dataset.
}
\label{fig:training-phase-ensemble}

\vskip -0.05in

\end{figure*}

\paragraph{Variance of losses is inversely proportional to the ensemble size. }

Let $p_{i} \in \left( 0, 1 \right]$ be a confidence of one NN prediction, and $\bar{p}^{(N)}$ be a confidence of $N$ ensemble, i.e., $\bar{p}^{(N)} = \frac{1}{N} \sum_{i=1}^{N} p_{i}$. Then, the variance of the NLL loss is:
\begin{align} 
\mathbb{V} \left[ \mathcal{L} \right] \label{eq:sharpness:proof}
&= \mathbb{V} \left[ \frac{1}{\vert \mathcal{D} \vert} \sum_{\mathcal{D}} - \log \bar{p}^{(N)} \right] \\
&= \frac{1}{\vert \mathcal{D} \vert} \mathbb{V} \left[ - \log \bar{p}^{(N)} \right] \\
&\simeq \frac{1}{\vert \mathcal{D} \vert} \mathbb{V} \left[ - \log \mu + \left( 1 - \frac{\bar{p}^{(N)}}{\mu} \right) \right] \label{eq:sharpness:taylor} \\
&= \frac{1}{\vert \mathcal{D} \vert} \mathbb{V} \left[ - \frac{\bar{p}^{(N)}}{\mu}  \right] \\
&= \frac{1}{N} \frac{\mathbb{V} \left[ p_i \right]}{\mu^2 \vert \mathcal{D} \vert} \\
&= \frac{1}{N} \frac{\sigma^2_{\text{pred}}}{\mu^2 \vert \mathcal{D} \vert} \label{eq:sharpness:variance}
\end{align}
where $\mu = \bar{p}^{(\infty)}$ and $\sigma^2_{\text{pred}}$ is predictive variance of confidence. We use the formula $\mathbb{V} \left[ \frac{1}{N} \sum_{i=1}^{N} \xi \right] = \frac{1}{N} \mathbb{V} \left[ \xi \right ]$ for arbitrary random variable $\xi$, and we take the first-order Taylor expansion with an assumption $\bar{p}^{(N)} \simeq \mu$ in \cref{eq:sharpness:taylor}. 
Therefore, the approximated sharpness is:
\begin{align}
\text{sharpness}_{\rho}^{2} \simeq \frac{1}{N} \frac{\sigma^2_{\text{pred}}}{\mu^2 \vert \mathcal{D} \vert}
\label{eq:sharpness}
\end{align}

In conclusion, \emph{the variance of NLL, (the square of) the sharpness, is proportional to the variance of predictions $\sigma^2_{\text{pred}}$ and inversely proportional to the ensemble size $N$}. As the ensemble size increases in the training phase, the loss landscape becomes smoother. Flat loss landscape results in better predictive performance and generalization \citep{foret2020sharpness}.

In these explanations, we only consider model uncertainty for the sake of simplicity. Extending the formulations to data uncertainty is straightforward. The predictive distribution of data-complemented BNN inference \citep{park2019vqbnn} is:
\begin{align} 
p (\bm{y} \vert \mathcal{S}, \mathcal{D}) 
&= \int p (\bm{y} \vert \bm{x}, \bm{w}) p(\bm{x} \vert \mathcal{S}) p(\bm{w} \vert \mathcal{D}) d\bm{x} d\bm{w} \\
&= \int p (\bm{y} \vert \bm{z}) p(\bm{z} \vert \mathcal{S}, \mathcal{D}) d\bm{z}
\end{align} 
where $\mathcal{S}$ is proximate data distribution, $\bm{z} = (\bm{x}, \bm{w})$, and $p(\bm{z} \vert \mathcal{S}, \mathcal{D}) = p(\bm{x} \vert \mathcal{S}) \, p(\bm{w} \vert \mathcal{D})$. This equation clearly shows that $\bm{w}$ and $\bm{x}$ are symmetric. Therefore, we obtain the formulas including both model and data uncertainty by replacing $\bm{w}$ with joint random variable of $\bm{x}$ and $\bm{w}$, i.e. $\bm{w} \rightarrow \bm{z} = (\bm{w}, \bm{x})$.

\paragraph{Experiment. }

Above, we claim two statements. First, the higher the dropout rate, the sharper the loss landscape. Second, the variance of the loss is inversely proportional to the ensemble size. 

To demonstrate the former claim quantitatively, we compare the Hessian eigenvalue spectra and the Hessian max eigenvalue spectra of MC dropout with various dropout rates. In these experiments, we use ensemble size of one for MC dropout. For detailed explanation of Hessian max eigenvalue spectrum, see \cref{sec:revisiting:gap}.

\cref{fig:dropout:hes} represents the spectra, which reveals that \emph{as the randomness of the model increases, the number of Hessian eigenvalue outliers increases}. Since outliers are detrimental to the optimization process \citep{ghorbani2019investigation}, dropout disturb NN optimization.

To show the latter claim, we evaluate the variance of NLL loss for ensemble size $N_\text{train}$ as shown in \cref{fig:training-phase-ensemble:variance}. As we would expect, \emph{the variance of the NLL loss---the sharpness of the loss landscape---is inversely proportional to the ensemble size} for large $N_\text{train}$.

\subsection{Training Phase Ensembles Lead to Better Performance}
\label{sec:extended-analysis:training-phase}

\Cref{sec:extended-analysis:dropout} raises an immediate question: \emph{Is there a performance difference between ``training with prediction ensemble'' and ``training with a low MC dropout rate, instead of no ensemble''?} Note that both methods reduce the sharpness of the loss landscape. 
This section answers the question by providing theoretical and experimental explanations that the ensemble in the training phase can improve predictive performance.

According to \citet{gal2016dropout}, the total predictive variance (in regression tasks) is:
\begin{align}
\sigma^{2}_{\text{pred}} = \sigma^{2}_{\text{model}} + \sigma^{2}_{\text{sample}} 
\end{align}
where $\sigma^{2}_{\text{model}}$ is model precision and $\sigma^{2}_{\text{sample}}$ is sample variance. Therefore, the model precision is the lower bound of the predictive variance, i.e.:
\begin{align}
\sigma^{2}_{\text{pred}} \geq \sigma^{2}_{\text{model}}
\end{align}
The model precision depends only on the model architecture. For example, in the case of MC dropout,  $\sigma^{2}_{\text{model}}$ is proportional to the dropout rate \citep{gal2016dropout} as follows:
\begin{align}
\sigma^{2}_{\text{model}} \propto \text{dropout rate}
\end{align}
These suggest that model precision dominate predictive variance if the MC dropout rate is large enough, i.e., even if the number of ensembles is increased in the training phase, the predictive variance is almost the same. In contrast, decreasing the MC dropout rate reduces prediction diversity, and it obviously leads to performance degradation. Therefore, in the training phase, \emph{it is better to ensemble predictions than to lower the MC dropout rate}. We believe that the training phase ensemble is strongly correlated with Batch Augmentation \citep{hoffer2020augment}. We leave concrete analysis for future work.

\paragraph{Experiment. }

The experiments below support the theoretical analysis. We train MC dropout by using training-phase ensemble method with various ensemble sizes $N_{\text{train}}$. 
As we would expect, \cref{fig:training-phase-ensemble:performance} shows that \emph{training phase ensemble significantly improves the predictive performance}. 

We also measure the predictive variances of NLL. The predictive variances of the model with $N_{\text{train}} = 1$ and with $N_{\text{train}} = 3$ are $\mathbb{V} \left[ \mathcal{L} \right] = 0.0169$ and $\mathbb{V} \left[ \mathcal{L} \right] = 0.0179$, respectively. Since the predictive variances of the two models are almost the same, we infer that there exists a lower bound.

%% file: appendix/extended-experiment.tex
\section{Extended Informations of Experiments}
\label{sec:extended}

This section provides additional information on the experiments in \cref{sec:experiments}.

\subsection{Image Classification}
\label{sec:extended:classification}

\input{resources/tab-classification}

\input{resources/fig-classification}

We present numerical comparisons in the image classification experiment and discuss the results in detail.

\paragraph{Computational performance. } 

The throughput of MC dropout and ``MC dropout + spatial smoothing'' is 755 and 675 image/sec, respectively, in training phase on ImageNet. As mentioned in  \cref{sec:experiments:classification}, NLL of ``MC dropout + spatial smoothing''  with ensemble size of 2 is comparable to or even better than that of MC dropout with ensemble size of 50. Therefore, ``MC dropout + spatial smoothing''  is 22$\times$ faster than MC dropout with similar predictive performance, in terms of throughput.

\input{resources/fig-corruption}

\paragraph{Predictive performance on test dataset. }

\cref{tab:classification} shows the predictive performance of various deterministic and Bayesian NNs with and without spatial smoothing on CIFAR-10, CIFAR-100, and ImageNet. 
This table suggests the following: 
First, spatial smoothing improves both accuracy and uncertainty in most cases. In particular, \emph{it improves the predictive performance of all models with MC dropouts}.
Second, spatial smoothing significantly improves the predictive performance of VGG, compared with ResNet. VGG has a chaotic loss landscape, which results in poor predictive performance \citep{li2017visualizing}, and spatial smoothing smoothens its loss landscape effectively.
Third, as the depth increases, the performance improvement decreases. Deeper NNs provide more overconfident results \citep{guo2017calibration}, but the number of spatial smoothing layers calibrating uncertainty is fixed.
Last, the performance improvement of ResNeXt, which includes an ensemble in its internal structure, is relatively marginal.

\cref{fig:classification} shows predictive performance of MC dropout and deep ensemble for ensemble size. A deep ensemble with an ensemble size of 1 is a deterministic NN.
This figure shows that spatial smoothing improves efficiency of ensemble size and the predictive performance at ensemble size of 50. 
In addition, spatial smoothing reduces the variance in performance, suggesting that it stabilizes NN training.

One peculiarity of the results on ImageNet is that spatial smoothing degrades ECE of ResNet-50. 
It is because spatial smoothing significantly improves the accuracy in this case, and there tends to be a trade-off between accuracy and ECE, e.g. as shown in \citep{guo2017calibration}, \cref{fig:dropoutrate}, and \cref{fig:temperature}. Instead, spatial smoothing shows the improvement in NLL, another uncertainty metric.

\paragraph{Predictive performance on training datasets. } 

Note that \emph{spatial smoothing helps NN learn strong representations}. In other words, \emph{spatial smoothing does not regularize NNs}, and it reduces the training loss. For example, the NLL of ResNet-18 with MC dropout on CIFAR-100 training dataset is $2.20 \times 10^{-2}$. The training NLL of the ResNet with spatial smoothing is $1.94 \times 10^{-2}$.

\paragraph{Corruption robustness. }

We measure predictive performance on CIFAR-100-C  \citep{hendrycks2019benchmarking} in order to evaluate the robustness of the models against 5 intensities and 15 types of data corruption.
The top row of \cref{fig:robustness} shows the results as a box plot. 
This box plot shows the median, interquartile range (IQR), minimum, and maximum of predictive performance for types. 
They reveal that spatial smoothing improves predictive performance for corrupted data. In particular, spatial smoothing undoubtedly helps in predicting reliable uncertainty.

To summarize the performance of corrupted data in a single value, \citet{hendrycks2019benchmarking} introduced a corruption error (CE) for quantitative comparison. $\text{CE}_{c}^{f}$, which is CE for corruption type $c$ and model $f$, is as follows:
\begin{align}
	\text{CE}_{c}^{f} = \left( \sum_{i=1}^{5} E^{f}_{i,c} \right) \bigg/ \left( \sum_{i=1}^{5} E^{\text{AlexNet}}_{i,c} \right)
	\label{eq:ce}
\end{align}
where $E^{f}_{i,c}$ is top-1 error of $f$ for corruption type $c$ and intensity $i$, and $E^{\text{AlexNet}}_{i,c}$ is the error of AlexNet.
Mean CE or \emph{mCE} summarizes $\text{CE}_{c}^{f}$ by averaging them over 15 corruption types such as Gaussian noise, brightness, and show.
Likewise, to evaluate robustness in terms of uncertainty, we introduce corruption NLL (\emph{CNLL}, $\downarrow$) and corruption ECE (\emph{CECE}, $\downarrow$) as follows:
\begin{align}
	\text{CNLL}_{c}^{f} = \left( \sum_{i=1}^{5} \text{NLL}^{f}_{i,c} \right) \bigg/ \left( \sum_{i=1}^{5} \text{NLL}^{\text{AlexNet}}_{i,c} \right)
	\label{eq:cnll}
\end{align}
and 
\begin{align}
	\text{CECE}_{c}^{f} = \left( \sum_{i=1}^{5} \text{ECE}^{f}_{i,c} \right) \bigg/ \left( \sum_{i=1}^{5} \text{ECE}^{\text{AlexNet}}_{i,c} \right)
	\label{eq:cece}
\end{align}
where $\text{NLL}^{f}_{i,c}$ and $\text{ECE}^{f}_{i,c}$ are NLL and ECE of $f$ for $c$ and $i$, respectively. \emph{mCNLL} and \emph{mCECE} are averages over corruption types. 

 The bottom row of \cref{fig:robustness} shows mCNLL, mCE, and mCECE for ensemble size. They consistently indicates that spatial smoothing improves not only the efficiency but corruption robustness across a whole range of ensemble size.

\paragraph{Adversarial robustness. }

\input{resources/tab-adversarial}

\input{resources/tab-perturbation}

We show that spatial smoothing also improves adversarial robustness. First, we measure the robustness, in terms of accuracy and attack success rate (ASR), of ResNet-50 on ImageNet against popular adversarial attacks, namely FGSM \citep{goodfellow2014explaining} and PGD \citep{madry2017towards}. 
\Cref{tab:adversarial} indicate that both MC dropout and spatial smoothing improve robustness against adversarial attacks.

Next, we find out how spatial smoothing improves adversarial robustness. To this end, similar to \cref{sec:spatial-smoothing:how}, we measure the accuracy on the test datasets with frequency-based FGSM adversarial perturbations. 
This experimental result shows that spatial smoothing is particularly robust against high frequency ($\geq 0.3 \pi$) adversarial attacks. This is because spatial smoothing is a low-pass filter, as we mentioned in \cref{sec:spatial-smoothing:how}. Since the ResNet is vulnerable 
against high frequency adversarial attack, an effective defense of spatial smoothing against high frequency attacks significantly improves the robustness.

\paragraph{Consistency. }

To evaluate the translation invariance of models, we use \emph{consistency} \citep{hendrycks2019benchmarking,zhang2019making}, a metric representing translation consistency for shift-translated data sequences $\mathcal{S} = \{ \bm{x}_1, \cdots, \bm{x}_{M+1} \}$, as follows:
\begin{align}
\text{Consistency} = \frac{1}{M} \sum_{i = 1}^{M} \mathds{1}(g(\bm{x}_i) = g(\bm{x}_{i+1}))
\label{eq:consistency}
\end{align}
where $g(\bm{x}) = \text{arg\,max} \, p(\bm{y} \vert \bm{x}, \mathcal{D})$. 
\Cref{tab:consistency} provides consistency of ResNet-18 on CIFAR-10-P \citep{hendrycks2019benchmarking}. The results shows that MC dropout and deep ensemble improve  consistency, and spatial smoothing improves consistency of both deterministic and Bayesian NNs.

\input{resources/tab-semantic-segmentation}

Prior works \citep{zhang2019making,azulay2018deep} reported qualitative examples in which fluctuating predictive confidence of conventional CNNs harms consistency.
However, surprisingly, we find that \emph{confidence fluctuation has little to do with consistency}. To demonstrate this claim, we introduce cross-entropy consistency (CEC, $\downarrow$), a metric that represents the fluctuation of confidence on a shift-translated data sequence $\mathcal{S} = \{ \bm{x}_1, \cdots, \bm{x}_{M+1} \}$, as follows:
\begin{align}
	\text{CEC} = - \frac{1}{M}  \sum_{i = 1}^{M} f(\bm{x}_i) \cdot \log ( f(\bm{x}_{i + 1} ) )
	\label{eq:cec}
\end{align}
where $f(\bm{x}) = p(\bm{y} \vert \bm{x}, \mathcal{D})$. In \cref{tab:consistency}, high consistency does not mean low CEC; conversely, high consistency tends to be high CEC. Canonical NNs predict overconfident probabilities, and their confidence sometimes changes drastically from near-zero to near-one. Correspondingly, it results in low consistency but low CEC. On the contrary, well-calibrated NNs such as MC dropout provide confidence that oscillates between zero and one, which results in high CEC.

\input{resources/fig-perturbation}

To represent the NN reliability appropreately, we propose \emph{relative confidence} ($\uparrow$) as follows:
\begin{align}
	\text{Relative confidence} = p(y_{\text{true}} \vert \bm{x}, \mathcal{D}) \big/ \, \text{max} \, p(\bm{y} \vert \bm{x}, \mathcal{D})
	\label{eq:relconf}
\end{align}
where $\text{max} \, p(\bm{y} \vert \bm{x}, \mathcal{D})$ is confidence of predictive result and $p(y_{\text{true}} \vert \bm{x}, \mathcal{D})$ is probability of the result for true label. It is 1 when NN classifies the image correctly, and less than 1 when NN classifies it incorrectly. Therefore, relative confidence is a metric that indicates the overconfidence of a prediction when NN's prediction is incorrect.

\Cref{fig:perturbation} shows a qualitative example of consistency on CIFAR-10-P by using relative confidence. This figure suggests that spatial smoothing improves consistency of both deterministic and Bayesian NN.

\subsection{Semantic Segmentation}
\label{sec:extended:semseg}

\cref{tab:extended:semseg-performance} shows the performance of U-Net on the CamVid dataset.
This table indicates that spatial smoothing improves accuracy, uncertainty, and consistency of deterministic and Bayesian NNs. 
In addition, temporal smoothing leads to significant improvement in efficiency of ensemble size, accuracy, uncertainty, and consistency by exploiting temporal information. Moreover, temporal smoothing requires only one ensemble to achieve high predictive performance, since it cooperates with the temporally previous predictions. 
\emph{We obtain the best predictive and computational performance by using both temporal smoothing and spatial smoothing.}

%% file: resources/tab-classification.tex
\begin{table*}
\vskip 0.1in
\setlength\extrarowheight{3pt}

\caption{
\textbf{Spatial smoothing improves both accuracy and uncertainty at the same time}.
Predictive performance of various models with spatial smoothing in image classification on CIFAR-10, CIFAR-100, and ImageNet.
}

\begin{center}
\begin{small}

  \begin{tabular}{cccccccccccc}

    \toprule

    \thead{\textsc{Model} \& \\ \textsc{Dataset}} & \textsc{MC dropout} & \textsc{Smooth} & \textsc{NLL} & \thead{\textsc{Acc}\\(\%)} & \thead{\textsc{ECE}\\(\%)} \\
    \midrule
    \multirow{4}{*}{\thead{VGG-19 \&\\CIFAR-10}}
    & $\cdot$ & $\cdot$ & 0.401 \color{Gray}(-0.000) & 93.1 \color{Gray}(+0.0) & 3.80 \color{Gray}(-0.00) \\
    & $\cdot$ & \checkmark & 0.376 \color{Green}(-0.002) & 93.2 \color{Green}(+0.1) & 5.49 \color{Red}(+1.69) \\
    \cline{2-6}
    & \checkmark & $\cdot$ & 0.238 \color{Gray}(-0.000) & 92.6 \color{Gray}(+0.0) & 3.55 \color{Gray}(-0.00) \\
    & \checkmark & \checkmark & \textbf{0.197 \color{Green}(-0.041)} & \textbf{93.3 \color{Green}(+0.7)} & \textbf{0.68 \color{Green}(-2.86)} \\
    \midrule
    \multirow{4}{*}{\thead{ResNet-18 \&\\CIFAR-10}}
    & $\cdot$ & $\cdot$ & 0.182 \color{Gray}(-0.000) & 95.2 \color{Gray}(+0.0) & 2.75 \color{Gray}(-0.00) \\
    & $\cdot$ & \checkmark & 0.173 \color{Green}(-0.009) & 95.4 \color{Green}(+0.2) & 2.31 \color{Green}(-0.44) \\
    \cline{2-6}
    & \checkmark & $\cdot$ & 0.157 \color{Gray}(-0.000) & 95.2 \color{Gray}(+0.0) & 1.14 \color{Gray}(-0.00) \\
    & \checkmark & \checkmark & \textbf{0.144 \color{Green}(-0.014)} & \textbf{95.5 \color{Green}(+0.2)} & \textbf{1.04 \color{Green}(-0.10)} \\
    \midrule
    \multirow{4}{*}{\thead{VGG-16 \&\\CIFAR-100}}
    & $\cdot$ & $\cdot$ & 2.047 \color{Gray}(-0.000) & 71.6 \color{Gray}(+0.0) & 19.2 \color{Gray}(-0.0) \\
    & $\cdot$ & \checkmark & 1.878 \color{Green}(-0.169) & \textbf{72.2 \color{Green}(+0.6)} & 20.5 \color{Red}(+1.3) \\
    \cline{2-6}
    & \checkmark & $\cdot$ & 1.133 \color{Gray}(-0.000) & 68.8 \color{Gray}(+0.0) & 3.66 \color{Gray}(-0.00) \\
    & \checkmark & \checkmark & \textbf{1.034 \color{Green}(-0.099)} & 71.4 \color{Green}(+2.6) & \textbf{1.06 \color{Green}(-2.60)} \\
    \midrule
    \multirow{4}{*}{\thead{VGG-19 \&\\CIFAR-100}}
    & $\cdot$ & $\cdot$ & 2.016 \color{Gray}(-0.000) & 67.6 \color{Gray}(+0.0) & 21.2 \color{Gray}(-0.0) \\
    & $\cdot$ & \checkmark & 1.851 \color{Green}(-0.165) & \textbf{71.7 \color{Green}(+4.0)} & 20.2 \color{Green}(-1.0) \\
    \cline{2-6}
    & \checkmark & $\cdot$ & 1.215 \color{Gray}(-0.000) & 67.3 \color{Gray}(+0.0) & 6.37 \color{Gray}(-0.00) \\
    & \checkmark & \checkmark & \textbf{1.071 \color{Green}(-0.144)} & 70.4 \color{Green}(+3.0) & \textbf{2.15 \color{Green}(-4.22)} \\
    \midrule
    \multirow{4}{*}{\thead{ResNet-18 \&\\CIFAR-100}}
    & $\cdot$ & $\cdot$ & 0.886 \color{Gray}(-0.000) & 77.9 \color{Gray}(+0.0) & 4.97 \color{Gray}(-0.00) \\
    & $\cdot$ & \checkmark & 0.863 \color{Green}(-0.023) & 78.9 \color{Green}(+1.0) & 4.40 \color{Green}(-0.57) \\
    \cline{2-6}
    & \checkmark & $\cdot$ & 0.848 \color{Gray}(-0.000) & 77.3 \color{Gray}(+0.0) & 3.01 \color{Gray}(-0.00) \\
    & \checkmark & \checkmark & \textbf{0.801 \color{Green}(-0.047)} & \textbf{78.9 \color{Green}(+1.6)} & \textbf{2.56 \color{Green}(-0.45)} \\
    \midrule
    \multirow{4}{*}{\thead{ResNet-50 \&\\CIFAR-100}}
    & $\cdot$ & $\cdot$ & 0.835 \color{Gray}(-0.000) & 79.9 \color{Gray}(+0.0) & 8.88 \color{Gray}(-0.00) \\
    & $\cdot$ & \checkmark & 0.834 \color{Green}(-0.002) & \textbf{80.7 \color{Green}(+0.8)} & 9.29 \color{Red}(+0.42) \\
    \cline{2-6}
    & \checkmark & $\cdot$ & 0.822 \color{Gray}(-0.000) & 79.1 \color{Gray}(+0.0) & \textbf{6.63 \color{Gray}(-0.00)} \\
    & \checkmark & \checkmark & \textbf{0.800 \color{Green}(-0.022)} & 80.1 \color{Green}(+1.0) & 7.25 \color{Red}(+0.62) \\
    \midrule
    \multirow{4}{*}{\thead{ResNeXt-50 \&\\CIFAR-100}}
    & $\cdot$ & $\cdot$ & 0.804 \color{Gray}(-0.000) & 80.6 \color{Gray}(+0.0) & 8.23 \color{Gray}(-0.00) \\
    & $\cdot$ & \checkmark & 0.825 \color{Red}(+0.022) & \textbf{80.8 \color{Green}(+0.3)} & 9.41 \color{Red}(+1.18) \\
    \cline{2-6}
    & \checkmark & $\cdot$ & 0.762 \color{Gray}(-0.000) & 80.5 \color{Gray}(+0.0) & \textbf{5.67 \color{Gray}(-0.00)} \\
    & \checkmark & \checkmark & \textbf{0.759 \color{Green}(-0.002)} & 80.7 \color{Green}(+0.2) & 6.62 \color{Red}(+0.94) \\
    \midrule
    \multirow{4}{*}{\thead{ResNet-18 \&\\ImageNet}}
    & $\cdot$ & $\cdot$ & 1.210 \color{Gray}(-0.000) & 70.3 \color{Gray}(+0.0) & 1.62 \color{Gray}(-0.00) \\
    & $\cdot$ & \checkmark & \textbf{1.183 \color{Green}(-0.027)} & \textbf{70.6 \color{Green}(+0.3)} & \textbf{1.22 \color{Green}(-0.40)} \\
    \cline{2-6}
    & \checkmark & $\cdot$ & 1.215 \color{Gray}(-0.000) & 70.0 \color{Gray}(+0.0) & 1.39 \color{Gray}(-0.00) \\
    & \checkmark & \checkmark & 1.190 \color{Green}(-0.032) & 70.6 \color{Green}(+0.6) & 2.25 \color{Red}(+0.86) \\
    \midrule
    \multirow{4}{*}{\thead{ResNet-50 \&\\ImageNet}}
    & $\cdot$ & $\cdot$ & 0.949 \color{Gray}(-0.000) & 76.0 \color{Gray}(+0.0) & 2.97 \color{Gray}(-0.00) \\
    & $\cdot$ & \checkmark & 0.916 \color{Green}(-0.033) & 76.9 \color{Green}(+0.9) & 3.46 \color{Red}(+0.49) \\
    \cline{2-6}
    & \checkmark & $\cdot$ & 0.945 \color{Gray}(-0.000) & 76.0 \color{Gray}(+0.0) & \textbf{1.89 \color{Gray}(-0.00)} \\
    & \checkmark & \checkmark & \textbf{0.905 \color{Green}(-0.040)} & \textbf{77.0 \color{Green}(+1.0)} & 2.49 \color{Red}(+0.60) \\
    \midrule
    \multirow{4}{*}{\thead{ResNeXt-50 \&\\ImageNet}}
    & $\cdot$ & $\cdot$ & 0.919 \color{Gray}(-0.000) & 77.7 \color{Gray}(+0.0) & 3.63 \color{Gray}(-0.00) \\
    & $\cdot$ & \checkmark & 0.907 \color{Green}(-0.012) & 78.0 \color{Green}(+0.3) & 4.60 \color{Red}(+0.97) \\
    \cline{2-6}
    & \checkmark & $\cdot$ & 0.895 \color{Gray}(-0.000) & 77.7 \color{Gray}(+0.0) & \textbf{2.53 \color{Gray}(-0.00)} \\
    & \checkmark & \checkmark & \textbf{0.887 \color{Green}(-0.008)} & \textbf{78.1 \color{Green}(+0.4)} & 3.28 \color{Red}(+0.75) \\
    \bottomrule
  \end{tabular}

\end{small}
\end{center}
\vskip 0.1in

\label{tab:classification}
\end{table*}

%% file: resources/fig-classification.tex
\newcommand{\size}{0.27}

\begin{figure*}
\begin{center}

\begin{subfigure}[b]{\textwidth}
\centering
\includegraphics[width=\size\textwidth]{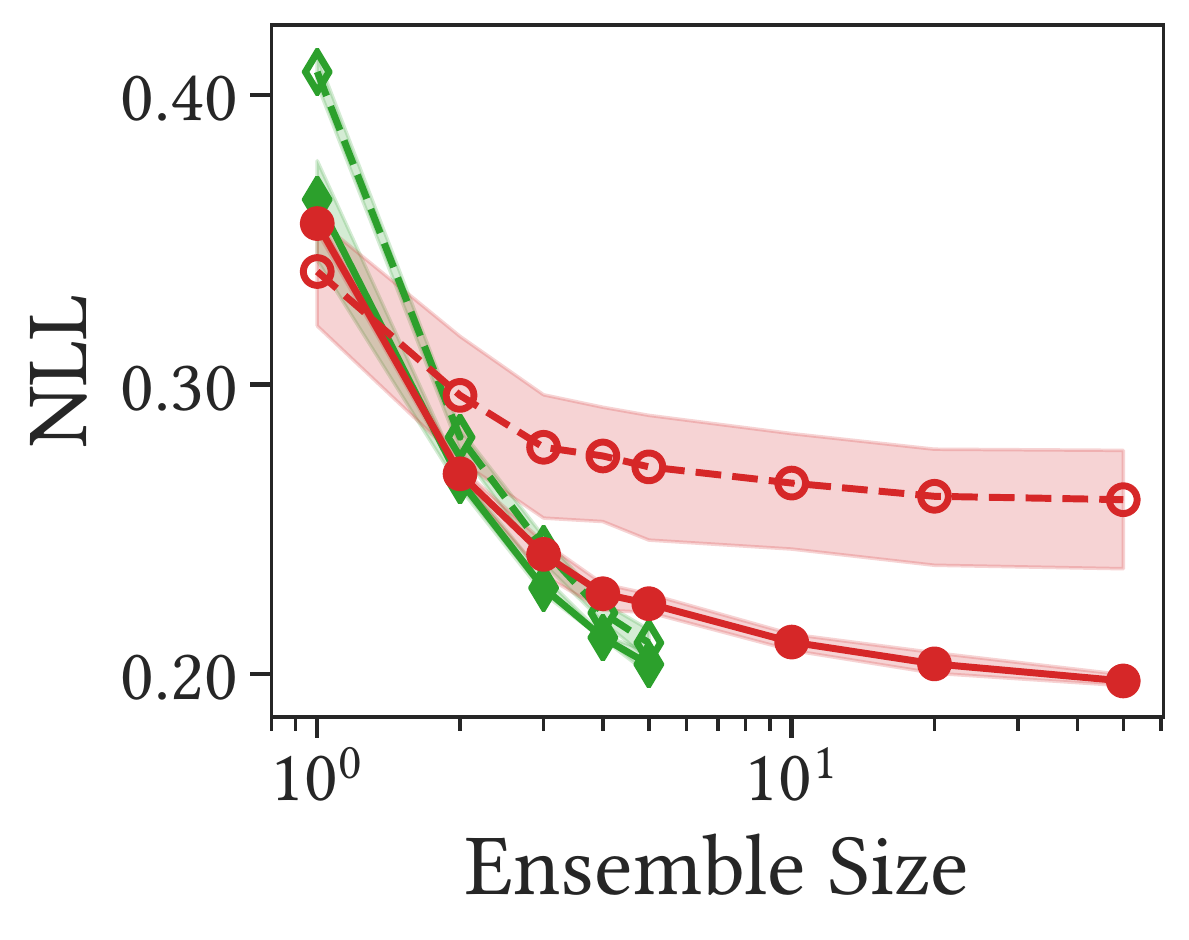}
\hspace{1pt}
\includegraphics[width=\size\textwidth]{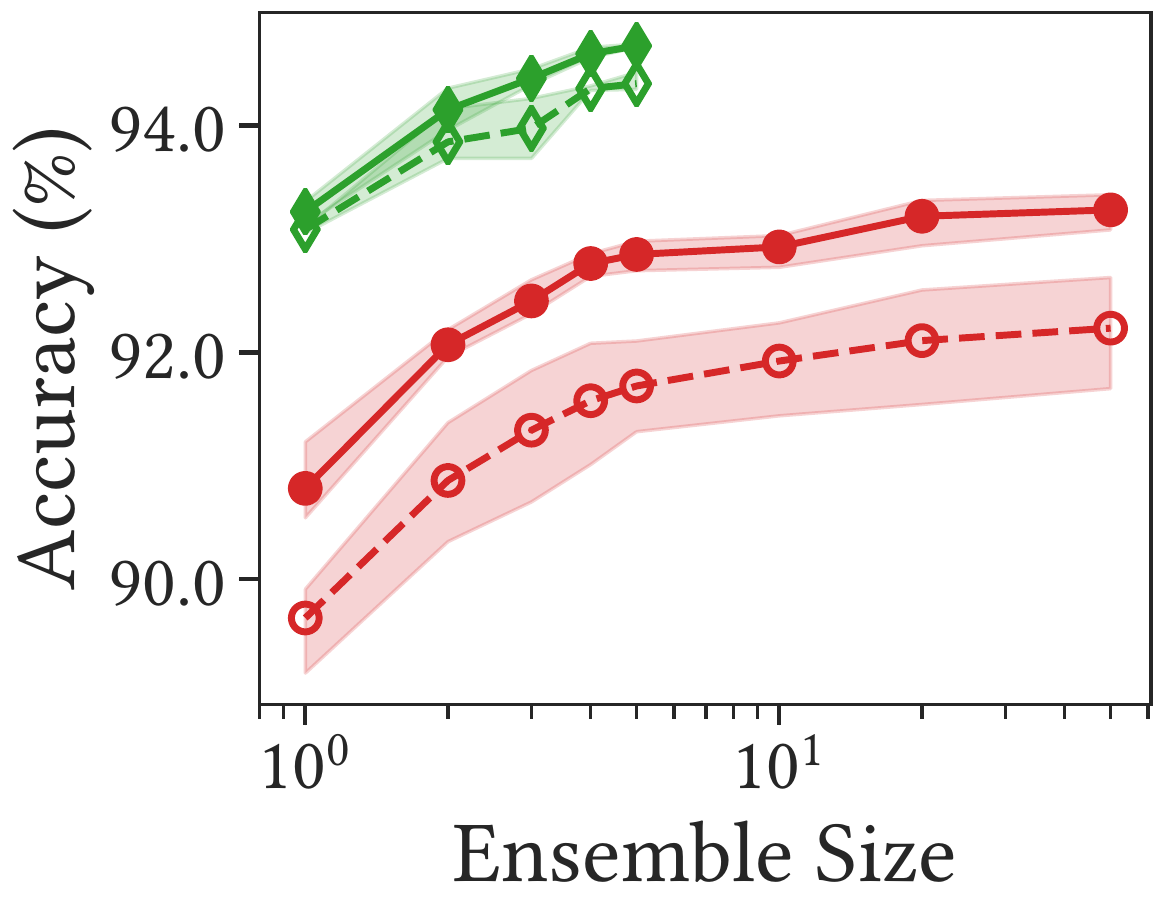}
\hspace{1pt}
\includegraphics[width=\size\textwidth]{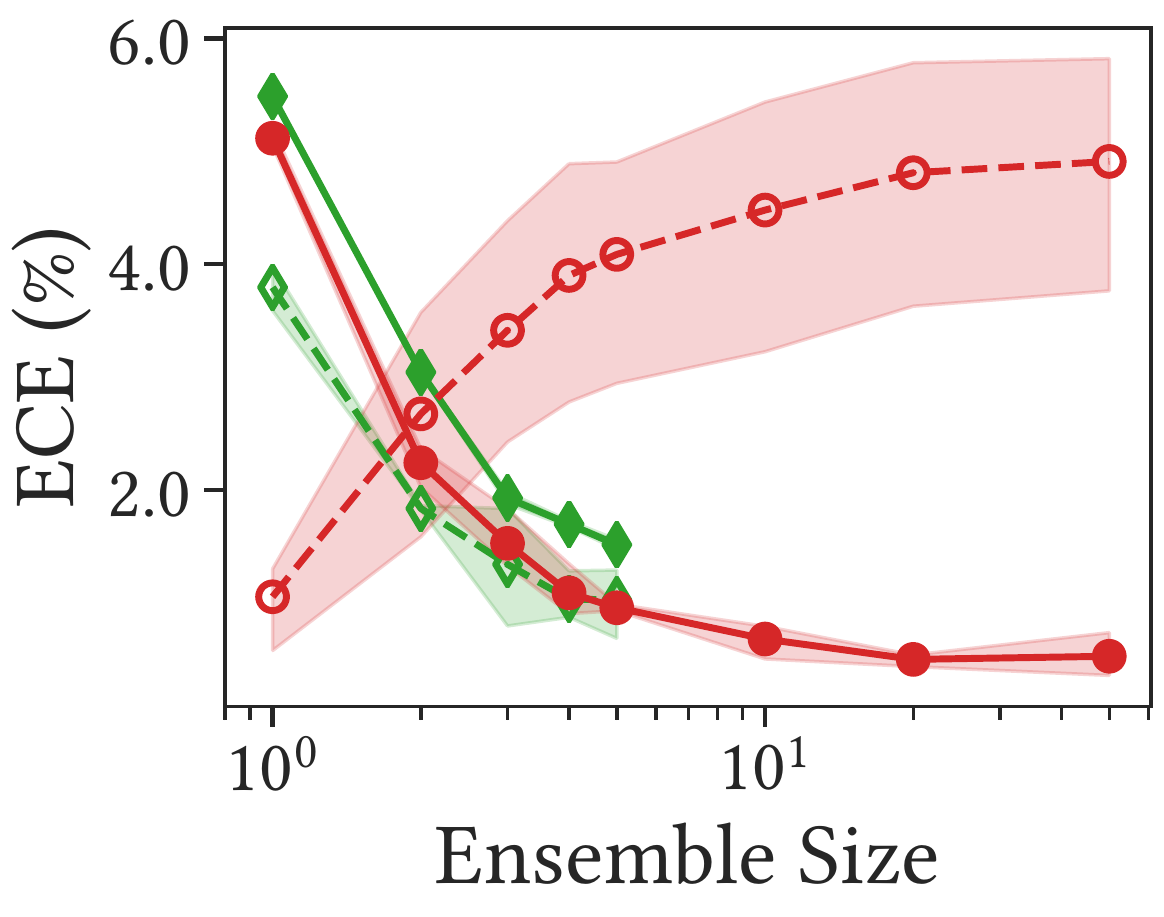}
\caption{VGG-19 on CIFAR-10}
\label{}
\end{subfigure}

\begin{subfigure}[b]{\textwidth}
\centering
\includegraphics[width=\size\textwidth]{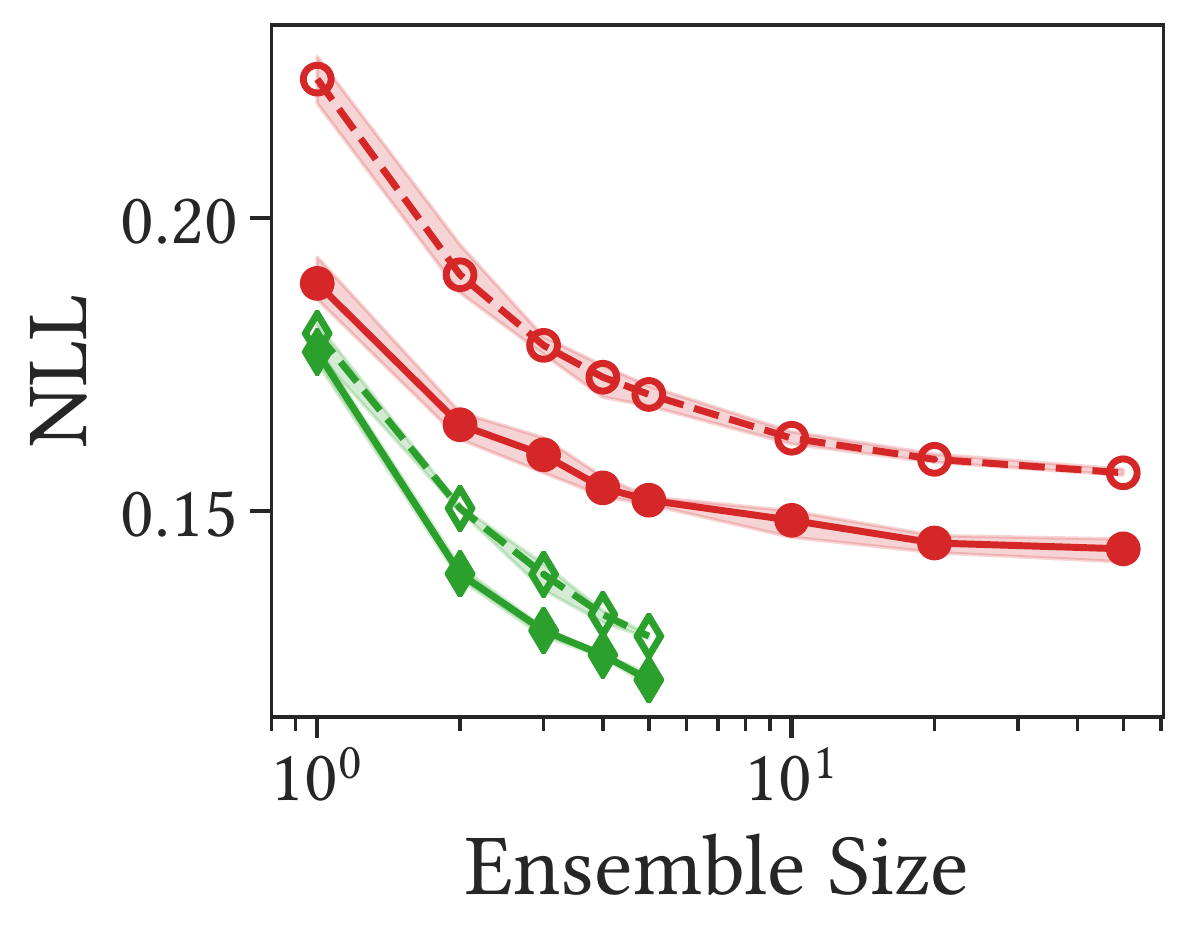}
\hspace{1pt}
\includegraphics[width=\size\textwidth]{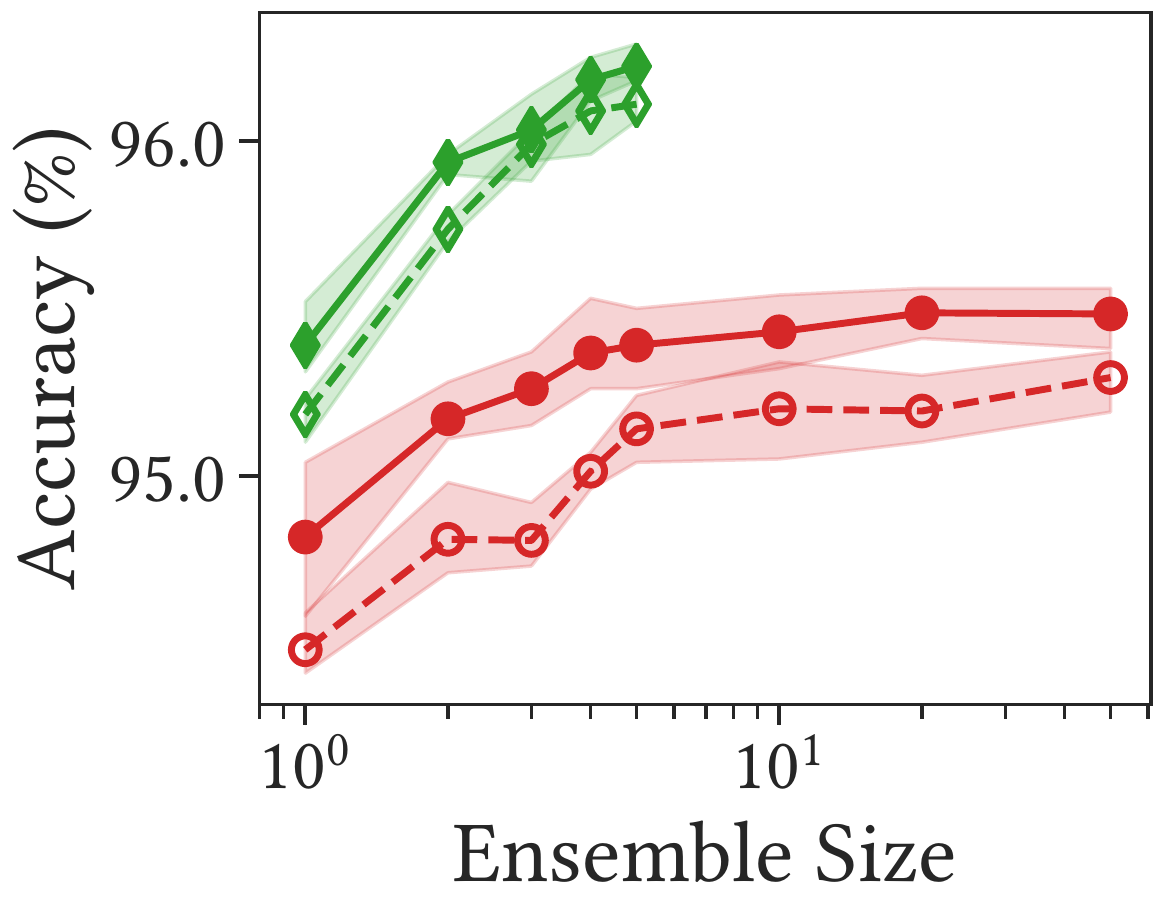}
\hspace{1pt}
\includegraphics[width=\size\textwidth]{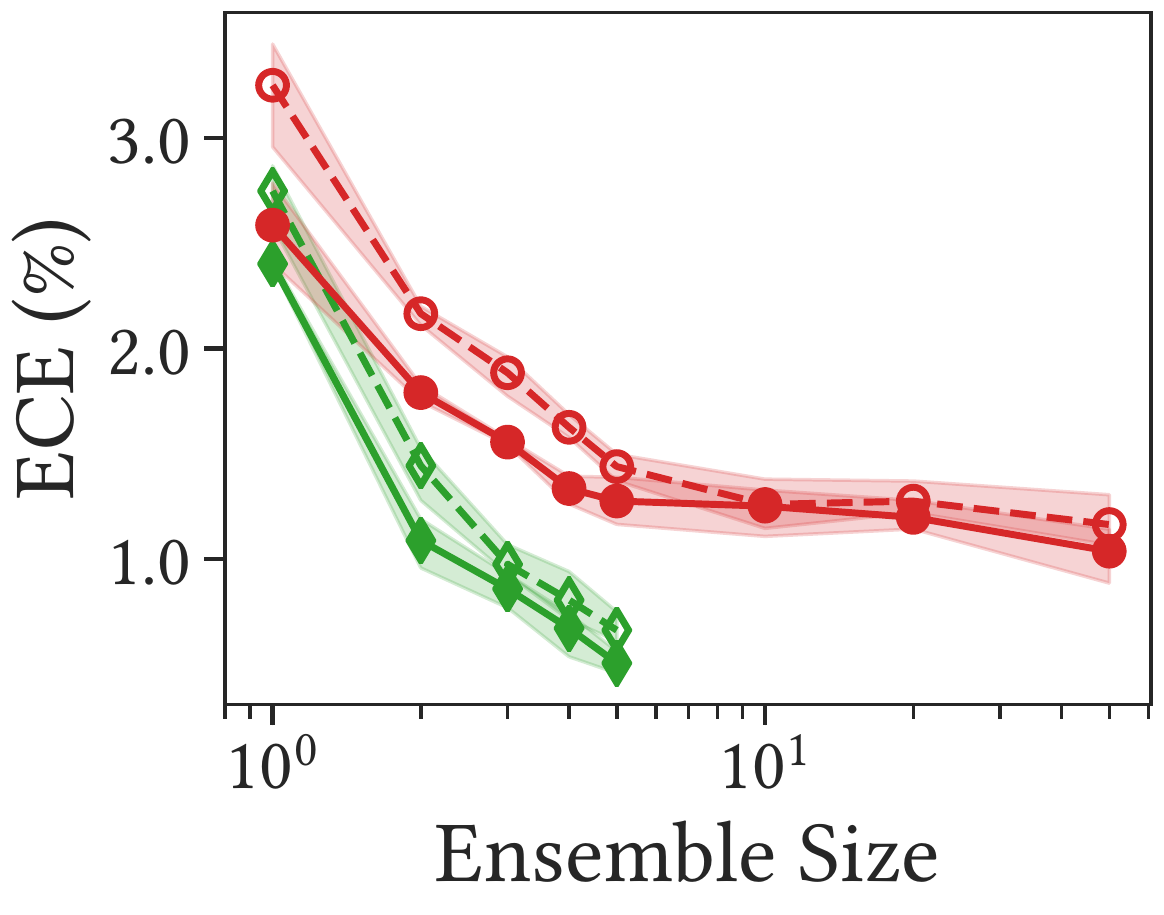}
\caption{ResNet-18 on CIFAR-10}
\label{}
\end{subfigure}

\begin{subfigure}[b]{\textwidth}
\centering
\includegraphics[width=\size\textwidth]{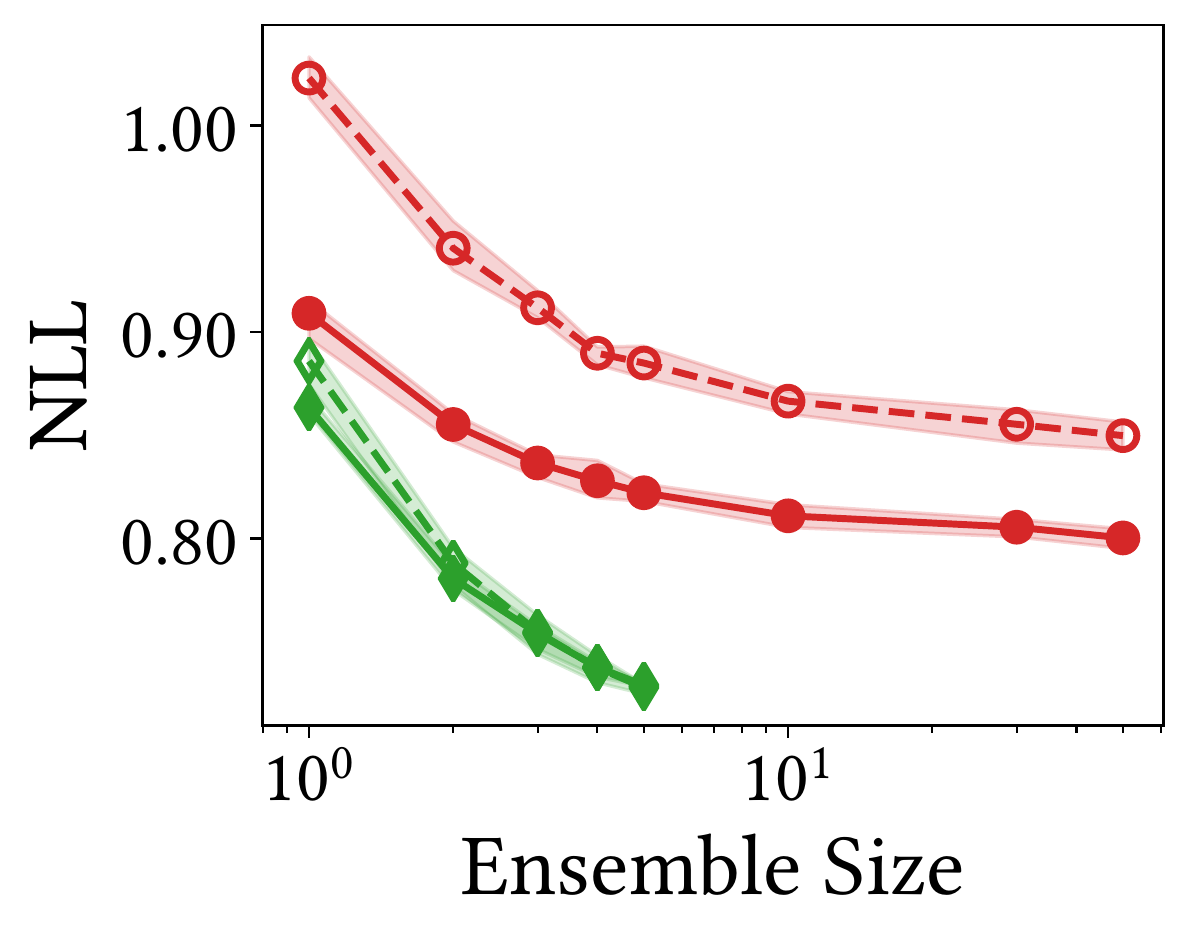}
\hspace{1pt}
\includegraphics[width=\size\textwidth]{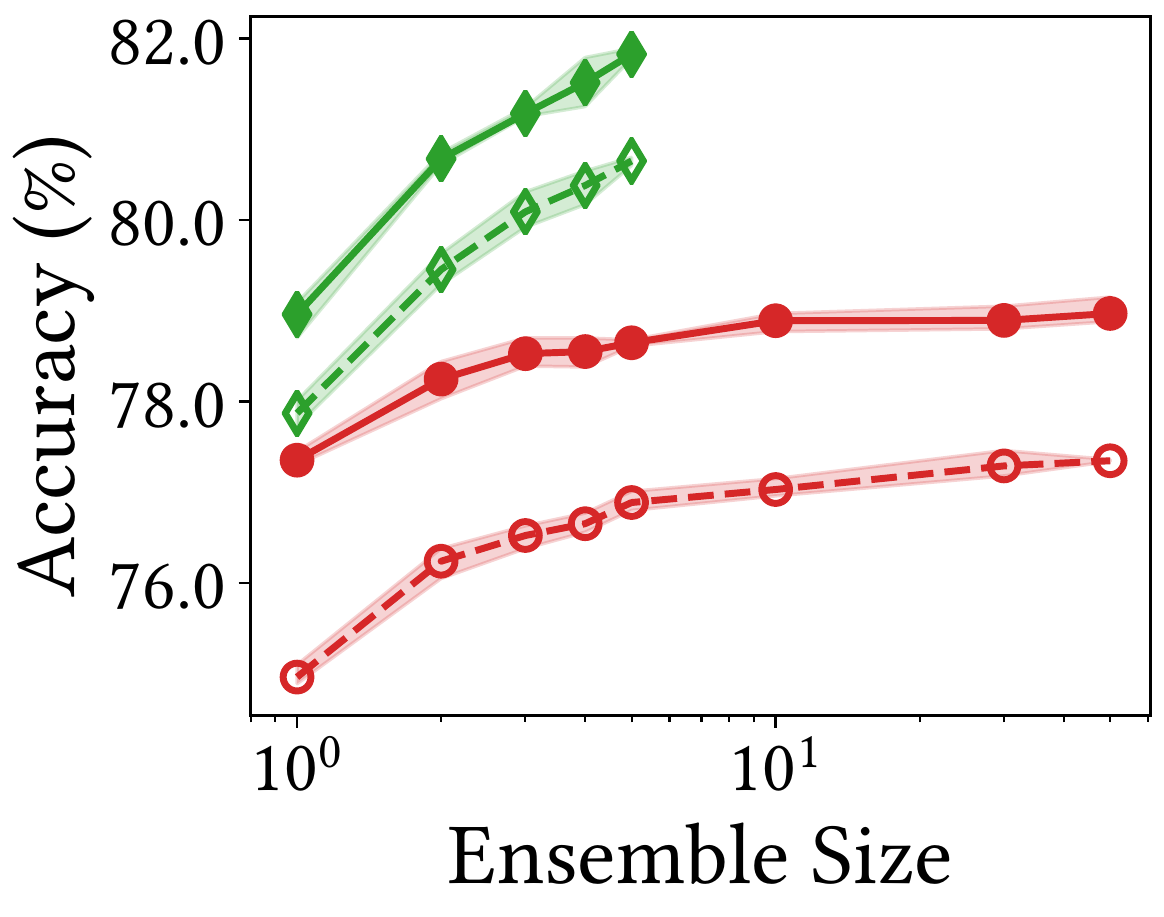}
\hspace{1pt}
\includegraphics[width=\size\textwidth]{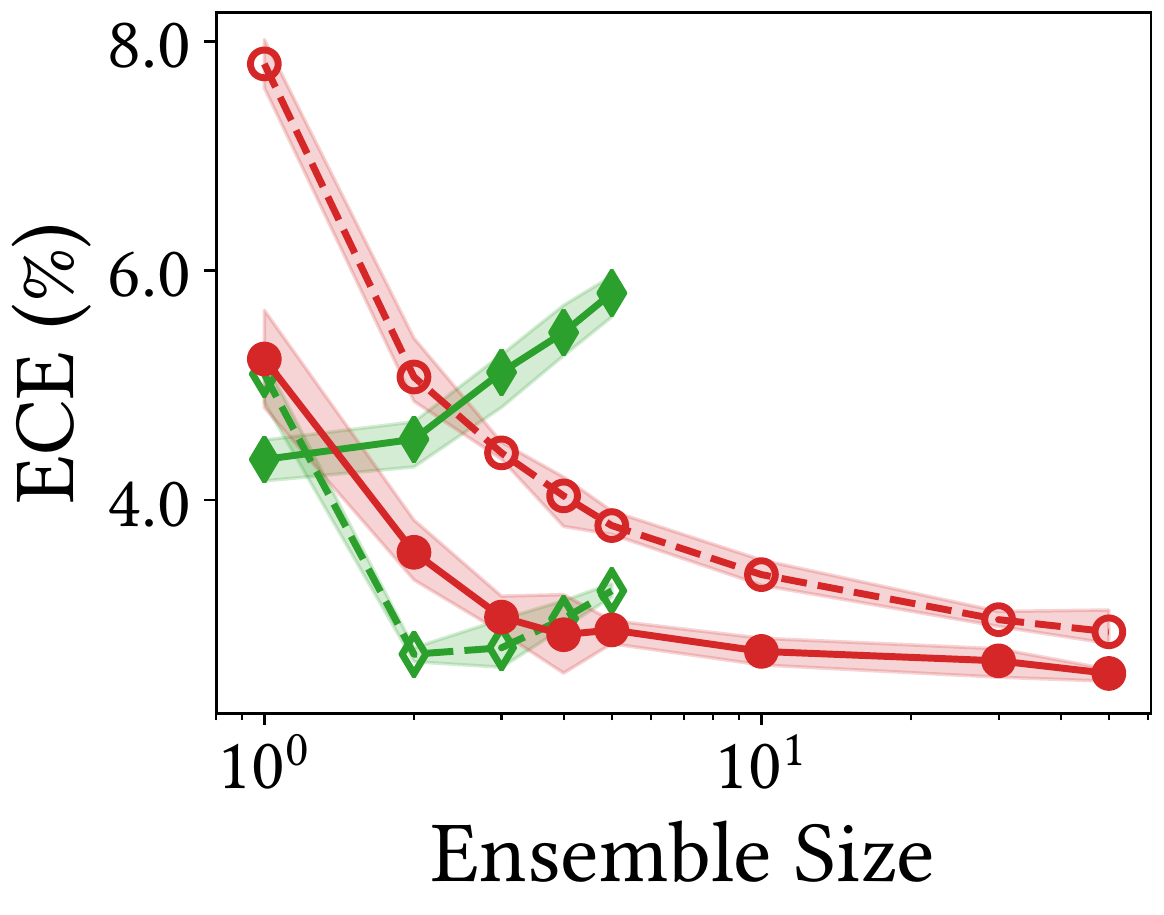}
\caption{ResNet-18 on CIFAR-100}
\label{}
\end{subfigure}

\begin{subfigure}[b]{\textwidth}
\centering
\includegraphics[width=\size\textwidth]{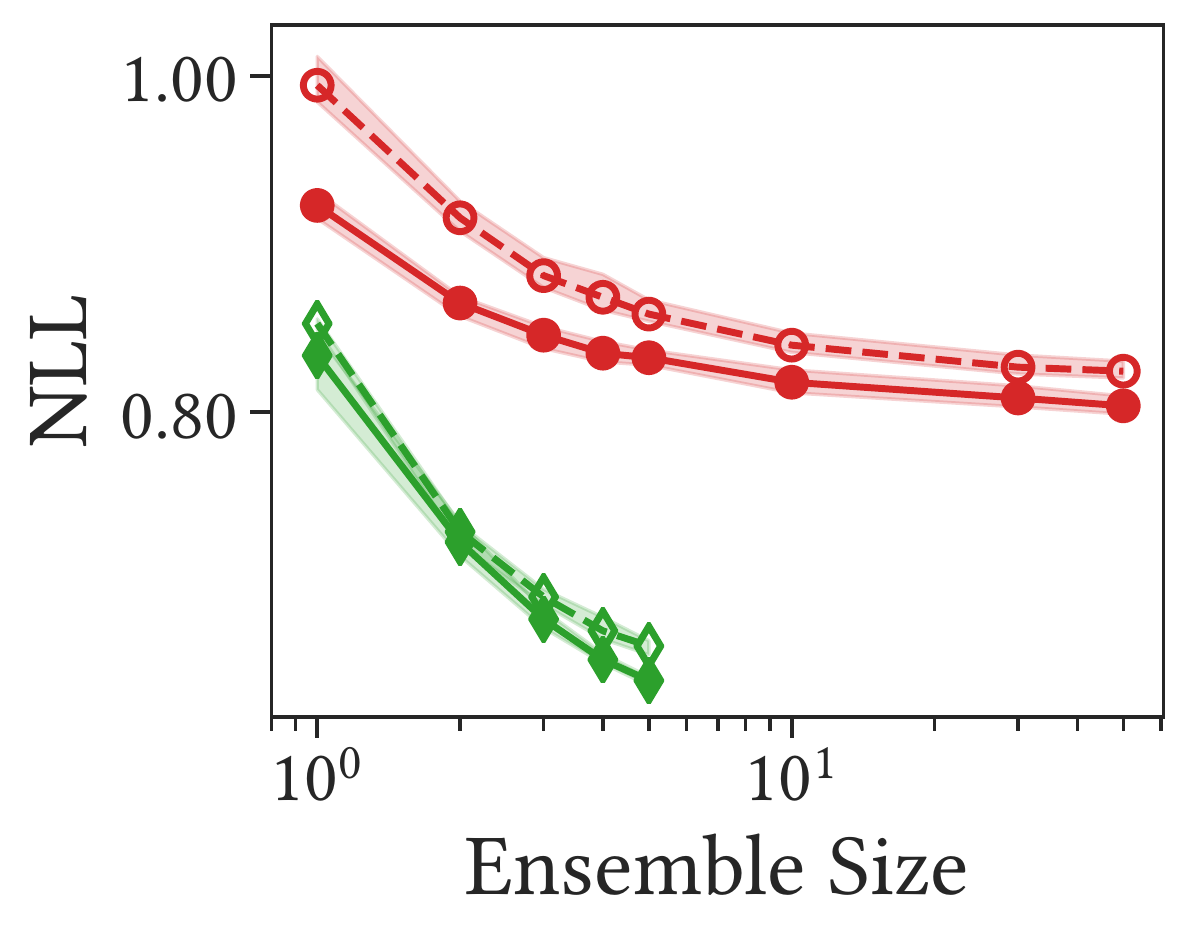}
\hspace{1pt}
\includegraphics[width=\size\textwidth]{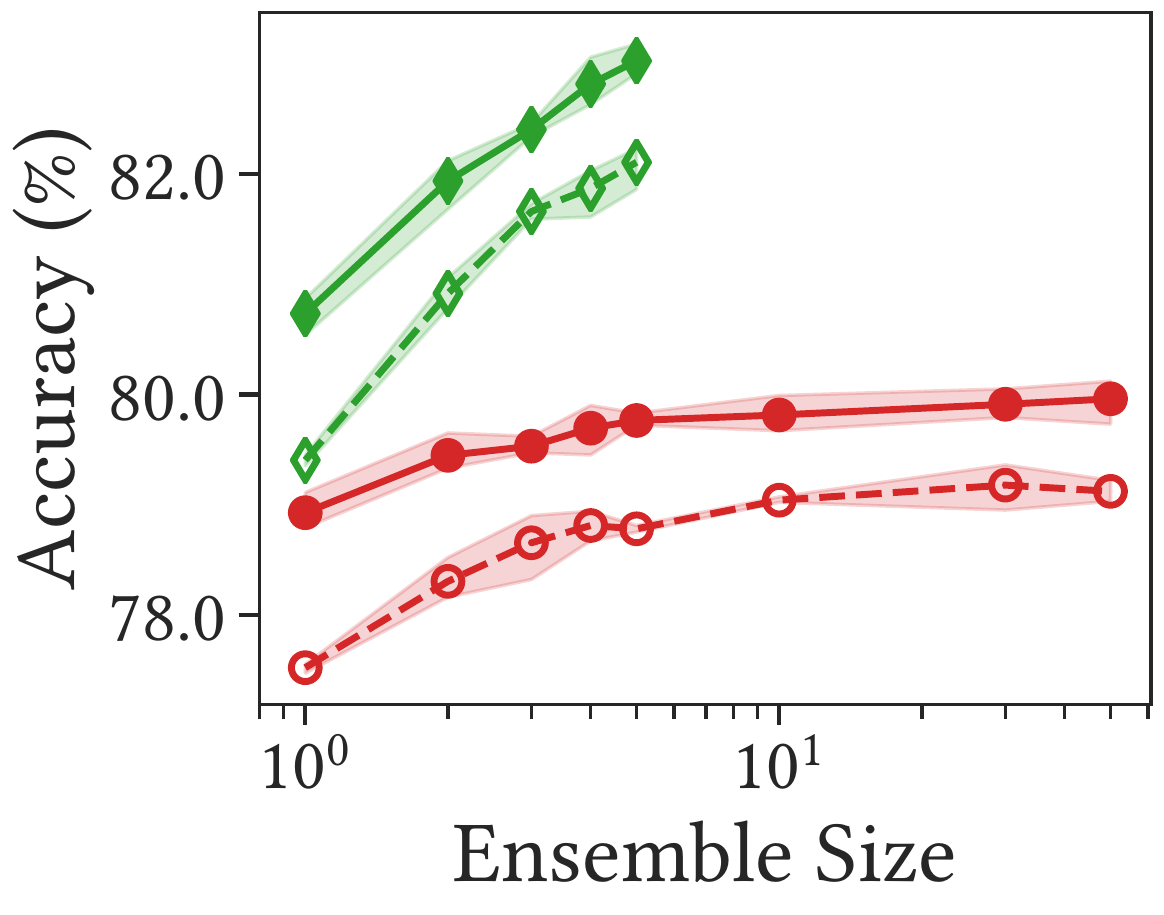}
\hspace{1pt}
\includegraphics[width=\size\textwidth]{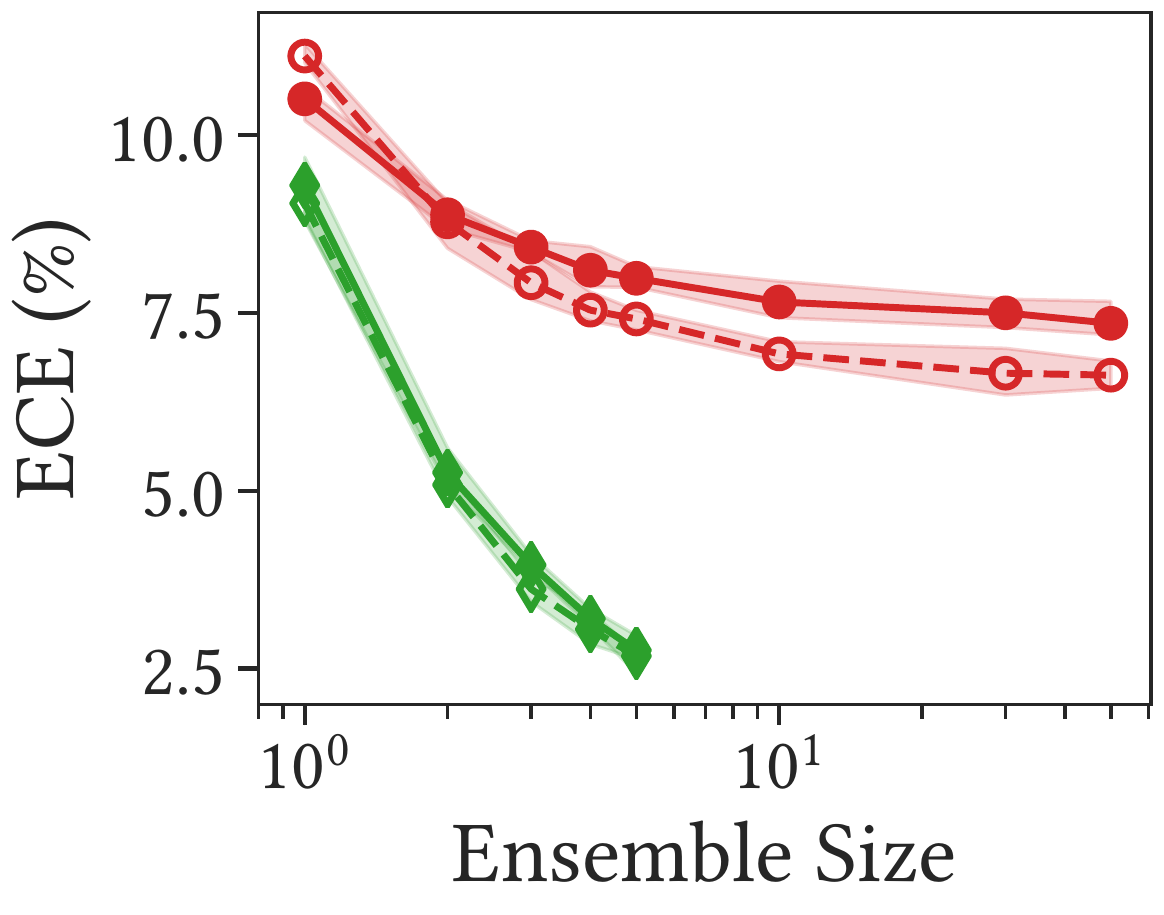}
\caption{ResNet-50 on CIFAR-100}
\label{}
\end{subfigure}

\begin{subfigure}[b]{\textwidth}
\centering
\includegraphics[width=\size\textwidth]{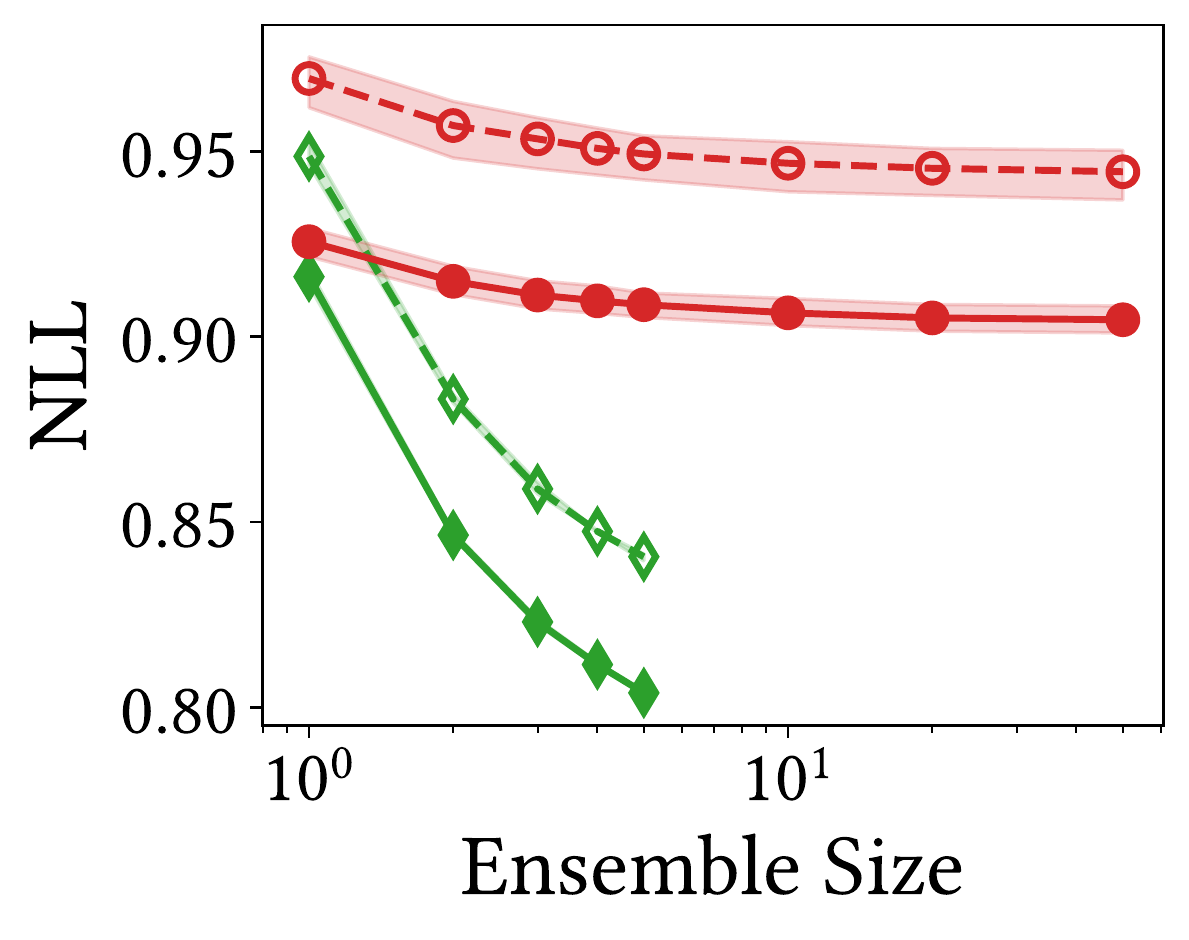}
\hspace{1pt}
\includegraphics[width=\size\textwidth]{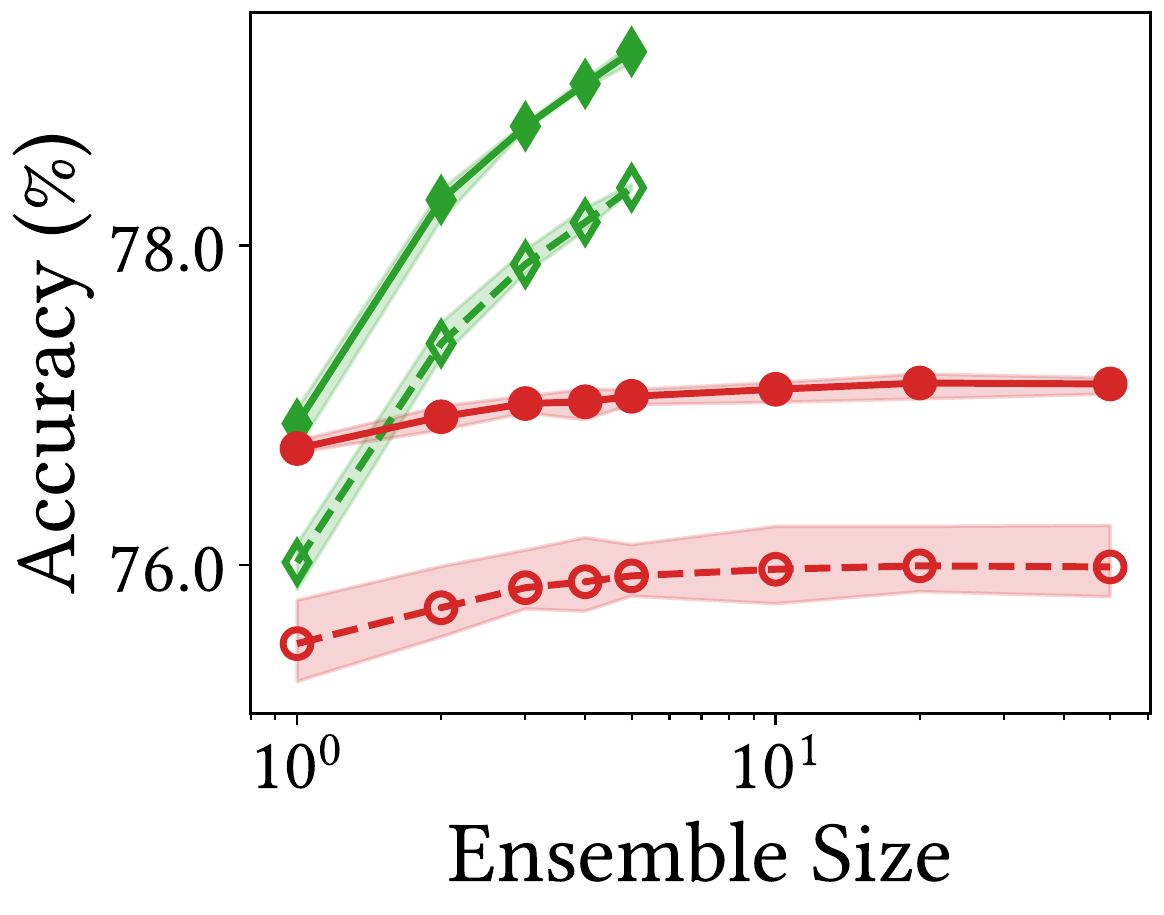}
\hspace{1pt}
\includegraphics[width=\size\textwidth]{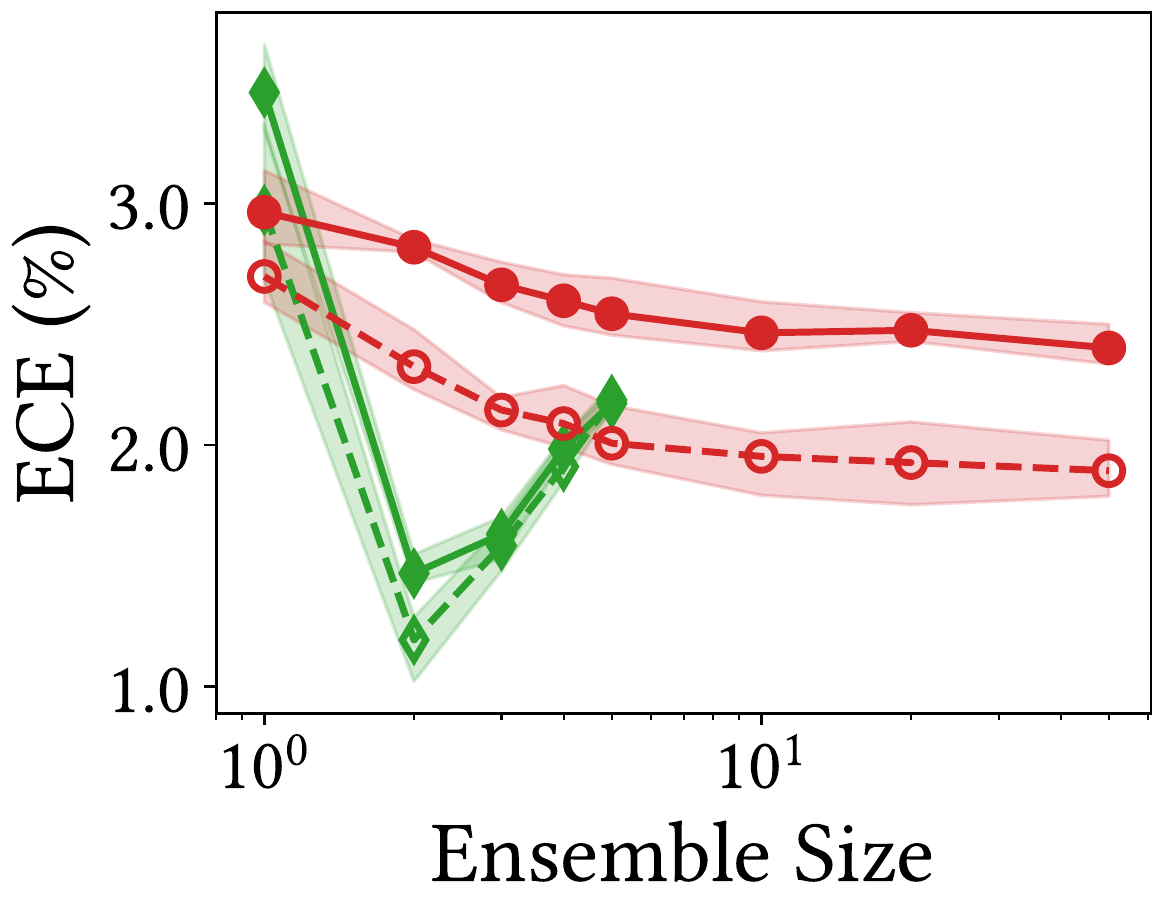}
\caption{ResNet-50 on ImageNet}
\label{}
\end{subfigure}

\vspace{3pt}
\centering
\includegraphics[height=0.027\textheight]{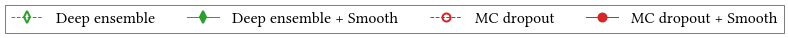}

\caption{
\textbf{Spatial smoothing improves both accuracy and uncertainty across a whole range of ensemble sizes.}
}
\label{fig:classification}
\end{center}
\vskip -0.2in
\end{figure*}

%% file: resources/fig-corruption.tex
\begin{figure*}
\centering

\includegraphics[width=0.30\textwidth]{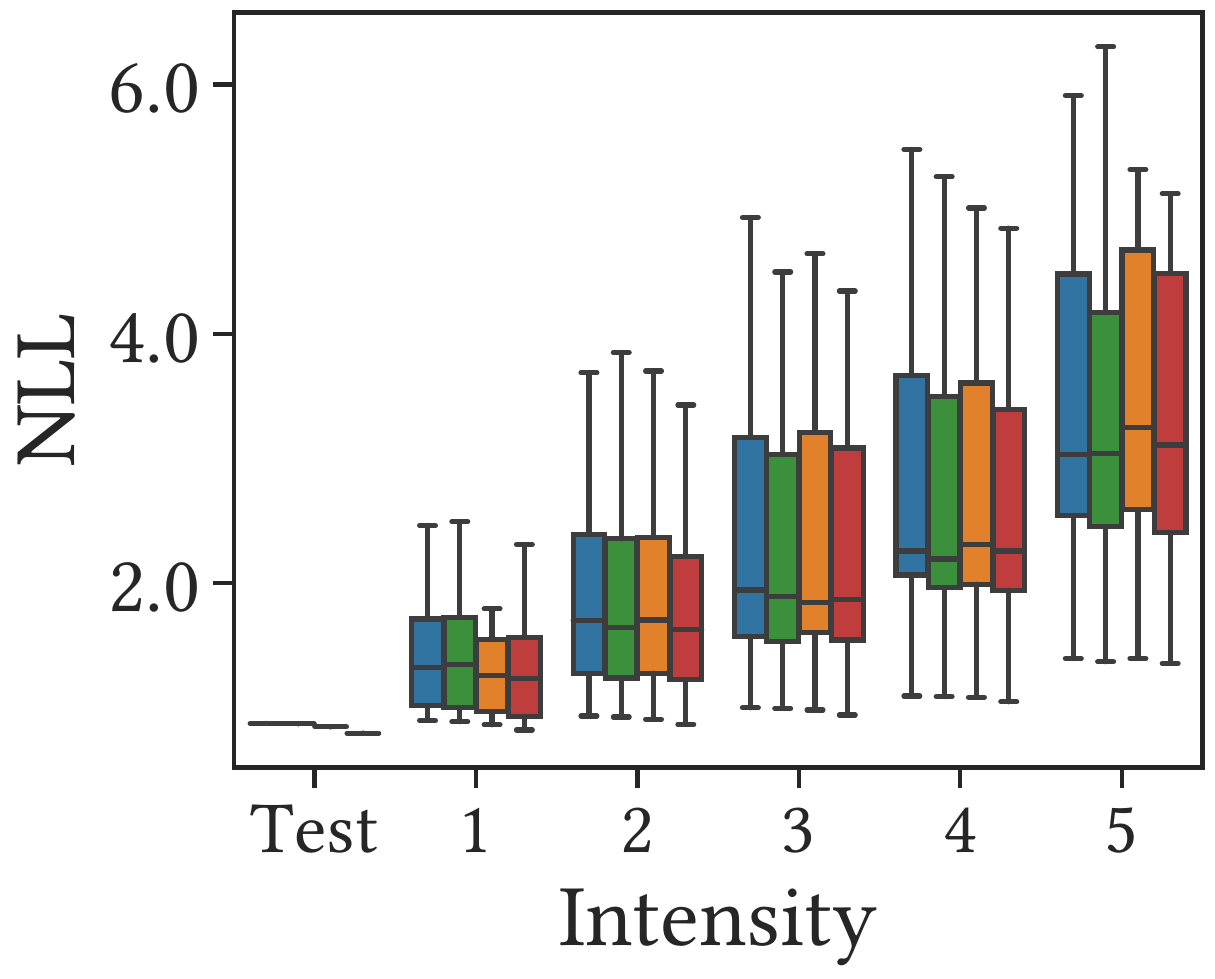}
\hspace{1pt} 
\includegraphics[width=0.30\textwidth]{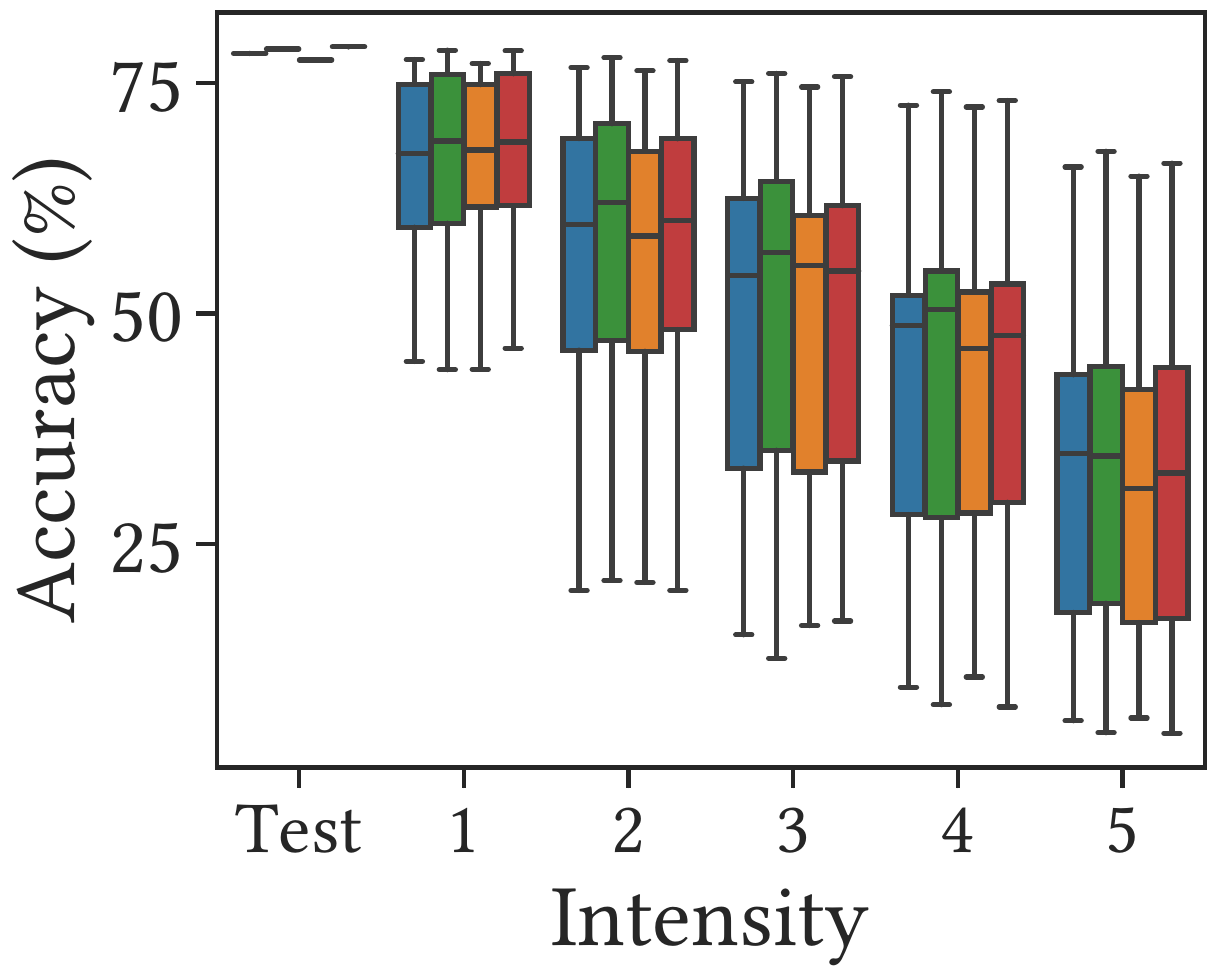}
\hspace{1pt}
\includegraphics[width=0.30\textwidth]{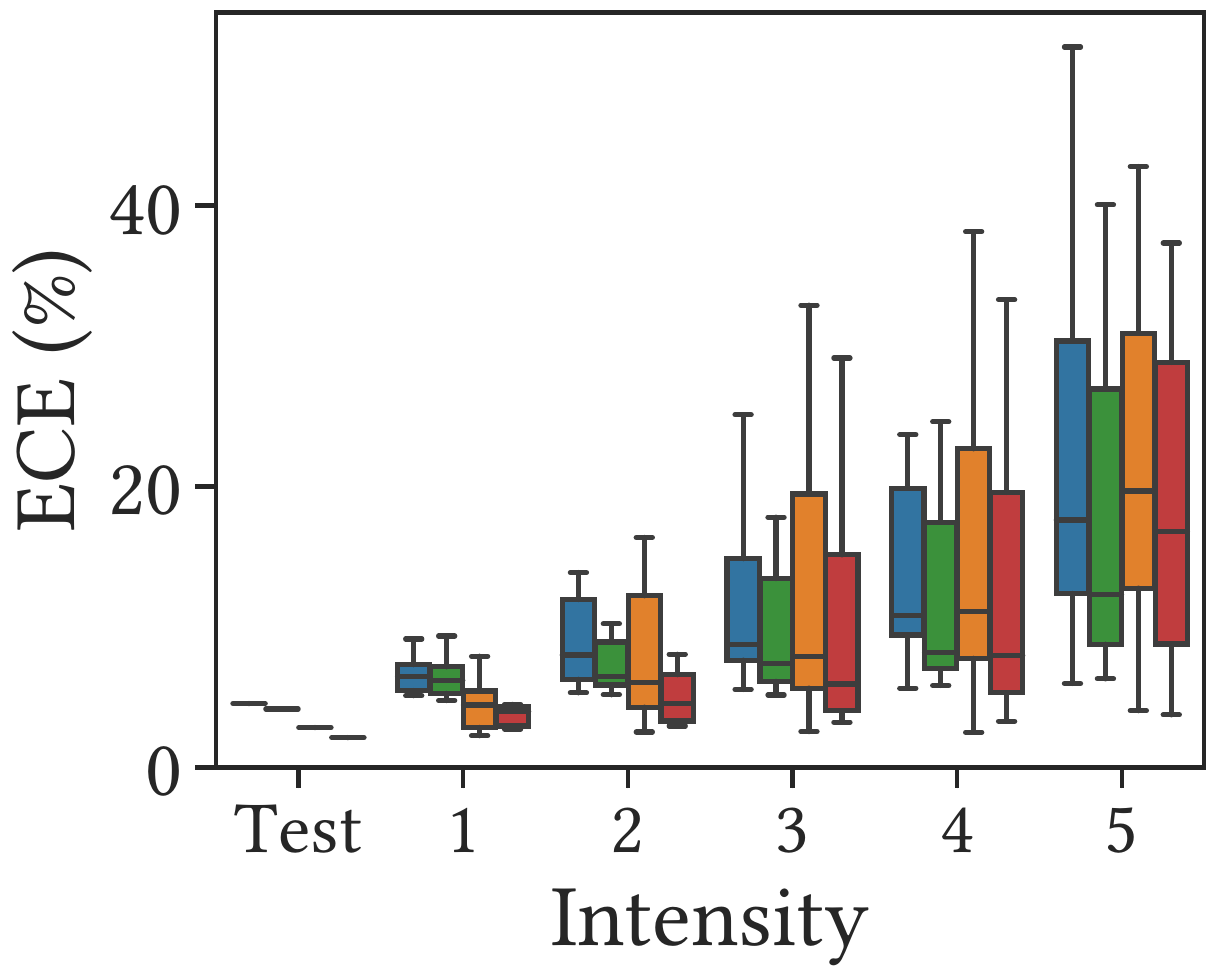}

\vspace{2pt} 
\centering
\includegraphics[width=0.79\textwidth]{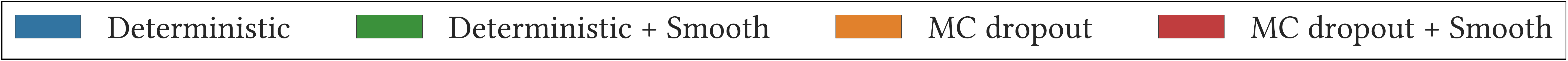}

\vspace{5pt}

\includegraphics[width=0.30\textwidth]{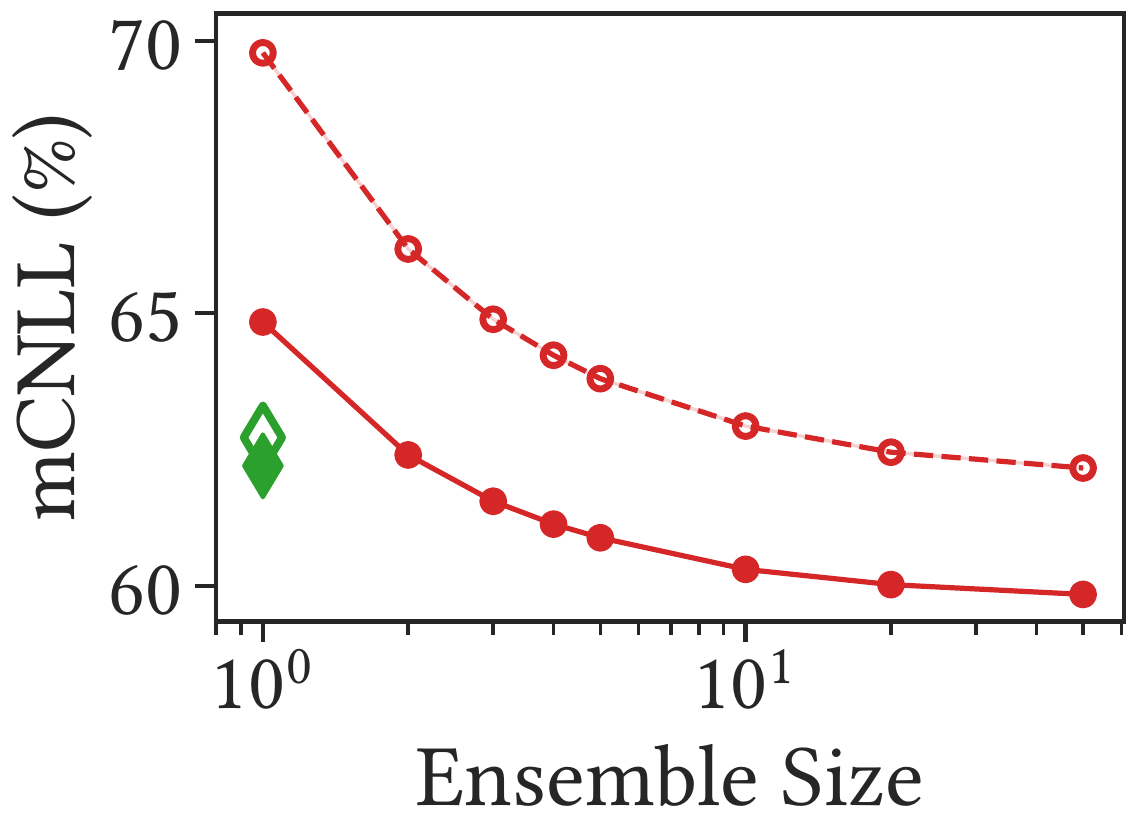}
\hspace{1pt}
\includegraphics[width=0.30\textwidth]{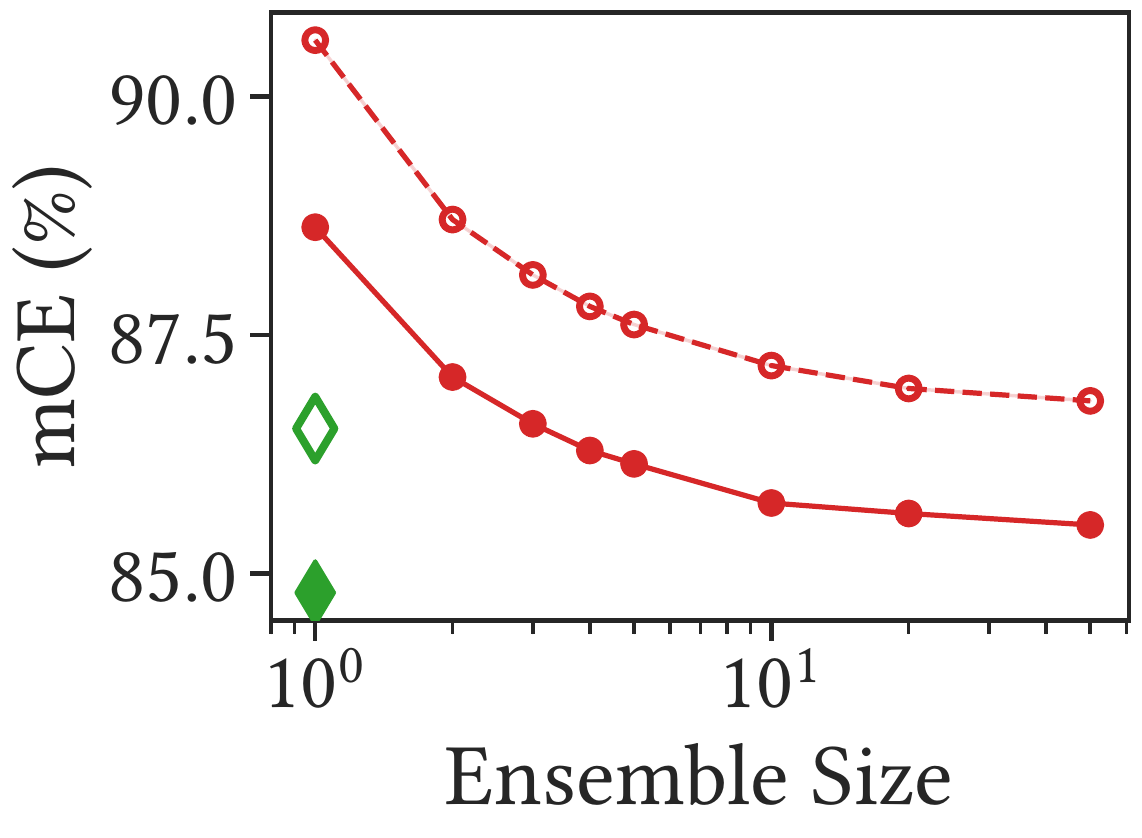}
\hspace{1pt}
\includegraphics[width=0.30\textwidth]{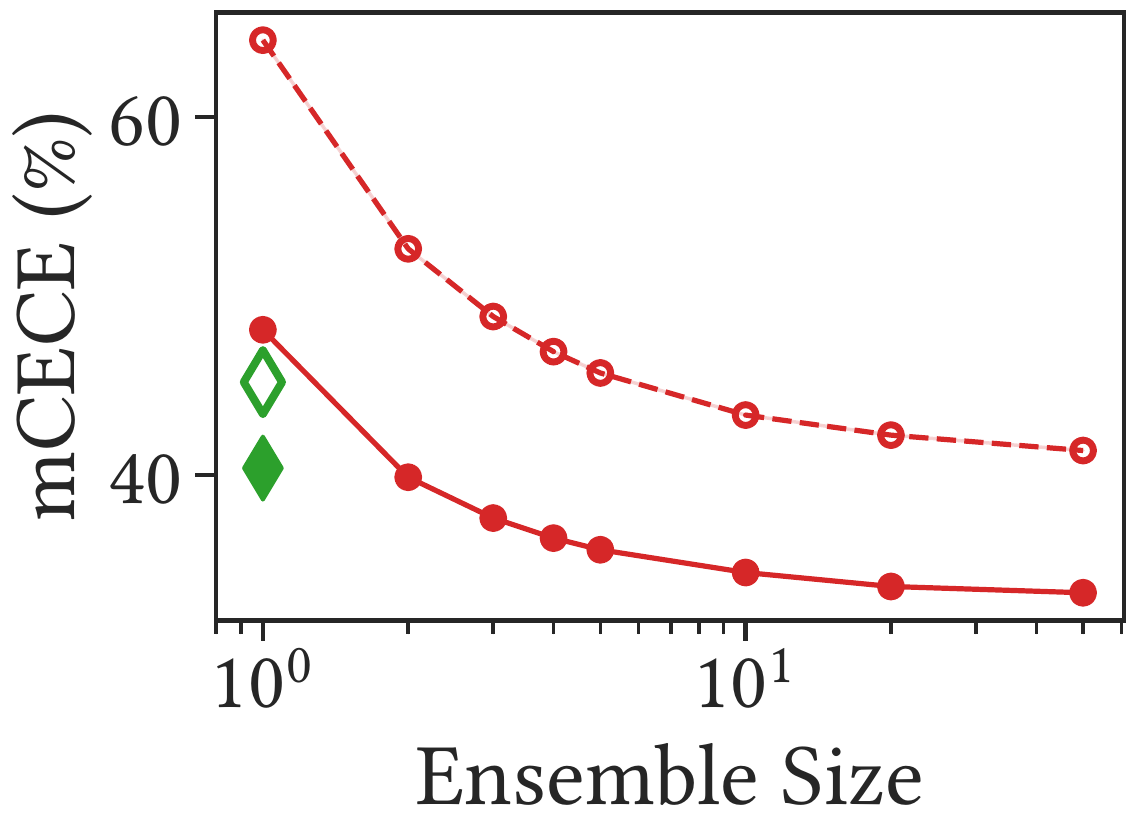}

\vspace{0pt}
\centering
\includegraphics[width=0.78\textwidth]{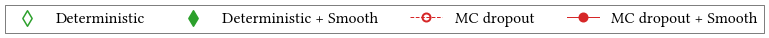}

\caption{
\textbf{Spatial smoothing improves corruption robustness}.
We measure the predictive performance of ResNet-18 on CIFAR-100-C. In the top row, we use an ensemble size of fifty for MC dropout with and without spatial smoothing.
}

\vskip -0.1in

\label{fig:robustness}
\end{figure*}

%% file: resources/tab-adversarial.tex
\begin{table*}
\vskip 0.1in
\setlength\extrarowheight{3pt} 

\caption{
\textbf{Spatial smoothing improves adversarial robustness. }
We measure the accuracy (\textsc{Acc}) and the Attack Success Rate (\textsc{ASR}) of ResNet-50 against adversarial attacks on ImageNet.
}

\begin{center}
\begin{small}

  \begin{tabular}{cccccccccccc}

    \toprule

    \thead{\textsc{Attack}} & \textsc{MC dropout} & \textsc{Smooth} & \thead{\textsc{Acc}\\(\%)} & \thead{\textsc{ASR}\\(\%)} \\
    \midrule
    \multirow{4}{*}{\thead{FGSM}}
    & $\cdot$ & $\cdot$ & 28.3 \color{Gray}(+0.0) & 62.9 \color{Gray}(-0.0) \\
    & $\cdot$ & \checkmark & 30.3 \color{Green}(+2.0) & 60.5 \color{Green}(-2.4) \\
    \cline{2-5}
    & \checkmark & $\cdot$ & 30.3 \color{Gray}(+0.0) & 59.8 \color{Gray}(-0.0) \\
    & \checkmark & \checkmark & \textbf{32.6 \color{Green}(+2.3)} & \textbf{57.4 \color{Green}(-2.4)} \\
    \midrule
    \multirow{4}{*}{\thead{PGD}}
    & $\cdot$ & $\cdot$ & 7.5 \color{Gray}(+0.0) & 90.1 \color{Gray}(-0.0) \\
    & $\cdot$ & \checkmark & 9.0 \color{Green}(+1.4) & 88.2 \color{Green}(-1.9) \\
    \cline{2-5}
    & \checkmark & $\cdot$ & 12.2 \color{Gray}(+0.0) & 83.7 \color{Gray}(-0.0) \\
    & \checkmark & \checkmark & \textbf{13.7 \color{Green}(+1.5)} & \textbf{82.1 \color{Green}(-1.6)} \\
    \bottomrule
  \end{tabular}

\end{small}
\end{center}
\vskip 0.1in

\label{tab:adversarial}
\end{table*}

%% file: resources/tab-perturbation.tex
\begin{table*}[t]

\vskip 0.2in
\setlength\extrarowheight{3pt} 

\caption{
\textbf{Spatial smoothing improves the consistency, robustness against shift-perturbation.}
We measure the consistency of ResNet-18 on CIFAR-10-P. Deterministic NN with $N=5$ means deep ensemble.
}

\begin{center}
\begin{small}
\begin{sc}

  \begin{tabular}{cccccccccccc}
    \toprule

    \textsc{MC dropout} & \textsc{Smooth} & \textsc{$N$} & \thead{\textsc{Cons}\\(\%)} & \thead{\textsc{CEC}\\($\times 10^{-2}$)} \\
    \midrule
    $\cdot$ & $\cdot$ & 1 & 97.9 \color{Gray}(+0.0) & \textbf{1.03 \color{Gray}(-0.00)} \\
    $\cdot$ & \checkmark & 1 & 98.2 \color{Green}(+0.3) & 1.16 \color{Red}(+0.13) \\
    \cline{1-5}
    $\cdot$ & $\cdot$ & 5 & 98.7 \color{Gray}(+0.0) & 1.22 \color{Gray}(-0.00) \\
    $\cdot$ & \checkmark & 5 & \textbf{98.9 \color{Green}(+0.2)} & 1.33 \color{Red}(+0.11) \\
    \cline{1-5}
    \checkmark & $\cdot$ & 50 & 98.2 \color{Gray}(+0.0) & 1.29 \color{Gray}(-0.00) \\
    \checkmark & \checkmark & 50 & 98.4 \color{Green}(+0.2) & 1.34 \color{Red}(+0.05) \\

    \bottomrule
  \end{tabular}

\end{sc}
\end{small}
\end{center}
\vskip 0.0in

\label{tab:consistency}
\end{table*}

%% file: resources/tab-semantic-segmentation.tex
\begin{table*}[t]
\vskip 0.1in
\setlength\extrarowheight{3pt} 

\caption{
\textbf{Spatial smoothing and temporal smoothing are complementary.} 
We provide predictive performance of MC dropout in semantic segmentation on CamVid for each method. \textsc{Spat} and \textsc{Temp} each stand for spatial smoothing and temporal smoothing. \textsc{Cons} stands for consistency. 
}

\begin{center}
\begin{small}

  \begin{tabular}{cccccccccccc}
    \toprule

    \textsc{MC dropout} & \textsc{Spat} & \textsc{Temp} & \thead{$N$} & \textsc{NLL} & \thead{\textsc{Acc}\\(\%)} & \thead{\textsc{ECE}\\(\%)} & \thead{\textsc{Cons}\\(\%)} \\
    \midrule
    $\cdot$ & $\cdot$ & $\cdot$ & 1 & 0.354 \color{Gray}(+0.000) & 92.3 \color{Gray}(+0.0) & 4.95 \color{Gray}(+0.00) & 95.1 \color{Gray}(+0.0) \\
    $\cdot$ & \checkmark & $\cdot$ & 1 & 0.318 \color{Green}(+0.036) & 92.4 \color{Green}(+0.1) & 4.54 \color{Green}(+0.41) & 95.5 \color{Green}(+0.4) \\
    $\cdot$ & $\cdot$ & \checkmark & 1 & 0.290 \color{Green}(+0.064) & 92.5 \color{Green}(+0.2) & 3.18 \color{Green}(+1.77) & 96.3 \color{Green}(+1.2) \\
    $\cdot$ & \checkmark & \checkmark & 1 & 0.278 \color{Green}(+0.076) & 92.5 \color{Green}(+0.2) & 3.03 \color{Green}(+1.92) & \textbf{96.6 \color{Green}(+1.5)} \\   
    \cline{1-8}
    \checkmark & $\cdot$ & $\cdot$ & 50 & 0.298 \color{Gray}(+0.000) & 92.5 \color{Gray}(+0.0) & 4.20 \color{Gray}(+0.00) & 95.4 \color{Gray}(+0.0) \\  
    \checkmark & \checkmark & $\cdot$ & 50 & 0.284  \color{Green}(+0.014) & 92.6 \color{Green}(+0.1) & 3.96 \color{Green}(+0.24) & 95.6 \color{Green}(+0.2) \\
    \checkmark & $\cdot$ & \checkmark & 1 & 0.273 \color{Green}(+0.025) & 92.6 \color{Green}(+0.1) & 3.23 \color{Green}(+0.97) & 96.4 \color{Green}(+1.0) \\  
    \checkmark & \checkmark & \checkmark & 1 & \textbf{0.260 \color{Green}(+0.038)} & \textbf{92.6 \color{Green}(+0.1)} & \textbf{2.71 \color{Green}(+1.49)} & 96.5 \color{Green}(+1.1) \\
    \bottomrule
  \end{tabular}

\end{small}
\end{center}
\vskip -0.1in

\label{tab:extended:semseg-performance}
\end{table*}

%% file: resources/fig-perturbation.tex
\begin{figure}[ht]

\vspace{7pt}

\centering

\includegraphics[width=0.39\textwidth]{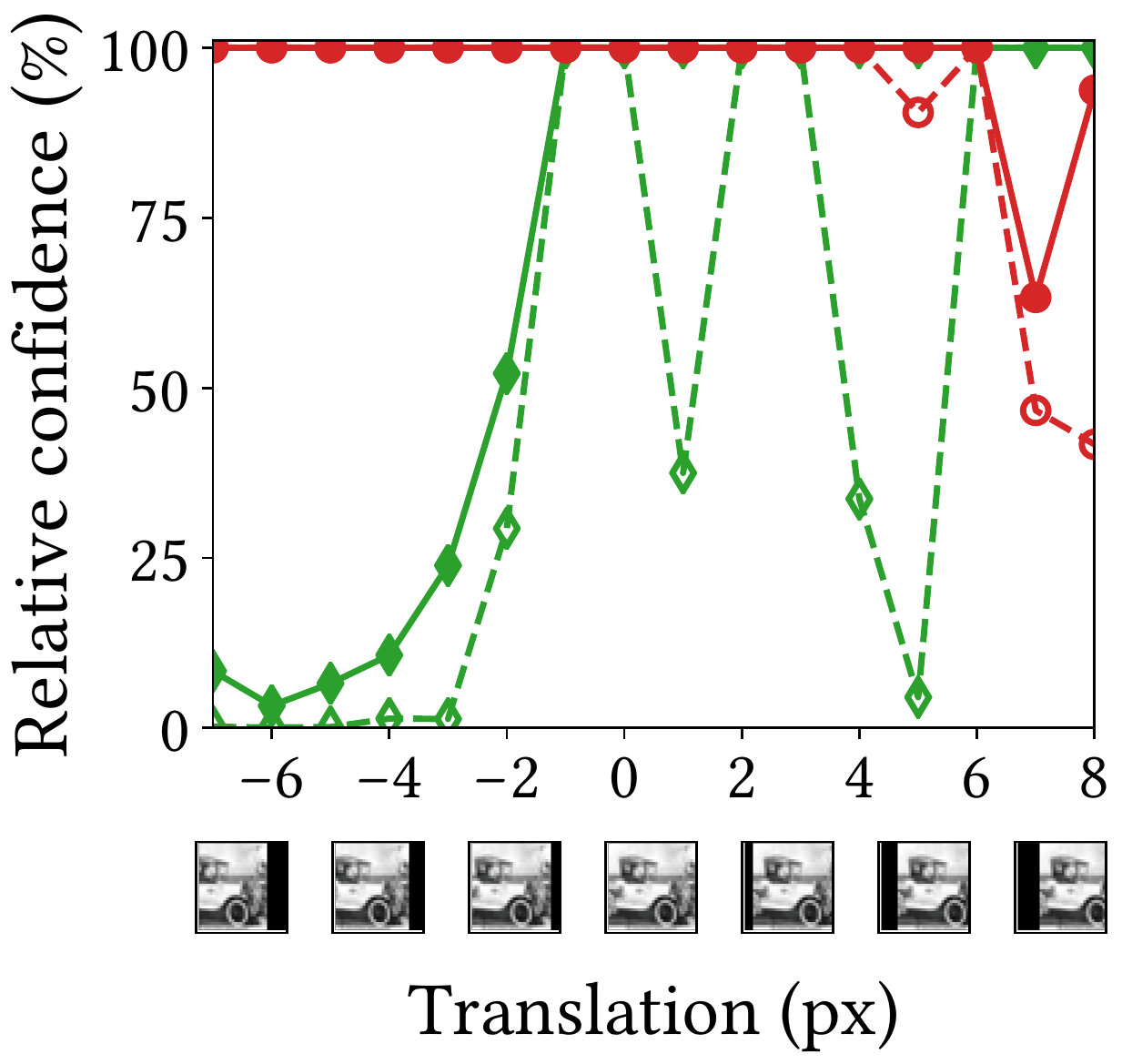}

\centering
\includegraphics[width=0.47\textwidth]{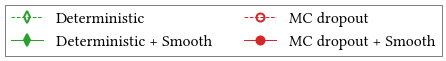}

\caption{
\textbf{Spatial smoothing improves the confidence \emph{when the predictions are incorrect}}.
We define relative confidence (See \cref{eq:relconf}), and measure the metric of ResNet-18 on CIFAR-10-P.
}
\label{fig:perturbation}

\vskip -0.15in

\end{figure}

%% file: appendix/prob-analysis.tex
\section{\texttt{Prob}s Play an Important Role in Spatial Smoothing}\label{sec:prob-role}

As discussed in \cref{sec:spatial-smoothing:architecture}, we take the perspective that each point in feature map is a prediction for binary classification by deriving the Bernoulli distributions from the feature map by using \texttt{Prob}. It is in contrast to previous works known as sampling-free BNNs \citep{hernandez2015probabilistic,wang2016natural,wu2018deterministic} attempting to approximate the distribution of feature map with one Gaussian distribution. We do not use any assumptions on the distribution of feature map, and exactly represent the Bernoulli distributions and their averages. 
However, sampling-free BNNs are error-prone because there is no guarantee that feature maps will follow a Gaussian distribution.

This \texttt{Prob} plays an important role in spatial smoothing. CNNs, such as VGG, ResNet, and ResNeXt, generally use post-activation arrangement. In other words, their stages end with \texttt{BatchNorm} and \texttt{ReLU}. 
Therefore, spatial smoothing layers $\texttt{Smooth} (\bm{z}) = \texttt{Blur} \circ \texttt{Prob} (\bm{z})$ in CNNs cooperates with \texttt{BatchNorm} and \texttt{ReLU} as follows:
\begin{align}
\texttt{Prob} (\bm{z}) 
&= \texttt{ReLU} \circ \texttt{tanh}_{\tau} \circ \texttt{ReLU} \circ \texttt{BatchNorm} \, (\bm{z}) \\
&= \texttt{ReLU} \circ \texttt{tanh}_{\tau} \circ \texttt{BatchNorm} \, (\bm{z})
\end{align}
since $\texttt{ReLU}$ and $\texttt{tanh}_{\tau}$ are commutative, and $\texttt{ReLU} \circ \texttt{ReLU}$ is $\texttt{ReLU}$. This $\texttt{Prob}$ is trainable and is a general form of \cref{eq:prob}. If we only use \texttt{Blur} as spatial smoothing, the activations \texttt{BatchNorm}--\texttt{ReLU} play the role of \texttt{Prob}.

\input{resources/fig-preact}

In order to analyze the roles of \texttt{Prob} and \texttt{Blur} more precisely, we measure the predictive performance of the model that does not use the post-activation. \Cref{fig:extended:preact} shows NLL of pre-activation VGG-16 on CIFAR-100. 
The result shows that \texttt{Blur} with \texttt{Prob} improves the performance, but \texttt{Blur} alone does not.
In fact, contrary to \citet{zhang2019making}, \emph{blur degrades the predictive performance since it results in loss of information}. We also measure the performance of VGG-19, ResNet-18, ResNet-50, and BlurPool \citep{zhang2019making} with pre-activation, and observe the same phenomenon. In addition, \texttt{BatchNorm}--\texttt{ReLU} in front of GAP significantly improves the performance of pre-activation ResNet. This observation also supports the claim.

As mentioned in \cref{sec:revisiting:preact}, pre-activation is a special case of spatial smoothing. Therefore, the performance improvement of pre-activation by spatial smoothing is marginal compared to that of post-activation.

%% file: resources/fig-preact.tex
\begin{figure}[ht]
\centering

\vskip 0.2in

\raisebox{0pt}[\dimexpr\height-0.6\baselineskip\relax]{
\includegraphics[width=0.37\textwidth]{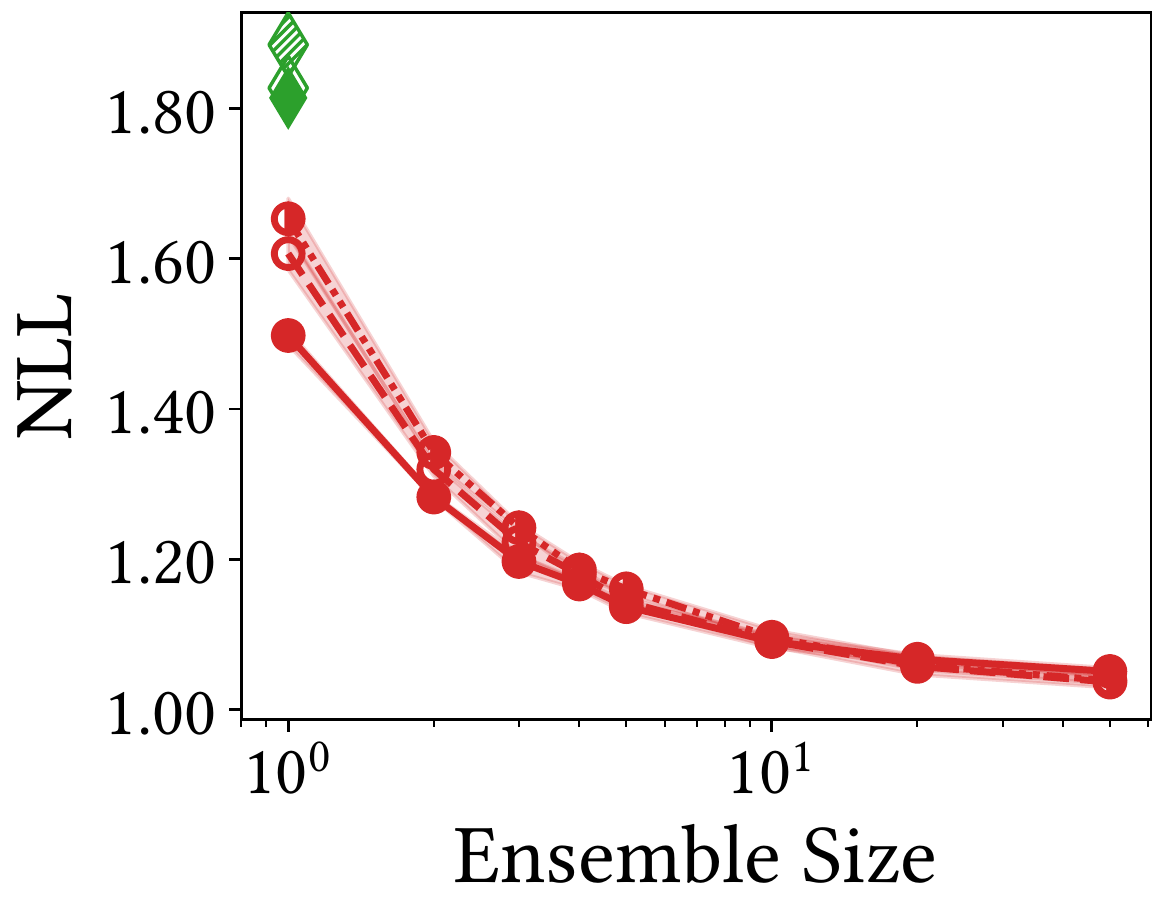}
}

\vspace{2pt}

\centering
\includegraphics[width=0.48\textwidth]{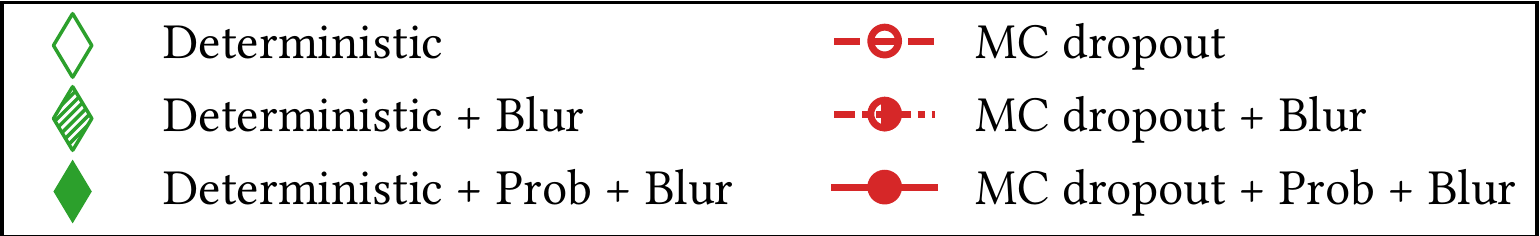}

\caption{
\textbf{\texttt{Blur} alone harms the predictive performance, although \texttt{Prob} + \texttt{Blur} improves it}.
We provide NLL of pre-activation VGG-16 on CIFAR-100.
}
\label{fig:extended:preact}
\end{figure}